\documentclass[10pt,twocolumn]{article}

\usepackage{graphicx}
\graphicspath{ {fig/} }

\usepackage{amsmath}
\usepackage{amssymb}
\usepackage{mathtools, cuted}	
\usepackage[toc,]{appendix}

\usepackage{tikz}
\usetikzlibrary{er,positioning,shapes,snakes,shadows,arrows, topaths,matrix,arrows,fit,intersections,shapes.arrows}

\makeatletter
\pgfkeys{/pgf/.cd,
  parallelepiped offset x/.initial=2mm,
  parallelepiped offset y/.initial=2mm
}
\pgfdeclareshape{parallelepiped}
{
  \inheritsavedanchors[from=rectangle] 
  \inheritanchorborder[from=rectangle]
  \inheritanchor[from=rectangle]{north}
  \inheritanchor[from=rectangle]{north west}
  \inheritanchor[from=rectangle]{north east}
  \inheritanchor[from=rectangle]{center}
  \inheritanchor[from=rectangle]{west}
  \inheritanchor[from=rectangle]{east}
  \inheritanchor[from=rectangle]{mid}
  \inheritanchor[from=rectangle]{mid west}
  \inheritanchor[from=rectangle]{mid east}
  \inheritanchor[from=rectangle]{base}
  \inheritanchor[from=rectangle]{base west}
  \inheritanchor[from=rectangle]{base east}
  \inheritanchor[from=rectangle]{south}
  \inheritanchor[from=rectangle]{south west}
  \inheritanchor[from=rectangle]{south east}
  \backgroundpath{
    \southwest \pgf@xa=\pgf@x \pgf@ya=\pgf@y
    \northeast \pgf@xb=\pgf@x \pgf@yb=\pgf@y
    \pgfmathsetlength\pgfutil@tempdima{\pgfkeysvalueof{/pgf/parallelepiped offset x}}
    \pgfmathsetlength\pgfutil@tempdimb{\pgfkeysvalueof{/pgf/parallelepiped offset y}}
    \def\ppd@offset{\pgfpoint{\pgfutil@tempdima}{\pgfutil@tempdimb}}
    \pgfpathmoveto{\pgfqpoint{\pgf@xa}{\pgf@ya}}
    \pgfpathlineto{\pgfqpoint{\pgf@xb}{\pgf@ya}}
    \pgfpathlineto{\pgfqpoint{\pgf@xb}{\pgf@yb}}
    \pgfpathlineto{\pgfqpoint{\pgf@xa}{\pgf@yb}}
    \pgfpathclose
    \pgfpathmoveto{\pgfqpoint{\pgf@xb}{\pgf@ya}}
    \pgfpathlineto{\pgfpointadd{\pgfpoint{\pgf@xb}{\pgf@ya}}{\ppd@offset}}
    \pgfpathlineto{\pgfpointadd{\pgfpoint{\pgf@xb}{\pgf@yb}}{\ppd@offset}}
    \pgfpathlineto{\pgfpointadd{\pgfpoint{\pgf@xa}{\pgf@yb}}{\ppd@offset}}
    \pgfpathlineto{\pgfqpoint{\pgf@xa}{\pgf@yb}}
    \pgfpathmoveto{\pgfqpoint{\pgf@xb}{\pgf@yb}}
    \pgfpathlineto{\pgfpointadd{\pgfpoint{\pgf@xb}{\pgf@yb}}{\ppd@offset}}
  }
}
\makeatother

\usepackage{xcolor}
\numberwithin{equation}{section}

\usepackage{natbib}

\usepackage{algorithm}
\usepackage{algorithmic}

\begin{document}
\title{Linear Algebra and Duality of Neural Networks}
\author{Galin Georgiev}
\date{June 1, 2015}

\maketitle

\begin{abstract} 
Bases, mappings, projections and metrics, natural for Neural network training, are introduced. Graph-theoretical interpretation is offered. Non-Gaussianity naturally emerges, even in relatively simple datasets. Training statistics, hierarchies and energies are analyzed, from physics point of view. \emph{Duality} between observables (for example, pixels) and observations is established. Relationship between exact and numerical solutions is studied. Physics and financial mathematics interpretations of a key problem are offered. Examples support all new concepts.
\end{abstract}

\tableofcontents
\section{Introduction}
\label{Introduction}
In modern Machine Learning, one typically studies a collection of \emph{observables} which are points in some N-dimensional ambient space. Examples are the collection of intensities of N  pixels in visual recognition, a collection of N frequency amplitudes derivatives in speech recognition, N stock prices or simply the collection of $N$ bits (or “spins chains”). They are often referred to also as \emph{''nodes''} or \emph{``states''}. If we have $P$  \emph{observations}, we are therefore looking at a $P \times N$ matrix $\mathbf X$ - the \emph{training}  matrix - where every row is an observation and every column is an observable. 

Notation-wise, we will refer to the collection $\{ \mathbf x_i\}, i = 1,..., N$ of the columns of the training matrix  as the collection of observables and it is generally highly correlated. The collection $\{ \mathbf x_\mu\}, \mu = 1,..., P$ of the rows of the training matrix will be referred to as the collection of (training) observations and also typically highly correlated among themselves. We will use \emph{latin} letters like $i,j,k$ etc for the index in the collection of observables and \emph{greek} letters like $\kappa, \mu, \nu $ etc for the index in the collection of observations 

In modern applications, $N$ values typically range in the hundreds or thousands but can easily be into the millions, for  1024 x 1024 pixel images for example. The dimension $P$ is typically in the thousands but can easily be in the millions. While P can in principle go to infinity, in practice, observations can be split up into batches describing different  phenomena. We will therefore assume that $P$ and $N$ are of similar order of magnitude.

When one attacks typical problems of Machine Learning like Classification, Interpolation (Regression), Dimension Reduction etc, one looks for robust structures in the collection of observables, assuming that observations are simply fleeting and often noisy snapshots  of these robust structures. The methods of Machine Learning are usually separated in two general buckets: 

i) In the so-called \emph{''generative''} methods like Bayesian Networks, Gaussian Mixture Models (GMM), Hidden Markov Models (HMM), Generative Neural Networks, etc., one is looking to compute a conditional probability distribution $Prob(\mathbf {x_{\mu}|X })_{\mu=1...P}$. The goal is  to generate the conditional probability for a new observation $Prob(\mathbf{x}_{\nu} | \mathbf{X})$, $\nu \not \in 1...P$.

ii) In the so-called \emph{''discriminatory''} methods like Logistic Regressions, Support Vector Machines (SVM), or Classifier Neural Networks, one estimates the mapping between the collection of observations  $Prob(\mathbf {x_{\mu}|X })_{\mu=1...P}$ and \emph{labels}. These methods do not naturally offer distributions of observations and hence do not allow to generate new observations.

The broader, generative methods, have \emph{learning time} more or less explicitly in their equations while discriminatory methods do not. But even when learning time is explicitly present, one is generally looking for “stability” and “robustness” of the structures and these concepts are associated with stationarity i.e. independence of time. Traditional Machine Learning is, in that sense, “equilibrium” Machine Learning i.e. looking for stationary structures. For this reason, with small exceptions, most of modern Machine Learning (including GMM and HMM) forces or assumes wrongly the \emph{independence} of observations. The approach makes a lot of sense when one deals with any fixed collection of observations: e.g. a collection of human faces or a collection of cats or a dictionary of words and we have reasonable success painstakingly ''machine learning'' those domains with their domain-specific methods. 

Unfortunately, the structures emerging within the different collections of observations, for example, music tunes on the one hand and human faces on the other, are completely unrelated to each other. There is no natural mapping between sensory collections like collections of pictures and collections of ''triphones'' (used in speech recognition), not to mention cognitive knowledge like chess-playing or the human ability to build abstract constructs. 

Since the human brain excellently manages all these distinct tasks, it is therefore generally believed that different parts of the brain have evolved and learned separately the skills for every specific task at hand. Because of the \emph{plasticity} of the brain, i.e. its ability to learn new tasks by brain areas which have been originally  designated for other tasks, there is a natural expectation that the learning approach is essentially universal. The only difference is that the input training data varies dramatically from task to task.

Problem is, after more than fifty years of trying really hard to replicate that hypothetical universal learning process, using machines, humans have not succeeded. We are getting close - in the last few years, speech and vision recognition, for example, were more or less united into an umbrella of similar Neural Networks, but they still require a myriad of idiosyncratic problem-specific techniques to perform well. Moreover, they are solving an essentially stationary problem (human language, for example,  is more or less a stationary set).

''Have we thrown the baby out with the bath water?'' asked pointedly in the late nineties David MacKay - one of the main backers of the generative Gaussian Mixture Models (GMM), \cite{MacKay98}. He was referring to the disappointing state of affairs at the time when the exciting and all-promising artificial Neural Networks  in the eighties had been shown to be mere smoothing devices via their equivalence to GMMs.

Well, it looks like we have. The ''connectionist'' approach to  Neural Networks (\cite{Bourlard93}, Ch. 5) took the wind out of the sails of the  Neural Networks by demonstrating that, with the then current computing capacity, and for the problems practically solvable at the time i.e. number of observables in the thousands (but no more!), Neural Networks do not really have any advantage over explicitly generative methods like HMM or GMM. It did not help that the single most dominant technique for ''training'' the  Neural Networks - Back-propagation - is believed to not be biologically plausible. This entirely justified  critique did not really offer viable generative, universal and biologically inspired alternatives. In  recent years, with the availability of more memory and ever more powerful massive parallel GPU computing, many research groups went back to the  Neural Networks skeleton closet and have had success improving many benchmarks in Machine Learning, using essentially the same back-propagation. They are becoming mainstream in industrial speech- and visual- recognition systems, but at their core, they appear to be better engineered copy-cats of the same miscreants the connectionists bemoaned loudly and rejected in the nineties.

\section{Definitions and notations.}
\label{Definitions and}
Matrices and tensors will be denoted with capital bold-faced letters like $\mathbf G$, vectors with regular bold-faced letters like $\mathbf q$, linear of affine spaces will be denoted with capitals like $\mathbb  R $ or $\mathbb O$.  

The training $P \times N$ matrix is $\mathbf {X }= \{X_{\mu i}\}_{\mu=1,...P, \\ i=1,...,N}$, with the collection of \emph{observables} - the column-vectors $\{ \mathbf{x}_{i}\}_{i=1}^{N}$ of  $\mathbf X$ and the collection of \emph{observations} - the row-vectors $\{ \mathbf{x}_{\mu}\}_{\mu=1}^{P}$ of $\mathbf X$.

Since our observables $\{ \mathbf{x}_{i}\}_{i=1}^{N}$ are column-vectors in $\mathbb{R}^P$, we will refer to $\mathbb{R}^P$ as the space of all, not necessarily training observables or simply \emph{observables space}. Similarly, the training observations  $\{ \mathbf{x}_{\mu}\}_{\mu=1}^{P}$ are row-vectors in $\mathbb{R}^N$ and we will refer to $\mathbb{R}^N$ as \emph{observation space}. Let us assume that the training matrix is of rank $M \leq min(N,P)$. We will call informally the M-dimensional  subspace spanned by the rows or columns of the training matrix \emph{training space}.  More formally, let us introduce the notion of M-dimensional linear \emph{space of training observables} $\mathbb{O} \approx \mathbb{R}^M \subseteq \mathbb{R}^P $  as the linear subspace of $\mathbb{R}^P$ spanned by the of observables $\{ \mathbf{x}_{i}\}_{i=1}^{N}$ .  Its linear dual space $\mathbb{O'} \approx \mathbb{R}^M \subseteq \mathbb{R}^N $ will be called the \emph{space of training observations} and is  spanned by the collection of training observations $\{ \mathbf{x}_{\mu}\}_{\mu=1}^{P}$. 

With the risk of abusing language, we will refer to arbitrary points in the space $\mathbb{O},$ which are not in the training set, as  \emph{hidden} training observables. Similarly, points in the space $\mathbb{O'},$ which are not in the training set will be referred to as \emph{hidden} training observations\footnote{In this sense, the training observations $\{ \mathbf{x}_{\mu}\}_{\mu=1}^{P}$ should strictly speaking be called \emph{visible} but we will often skip the adjective ``visible''.}. In other words both  $\mathbb{O}$ and  $\mathbb{O'}$ can be broken down into visible and hidden subsets:
\begin{align}
 \mathbb {O} &= \mathbb{O}_{visible} \cup \mathbb{O}_{hidden}, \nonumber \\
  \mathbb {O'} &= \mathbb{O'}_{visible} \cup \mathbb{O'}_{hidden}
\label{0.-1}
\end{align}
It is in principle of course possible that with  the increase of the training set, a hidden observation may become visible.

\begin{figure}[ht]
\vskip 0.2in
\begin{center}
\centerline{\includegraphics[width=\columnwidth]{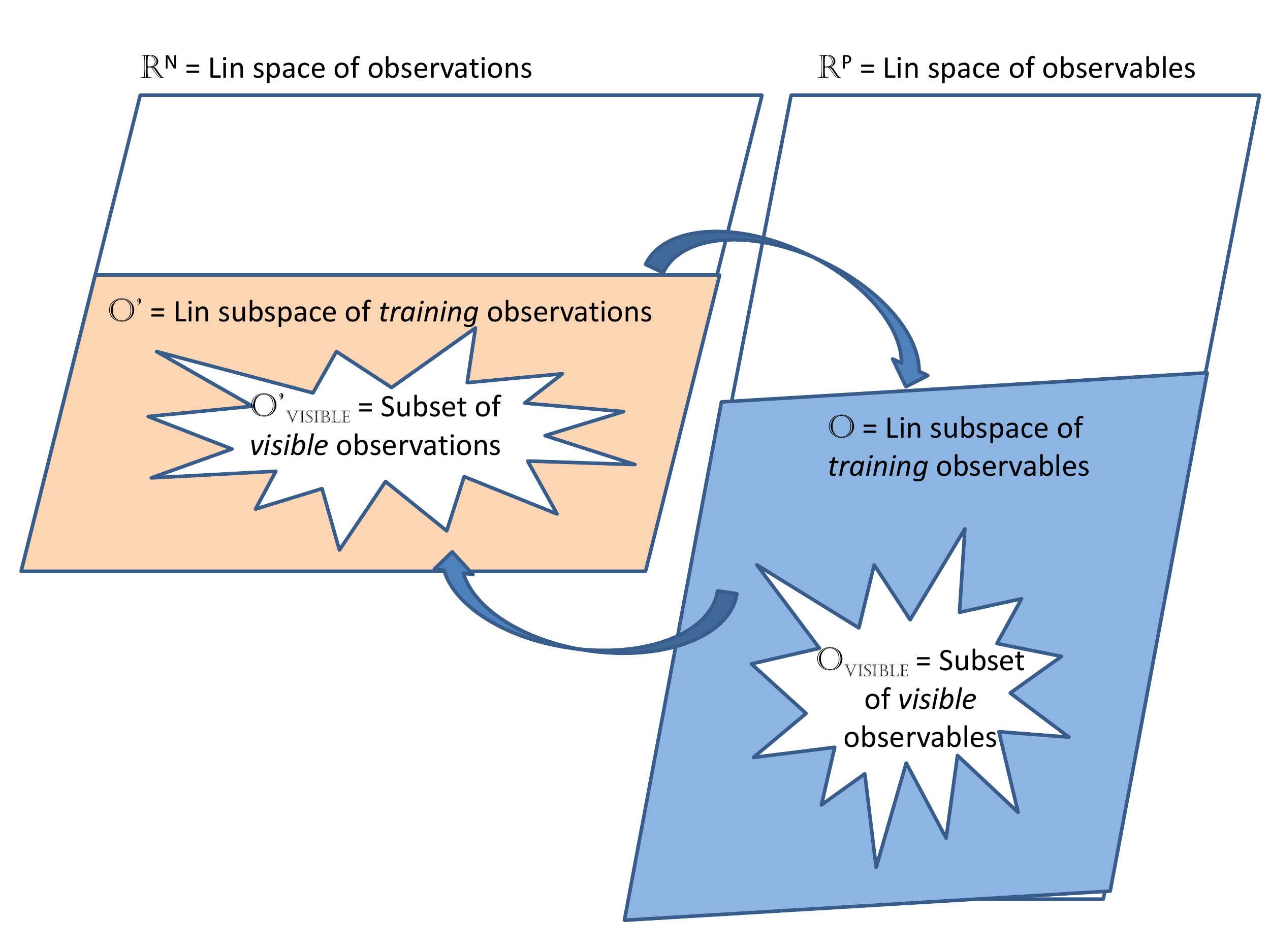}}
\caption{Hierarchy of the subsets and linear spaces related to the training matrix $\mathbf X$. The arrows indicate that $\mathbb O$ and $\mathbb O'$ are interchangeable , since they are isomorphic.}
\label{Fig3.-21}
\end{center}
\vskip -0.2in
\end{figure}

The hierarchy of the different sets and linear subspaces related to $\mathbf X$  is shown on Figure \ref{Fig3.-21}.

Note that an arbitrary observation i.e. a row-vector $\mathbf q' $ $ =[q'_1,q'_2,...,q'_N]$ $ \in \mathbb R^N$ is not necessarily a training observation in $\mathbb{O'}$, neither visible, nor hidden.

\section{The training set: bases, conjugates, projections, and metrics.}
\label{The training set}
We will analyze here the structure of the training set from pure linear-algebraic point of view and introduce some basic concepts like bases, overlaps, conjugates, projections and related to them Gram matrices,  metrics, etc. We will also suggest a \emph{graph-theoretical} interpretation of the training data  where training observations will be the graph \emph{vertices} and their overlaps will become the graph \emph{edges}.

\subsection{Basis.}
\label{Basis}

For a typical training dataset,  neither the training observations $\{ \mathbf{x}_{\mu}\}_{\mu=1}^{P}$, nor the training observables  $\{ \mathbf{x}_{i}\}_{\i=1}^{N}$ are independent among themselves, and hence do not form a basis. Having a basis comes in very handy, so we will introduce a basis for both the observations space $\mathbb{R}^N$ and observables space $\mathbb {R}^P$.  

For the observations space, the most obvious choice is the set of the ( likely hidden ) ``indicator'' observations $\{ \mathbf{e'}_i\}_{i=1}^N$ ,
\begin{align}
 \mathbf{e'}_1&= (1,0,...,0) , \nonumber \\
 \mathbf{e'}_2 &= (0,1,...,0) , \nonumber \\
 ... \nonumber \\
  \mathbf{e'}_N &= \underbrace{(0,0,...,1)}_{N}
\label{0.-2}
\end{align}
which are row-vectors and form an orthonormal basis of $\mathbb R^N$ i.e. $<\mathbf{e'}_i,\mathbf{e'}_j>_{\mathbb E}$ $=\delta_{ij}$ in the plain Euclidean metric\footnote{ $ \delta_{ij}$ is the Kronecker symbol: $ \delta_{ij} = 1$ , if $i=j$ and $0$ otherwise.}. One can express the training observations via this basis in an obvious way:
\begin{align}
\mathbf{x}_{\mu}=\sum_{i=1}^{N} X_{\mu i} \mathbf{e'}_i.
\label{0.-3}
\end{align}
and hence,
\begin{align}
<\mathbf{x}_{\mu}, \mathbf{e'}_i>_{\mathbb E} = X_{\mu i}
\label{0.-3.1}
\end{align}
is the $i$-th coordinate of $\mathbf{x}_{\mu}.$ Because of the completeness of the basis, two observations are identical if and only if their coordinates in this basis are the same. 

Similarly, the column-vectors $\{ \mathbf{e}_{\mu}\}_{{\mu}=1}^P$ ,
\begin{align}
 \mathbf{e}_1&= (1,0,...,0)^T , \nonumber \\
 \mathbf{e}_2 &= (0,1,...,0)^T , \nonumber \\
 ... \nonumber \\
  \mathbf{e}_P &= \underbrace{(0,0,...,1)^T}_{P}
\label{0.-2a}
\end{align}
form an orthonormal basis of the observables space $\mathbb {R}^P$. We have for the training observables:
\begin{align}
\mathbf{x}_{i}=\sum_{\mu=1}^{P} X_{\mu i} \mathbf{e}_{\mu}.
\label{0.-3a}
\end{align}
and hence, the ``dual'' to (\ref{0.-3.1}) identity holds:
\begin{align}
<\mathbf{x}_{i}, \mathbf{e}_{\mu}>_{\mathbb E} = X_{\mu i}.
\label{0.-3.a.1}
\end{align}

\subsection{Training mappings $\mathbb{T}$ and $\mathbb{T'}$. Overlaps.}
\label{Training mappings}

\begin{figure}[ht]
\vskip 0.2in
\begin{center}
\centerline{\includegraphics[width= \columnwidth]{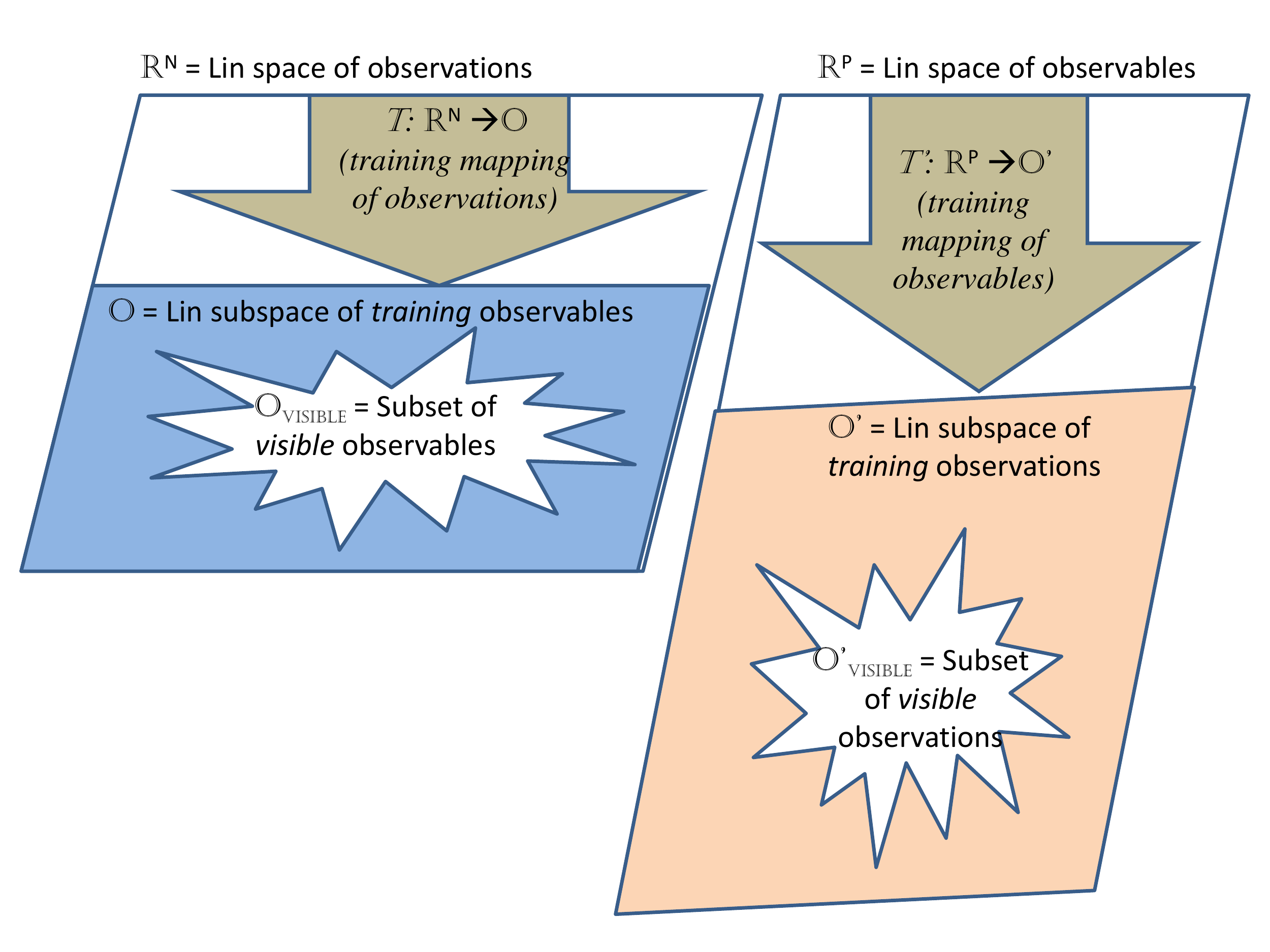}}
\caption{ On the left is the training mapping of  observations $\mathbb{T}$ from (\ref{0.-4}) and on the right is the training mapping  of observables $\mathbb{T'}$ from (\ref{0.-4a}). In the framework of Figure \ref{Fig3.-21}, we have swapped the isomorphic subspaces $\mathbb O$ and $\mathbb O' \cong \mathbb{R}^M$.}
\label{Fig3.-23}
\end{center}
\vskip -0.2in
\end{figure}

The introduction of basis in the observations and observables spaces paves the way for an elegant description of their respective training sub-spaces. The \emph{training mapping} of observations $\mathbb T$ is simply the multiplication from the left by $\mathbf{X}$ of the transposed observations: 
\begin{align}
\mathbb{T}: \mathbb{R}^N & \rightarrow \mathbb{O} \subseteq \mathbb{R}^P, \nonumber \\
\mathbb{T}\mathbf{e'}_i &= \mathbf{X}\mathbf{e'}_i^T = \mathbf{x}_i , \quad  i =1,...,N.
\label{0.-4}
\end{align}
where $\mathbf { X e'}_i^T$ is the matrix product of $\mathbf { X }$ and the column vector $\mathbf{e'}_i'^T$ (cf. the left side of Figure \ref{Fig3.-23}). It nicely maps the basis $\{ \mathbf{e'}_{i}\}_{i=1}^N$ of the  observation space $\mathbb{R}^N$ into the set of training observables $\{ \mathbf{x}_{i}\}_{i=1}^N$. Translated for an arbitrary observation  $\mathbf q' $  $ =\sum_{i=1}^N q'_i\mathbf{e'}_i$ $\in \mathbb{R}^N$, this reads:
\begin{align}
\mathbb{T}\mathbf{q'} &= \sum_{i=1}^N  \mathbf{x}_{i} q'_i = \mathbf{X q'}^T = \left\{\sum_{i=1}^N X_{\mu i} q'_i\right\}_{\mu=1}^P ,
\label{0.0a}
\end{align}
where $\mathbf { X q'}^T$ is the matrix product of  $\mathbf { X }$ and the column vector $\mathbf q'^T$.  The $\mu$-th coordinate of the training mapping of  observation $\mathbf q'$:
\begin{align}
(\mathbb{T}\mathbf{q'})_{\mu} = \mathbf { (Xq'^T)_{\mu}} = \sum_{i} q'_i X_{\mu i} = <\mathbf{x_{\mu}, q'} >_{\mathbb E} 
\label{0.1.0}
\end{align}
is called $\mu$-th \emph{training overlap}, or simply overlap, because it defines the proximity  between the readings of $\mathbf q'$ and the training observation $\mathbf{x}_{\mu}$ (cf. \cite{Coolen05}, (3.19)).

It is natural to ask what is the training mapping $\mathbb{T}\mathbf{x}_{\mu}$ of the $\mu$-th training observation $\mathbf{x}_{\mu}$? It turns out to be the $\mu$-th row of the \emph{Gram matrix } $\mathbf{G'}$  of training observations:
\begin{align}
\mathbf{G'} & :=\mathbf{X X^T}   ,  \quad
G'_{\mu \nu}  = \mathbf{<x_{\mu}, x_{\nu}>}_{\mathbb E} =  \sum_{i} X_{\mu i} X_{\nu i} . 
\label{0.1.1b}
\end{align} 
This follows directly from (\ref{0.-3}), (\ref{0.1.0}): the  $\nu$-th overlap  of $ \mathbf{x_{\mu}}$ is:
\begin{align}
 (\mathbb{T}\mathbf{x}_{\mu})_{\nu} =  (\mathbf{XX}^T)_{\mu \nu } = G'_{\mu \nu}.
\label{0.1.2c.1}
\end{align}
The training overlaps of the training observations with themselves can naturally be called  \emph{self-overlaps}.

The training mapping $\mathbb T$ is surjective but is not a projection in the linear-algebraic sense, \cite{Kostrikin89}: it is not \emph{idempotent} because $\mathbb{TT} \neq \mathbb{T}$. We will construct in (\ref{0.-5}), (\ref{0.-21})  genuine training projections $\mathbb {P, P'},$  so the distinction between a plain ``mapping'' and ``projection'' is not incidental (compare Figure \ref{Fig3.-23} and Figure \ref{Fig3.-25}).

The obvious counterpart of $\mathbb{T}$ for observables maps the basis vector $\mathbf{e}_{\mu}$ onto the training observation $\mathbf{x}_{\mu}$:
\begin{align}
\mathbb{T'}: \mathbb{R}^P & \rightarrow \mathbb{O'} \subseteq \mathbb{R}^N, \nonumber \\
 \mathbb{T'} \mathbf{e}_{\mu}&=  \mathbf{X }^T \mathbf{e}_{\mu} = \mathbf{x}_{\mu}^T , \quad  \mu =1,...,P.
\label{0.-4a}
\end{align}
It  maps the basis $\{ \mathbf{e}_{\mu}\}_{\mu=1}^P$ into the set of training observations $\{ \mathbf{x}_{\mu}\}_{\mu=1}^P$. We will also refer to $\mathbb T'$  as training mapping because it will be clear from the context whether the domain is space of observations or observables. For an arbitrary observable $\mathbf q $ $ =\sum_{\mu=1}^P q_{\mu}\mathbf{e}_{\mu}$  $ \in \mathbb R^P$:
\begin{align}
 \mathbb{T'} \mathbf{q}&= \sum_{\mu=1}^P  q_{\mu} \mathbf{x}_{\mu}^T  = \mathbf{X }^T \mathbf{q} = \left\{\sum_{\mu=1}^P X_{i\mu} q_{\mu}  \right\}_{i=1}^N.
\label{0.0b}
\end{align}
 The $i$-th coordinate of this mapping  :
\begin{align}
 (\mathbb{T'} \mathbf{q})_{i} = \mathbf {(X}^T\mathbf{q)}_i = \sum_{\mu}  X_{i\mu}q_{\mu} = <\mathbf{x}_{i}, \mathbf{q} >_{\mathbb E}.
\label{0.1.0a}
\end{align}
defines again a $i$-th training overlap, i.e. proximity between the readings of $\mathbf q$ and the training observable $\mathbf{x}_{i}$. The training mapping $\mathbb{T}' \mathbf{x}_{i}$ of the $i$-th training observable $\mathbf{x}_{i}$ is the $i$-th row of the Gram matrix  $\mathbf{G}$ of training observables:
\begin{align}
\mathbf{G  :=X^T X} , \qquad G_{ij} = \mathbf{<x_i, x_j>}_{\mathbb E} = \sum_{\mu} X_{\mu i} X_{\mu j},
\label{0.1a}
\end{align}
because the self-overlaps of observables are:
\begin{align}
 (\mathbb{T'}\mathbf{x}_{i})_{j} =  \sum_{\mu=1}^P X_{i \mu } X_{\mu j} = G_{ij} = <\mathbf{x}_{i}, \mathbf{x}_{j}>_{\mathbb E}.
\label{0.1.2e.1a}
\end{align}

\subsection{Conjugate and inverse observables/observations. }
\label{Conjugate and}

Let us now ``chain'' the training mappings $\mathbb{T}$ and $\mathbb{T'}$ and see if their composition amounts to anything? From the mappings definitions (\ref{0.0a}) , (\ref{0.1.0}), (\ref{0.0b}),  one has for an observation $\mathbf{q'} \in \mathbb{R}^N$:
\begin{align}
\mathbb{ T' T}: \mathbb{R}^N & \overset{\mathbb T} {\rightarrow}\mathbb{O} \overset{\mathbb T'}{\rightarrow} \mathbb{O'}, \nonumber \\
\mathbb{T'} \mathbb{T}\mathbf{q'} &= \sum_{\mu=1}^P  <\mathbf{x_{\mu}, q'} >_{\mathbb E}  \mathbf{x}_{\mu}^T ,
\label{0.0c}
\end{align}
which looks awfully similar to a \emph{Fourier decomposition} but is not, because $\{\mathbf{x}_{\mu}\}$ do not form an orthonormal basis. Using the matrix form of the training mappings, one can re-write the training composition $\mathbb{T}' \mathbb{T}$ as:
\begin{align}
\mathbb{T'} \mathbb{T}\mathbf{q'}  &=  \mathbf{X}^T\mathbf{Xq'} = \mathbf{Gq'}  ,
\label{0.0c.-1}
\end{align}
 where $\mathbf G$ is the  Gram matrix (\ref{0.1a}) of the training observables. The inverse of an observation $\mathbf{q'}$ under the  training composition (if it exists) will be called its \emph{conjugate}  $\mathbf{\check{q}'}$:
\begin{align}
\mathbb{T'} \mathbb{T}\mathbf{\check{q'}} &:= \mathbf{q'} .
\label{0.0d}
\end{align}
Because $\mathbb{T, T'}$ are in general surjective mappings, the conjugate is not uniquely defined. We will make a special choice inspired by (\ref{0.0c.-1}). Let us  assume for simplicity that $rank(\mathbf{X}) = M = N$. Then $\mathbf{G}^{-1}$ is well defined\footnote{In the  case when $rank(\mathbf{X}) = M$ $< N \leq P$, one has to work with the restricted, with rank = M, version of the inverse matrix $\mathbf{G'}^{-1}$, defined via (\ref{1.1.3}).\label{fn1.1.3}} and for any observation  $\mathbf{q}'  \in \mathbb R^N$, its  conjugate $\mathbf{\check{q'}}$ from (\ref{0.0c.-1}), (\ref{0.0d}) will be chosen to equal:
\begin{align}
\mathbf{\check{q'}}^T &=   \mathbf{G}^{-1} \mathbf{q'}^T. 
\label{0.1.0.1b}
\end{align}
 The conjugate observation $\mathbf{\check{q'}}$  is in other words the \emph{covariant} vector corresponding to the original \emph{contravariant} vector $\mathbf q'$ in the metric defined by the metric tensor $\mathbf {G} ^{-1}$. The conjugates of the training observations  $\{\mathbf{\check{x}}_{\mu}\}$ form the rows of the $P \times N $ \emph{left conjugate} training matrix $\mathbf{\check{X}}$:
 \begin{align}
\mathbf{\check{X}}^T &=  \mathbf{G}^{-1} \mathbf{X}^T . 
\label{0.1.0.1c}
\end{align}
The ``left'' refers to the fact that $\mathbf{\check{X}}^T$ is exactly the left inverse of $\mathbf{X}$:
  \begin{align}
\mathbf{\check{X}}^T \mathbf{X} &=  \mathbf{G}^{-1} \mathbf{X}^T \mathbf{X} = \mathbf{I}_N, 
\label{0.1.0.1d}
\end{align}
i.e. the columns of $\mathbf{\check{X}}$ \emph{invert} the training observables (columns of $\mathbf X$).\footnote{The  alternative product $\mathbf{\check{X}} \mathbf{X}^T$ forms the important training projection matrix $\mathbf P'$ introduced in (\ref{0.-5.-2}).}. The rows of $\mathbf{\check{X}}$ are by construction  conjugate training observations, while, its columns are \emph{inverse training observables}.

In general, an observation and its conjugate are very different because of the  highly non-trivial nature of the Gram matrix $\mathbf G$ (see Figure \ref{Fig3.-1b} for examples). There is nevertheless a special class of  hidden observations which equal their conjugates, up to a scaling factor. They are the so-called \emph{eigen-observations} of $\mathbf G$ introduced in (\ref{1.1.1}). 

Switching to observables, one has in full analogy, for any observable $\mathbf{q} \in \mathbf{R}^P$:
\begin{align}
\mathbb{T} \mathbb{T'} \mathbf{q} &= \sum_{\i=1}^N  <\mathbf{x_{i}, q} >_{\mathbb E}  \mathbf{x}_{i}  ,
\label{0.0c.1}
\end{align}
and the image of  $\mathbf{q}$ under the  training composition is again called its conjugate  $\mathbf{\check{q}}$:
\begin{align}
\mathbb{T} \mathbb{T'} \mathbf{\check{q}}  &= \mathbf{q} = \mathbf{G'}\mathbf{\check{q}}.
\label{0.0d.1}
\end{align}
where $\mathbf G'$  is the Gram matrix (\ref{0.1.1b}) of  training observations. Similarly to observations, we will  assume that the inverse $\mathbf{G'}^{-1}$ is well defined\footnote{In the  case when $rank(\mathbf{X}) = M$ $< P \leq N$, one  has to work with the restricted version of the inverse matrix $\mathbf{G'}^{-1}$ - see footnote \ref{fn1.1.3}.}. Then  for any observable  $\mathbf{q}  \in \mathbb R^P$, its  conjugate $\mathbf{\check{q}}$ from (\ref{0.0d.1}) exists and is defined as:
\begin{align}
\mathbf{\check{q}} &= \mathbf{G'}^{-1} \mathbf{q}. 
\label{0.1.1b.1}
\end{align}
 The conjugate observable $\mathbf{\check{q}}$  is in other words the \emph{covariant} vector corresponding to the original \emph{contravariant} vector $\mathbf q$ in the metric defined by the metric tensor $\mathbf {G'} ^{-1}$.  In analogy with observations, the conjugates of the training observables  $\{\mathbf{\check{x}}_{i}\}$ form the columns of the $P \times N $ \emph{right conjugate} training matrix $\mathbf{\check{X'}}$:
 \begin{align}
\mathbf{\check{X'}} &=  \mathbf{G'}^{-1}\mathbf{X}, 
\label{0.1.1b.1a}
\end{align}
where $\mathbf{\check{X'}}^T$ is  the right inverse of $\mathbf{X}$:
  \begin{align}
 \mathbf{X}\mathbf{\check{X'}}^T &=  \mathbf{X X}^T \mathbf{G'}^{-1} = \mathbf{I}_P. 
\label{0.1.1b.1b}
\end{align}
The rows of $\mathbf{\check{X'}}$ are in other words orthogonal to the  training observations and can be referred to as   \emph{inverse training observations}. There is again a special class of  \emph{eigen-observables} which equal their conjugates,  up to a scaling factor (cf. (\ref{1.1.1b})).

\subsection{Example: MNIST dataset.}
\label{Example: MNIST}

\begin{figure}[!ht]
\vskip 0.2in
\begin{center}
\centerline{\includegraphics[width= \columnwidth]{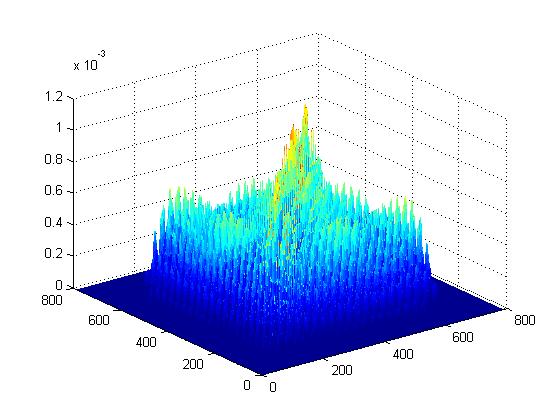}}
\caption{$\mathbf{G}$ for the first 5,000 images from  MNIST dataset (i.e., $P = 5,000$).}
\label{Fig3.-10}
\end{center}
\vskip -0.2in
\end{figure}

We plot the Gram matrix $\mathbf G$ for a part of the popular MNIST dataset in Figure \ref{Fig3.-10} (the dataset is composed of 60,000 training and 10,000 testing  images of the digits from 0 to 9, ordered randomly; every digit is displayed in a grid of $28 \times 28 $ $=784$ pixels; see \cite{LeCun98}, where dataset is defined).

\begin{figure}[!ht]
\vskip 0.2in
\begin{center}
\centerline{\includegraphics[width= \columnwidth]{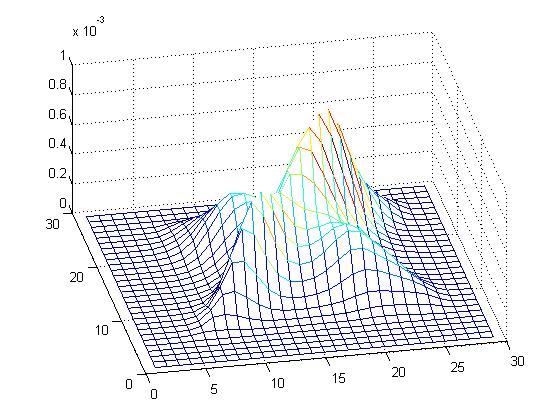}}
\caption{The $28$ pixel sub-block of $\mathbf{G}$ comprising the middle row $14$ on the $28 \times 28$ pixel grid ($P = 5,000$).}
\label{Fig3.-9}
\end{center}
\end{figure}

\begin{figure}[!ht]
\vskip 0.2in
\begin{center}
\centerline{\includegraphics[width= \columnwidth]{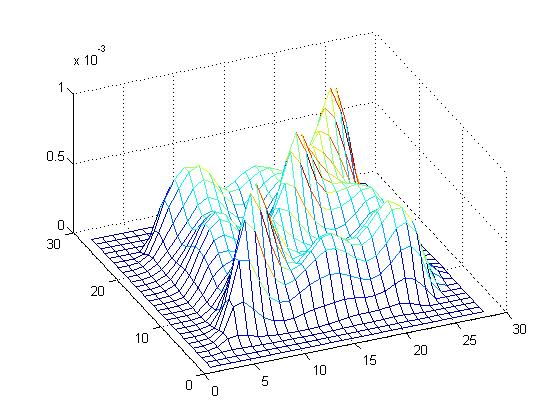}}
\caption{The $28$ pixel sub-block of $\mathbf{G}$ comprising the middle column $14$ on the $28 \times 28$ pixel grid ($P = 5,000$).}
\label{Fig3.-8}
\end{center}
\vskip -0.2in
\end{figure}

The number of observables i.e. pixels in this dataset is $N = 28 \times 28$ $=784$ and the digits from 0 to 9 are randomly dispersed throughout the dataset. One can appreciate better the overall symmetries in the dataset by examining separately two special sub-blocks of $\mathbf G$: i) the $28$ pixels comprising the middle row 14 on the grid (Figure \ref{Fig3.-9}) and ii) the $28$ pixels comprising the middle column $14$ of the grid (Figure \ref{Fig3.-8}). 

In both cases, we have dominant diagonal and first few sub-diagonals because the neighboring pixels in both the horizontal and vertical direction tend to ``fire-up'' together in humanly recognizable images. The extra peak on either side in Figure \ref{Fig3.-9} is due to the fact that a few digits, namely, $0,1,8$ are  (on average) symmetric with respect to the middle vertical line. The extra three peaks on either side in Figure \ref{Fig3.-8} are due to the fact that the digits $2,3,4,5,6,8,9$ are  (on average)  ``simultaneously busy'' in the top , middle and bottom section of middle vertical line.

\begin{figure}[!ht]
\vskip 0.2in
\begin{center}
\centerline{\includegraphics[width= \columnwidth]{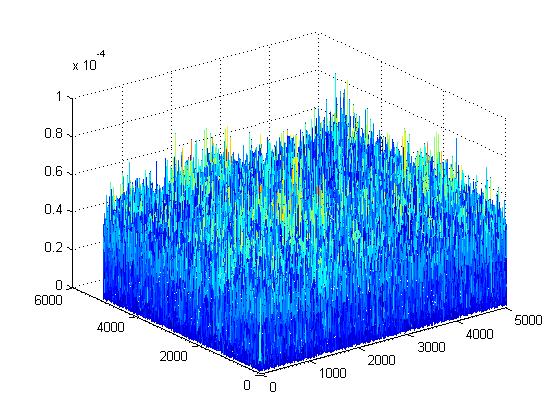}}
\caption{$\mathbf{G'}$ for the first 5,000 MNIST images (i.e., $P = 5,000$).}
\label{Fig3.0}
\end{center}
\end{figure}

\begin{figure}[!ht]
\begin{center}
\centerline{\includegraphics[width= \columnwidth]{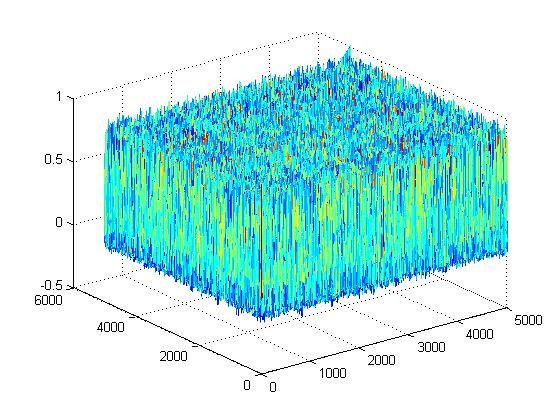}}
\caption{Correlation matrix (Pearson) of the first 5,000 MNIST images (i.e., $P = 5,000$), treated as random variables in $\mathbb{R} ^N$.}
\label{Fig3.-1}
\end{center}
\vskip -0.2in
\end{figure}

We also plot in Figure \ref{Fig3.0} the matrix $\mathbf G'$ for the first 5,000 observations of the  MNIST dataset introduced above. Because the random order of the digits in the dataset, unlike $\mathbf G$, the matrix $\mathbf G'$ has no visible structure or  any symmetries to speak of. Humanly recognizable images in a typical dataset, including MNIST, tend to be very highly correlated between themselves in the Euclidean metric in $\mathbb{R}^N$ (Figure \ref{Fig3.-1}).  

\begin{figure}[!ht]
\vskip 0.2in
\begin{center}
\centerline{\includegraphics[width= \columnwidth]{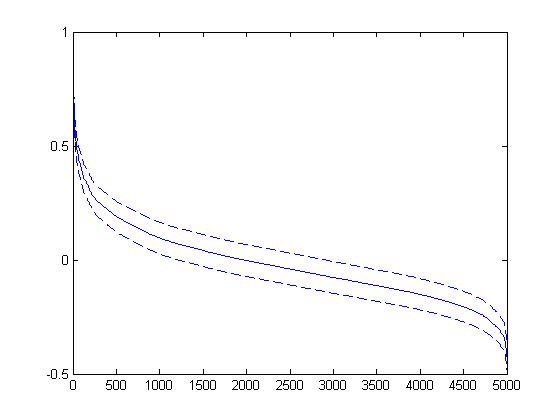}}
\caption{ The first off-diagonal of the correlation matrix from Figure \ref{Fig3.-1}, sorted in descending order ($P = 5,000$). The dashed lines are the corresponding 95\% confidence intervals. }
\label{Fig3.-1a}
\end{center}
\vskip -0.2in
\end{figure}

To appreciate the actual magnitudes better, we plot on Figure \ref{Fig3.-1a} only the pairwise correlations between  MNIST images which are neighbors in the original dataset order. The correlations are sorted subsequently in descending order. The expected obvious dependence between observations runs counter to the common assumption of independent observations in Neural Networks, which result in target minimization functions averaged uniformly across all observations (see the Introduction to Section \ref{Statistics and hierarchy} for more details).

\begin{figure}[!ht]
\vskip 0.2in
\centerline{\includegraphics[width= \columnwidth]{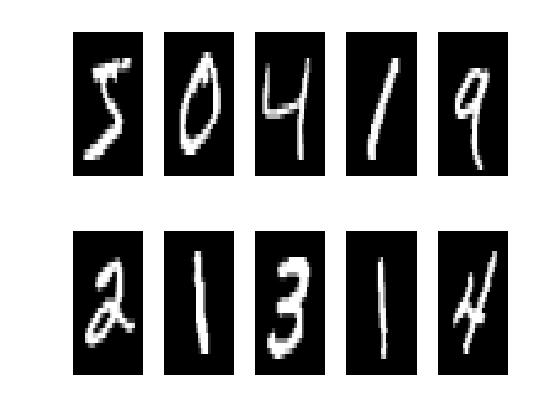}}
\vskip 0.2in
\centerline{\includegraphics[width= \columnwidth]{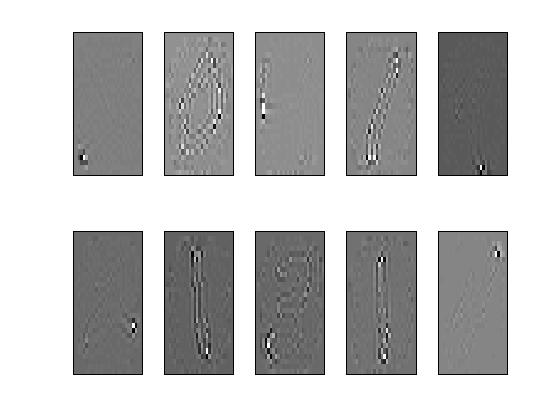}}
\caption{The first ten observations $\{ \mathbf{x}_i\}_{i=1}^{10}$ (top) in the original order of MNIST ($P = 5,000$) and their respective conjugate observations $\{ \mathbf{\check{x}}_i\}_{i=1}^{10}$ (bottom). The conjugates are spatially very localized and highlight outstanding parts of the original images.}
\label{Fig3.-1b}
\vskip -0.2in
\end{figure}

To demonstrate the nature of the conjugates, we plot on Figure \ref{Fig3.-1b} the first ten images from MNIST and their respective conjugate images.

 We also plot on Figures \ref{Fig1.2.2} - \ref{Fig1.2.3} the distributions of the training mappings of the so-called eigen-observations from Figure \ref{Fig1.2.0}.

\subsection{Training projections  $\mathbb {P, P'}$ .}
\label{Training projections}

\begin{figure}[!ht]
\vskip 0.2in
\begin{center}
\centerline{\includegraphics[width=\columnwidth]{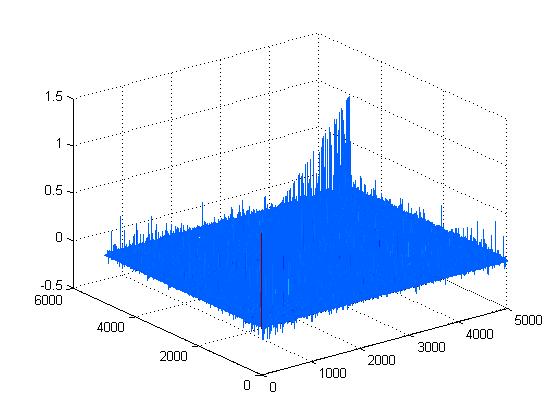}}
\vskip 0.2in
\centerline{\includegraphics[width=\columnwidth]{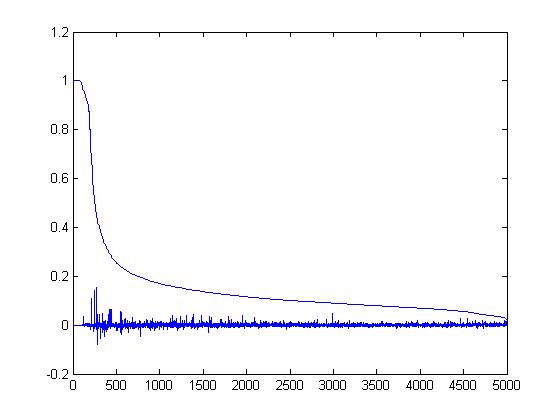}}
\caption{On the top chart is the training projection matrix $\mathbf{P'}$ for the first 5,000 MNIST images (i.e., $P = 5,000$). On the bottom chart are its diagonal $P'_{\mu \mu}$ (solid line) and first off-diagonal $P'_{\mu, \mu+1}$ (dashed line) elements, sorted in descending order of the diagonal elements.}
\label{Fig3.-26}
\end{center}
\vskip -0.2in
\end{figure}

Let us start with the case $P \geq N$ and  assume for simplicity that $rank(\mathbf{X}) = M = N$ and $\mathbf{ G}^{-1}$ exists. The $P \times P$ \emph{training projection matrix} is defined as:
 \begin{align}
 \mathbf{P'} :=\mathbf{\check{X}} \mathbf{X}^T = \mathbf{X G}^{-1}\mathbf{X}^T = \mathbf{X (X}^T \mathbf{X)}^{-1}\mathbf{X}^T
 \label{0.-5.-2}
\end{align} 
 with matrix elements:
\begin{align} 
 P'_{\mu \nu} = \mathbf{x}_{\mu} \mathbf{G}^{-1}\mathbf{x}_{\nu}^T,
 \label{0.-5.-1}
\end{align}
and  the following properties:
\begin{align}
 &i) \quad \mathbf{P'}  = \mathbf{P'}^T,  & \tag*{symmetric} \nonumber \\
 &ii) \quad \mathbf{P'} ^2 = \mathbf{P'}, \quad (\mathbf{I}_{P} - \mathbf{P}) ^2 = \mathbf{I}_{P} - \mathbf{P'},  & \tag*{projection} \nonumber \\
 & iii) \quad (\mathbf{I}_{P} - \mathbf{P'}) \perp \mathbf{P'},  & \tag*{orthogonal projection} \nonumber \\
 &iv) \quad \mathbf{P'X} = \mathbf{X}, (\mathbf{I}_{P} - \mathbf{P'})\mathbf{X} = 0.  & \tag*{invariant on \bf{X}} \\ 
\label{0.-6}
\end{align}
This matrix emerges naturally in multi-dimensional linear regression where it is often called \emph{hat matrix}\footnote{ For a multi-dimensional linear regression model $ \mathbf{y}$ $\sim \mathbf{x B} + \boldsymbol{\varepsilon}$, where $y \in \mathbb{R}^P$ and $\mathbf{B}$ is $ N \times N$ matrix, the model-predicted values, often denoted by $\mathbf{\hat{y}}$, are related to the empirical values $\mathbf y$ via $\mathbf{\hat{y}} = \mathbf {P' y}$ and the residuals are therefore $\boldsymbol{\varepsilon} = (\mathbf {I}_{P} - \mathbf{P') y}$.\label{MultLinReg}} (cf. \cite{Hamilton94}, Section 8). The matrix $\mathbf{P'}$ is plotted on Figure \ref{Fig3.-26} for the first 5,000 MNIST observations: compared to the Gram matrix $\mathbf{G'}$ with same dimension from Figure \ref{Fig3.0}, it is a lot more sparse-looking and will converge to $\mathbf{I}_N$ when $P \rightarrow N$. 

This symmetric projection matrix has the beautiful property of being the observation Gram matrix:
\begin{align}
\mathbf {P'} = \mathbf{V V} ^T , \qquad \mathbf{V}^T \mathbf{V} =\mathbf{I}_n,
\label{0.-6a}
\end{align}
of some $ P \times n$ data matrix $\mathbf V$, whose corresponding observables are orthonormal ($\mathbf{V}^T \mathbf{V} =\mathbf{I}_n$) , and   $n < P$ is the rank of projection matrix. This follows from the diagonalization property of symmetric matrices and the fact that the eigenvalues of a projection matrix equal either $1$ or $0$ (\cite{Hamilton94}, (8.1.20), (8.1.21)). When we introduce \emph{ singular value decomposition} in Sub-section \ref{Singular value}, we will recognize the training projection matrix as the Gram matrix of the left singular matrix $\mathbf V$ of $\mathbf X$ (\ref{1.1.3}). 

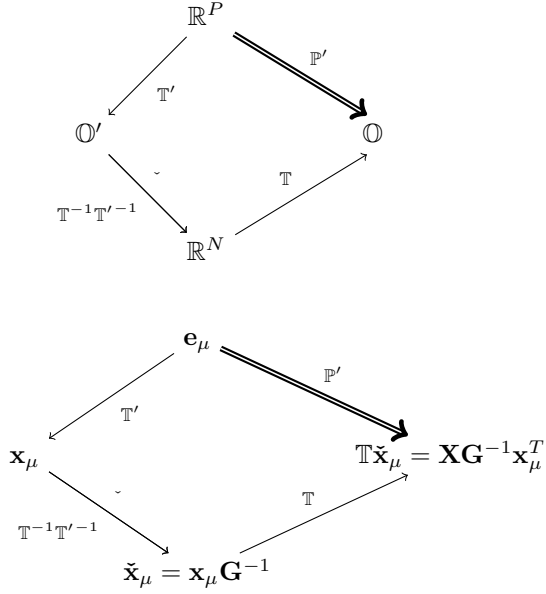
\begin{figure}[!ht]
\vskip 0.2in
\begin{center}
\begin{tikzpicture}[description/.style={fill=white,inner sep=2pt}]
\matrix (m) [matrix of math nodes, row sep=3em,column sep=2.5em, text height=1.5ex, text depth=0.25ex]
{  & \mathbb{R}^P & \\
\mathbb{O'} & & \mathbb{\qquad O  \qquad \qquad} \\
& \mathbb{R}^N &  \\};
\path[->,font=\scriptsize]
(m-1-2) edge node[auto] {$ \mathbb{T'} $} (m-2-1)
		  edge[double, thick] node[auto] {$ \mathbb{P'} $} (m-2-3)
(m-2-1) edge node[auto] {$ \check{} $} (m-3-2)
        edge node[auto,swap] {$ \mathbb{T}^{-1}\mathbb{T'}^{-1} $} (m-3-2)
(m-3-2) edge node[auto] {$ \mathbb{T} $} (m-2-3);
\end{tikzpicture}
\end{center}
\begin{center}
\begin{tikzpicture}[description/.style={fill=white,inner sep=2pt}]
\matrix (m) [matrix of math nodes, row sep=3em,column sep=2.5em, text height=1.5ex, text depth=0.25ex]
{  & \mathbf{e}_{\mu} & \\
\mathbf{x}_{\mu} & & \mathbb{T}\mathbf{\check{x}}_{\mu} = \mathbf{X}\mathbf{G}^{-1}\mathbf{x}_{\mu}^T \\
& \mathbf{\check{x}}_{\mu} = \mathbf{x}_{\mu}\mathbf{G}^{-1} &  \\};
\path[->,font=\scriptsize]
(m-1-2) edge node[auto] {$ \mathbb{T'} $} (m-2-1)
		  edge[double, thick] node[auto] {$ \mathbb{P'} $} (m-2-3)
(m-2-1) edge node[auto] {$ \check{} $} (m-3-2)
        edge node[auto,swap] {$ \mathbb{T}^{-1}\mathbb{T'}^{-1} $} (m-3-2)
(m-3-2) edge node[auto] {$ \mathbb{T} $} (m-2-3);
\end{tikzpicture}
\caption{Training projection $\mathbb{P'}$ in terms of linear spaces (top) and in terms of basis vectors (bottom).}
\label{Fig3.-24}
\end{center}
\vskip -0.2in
\end{figure}

There is in fact a whole family of projections which satisfy (\ref{0.-6}) and (\ref{0.-6a}). Pick an arbitrary $P \times P$  orthogonal matrix $\mathbf S$ which preserves $\mathbf X$, i.e. $\mathbf{S X} = \mathbf {S}^T\mathbf{X} $ $=\mathbf X$ when acting from the left\footnote{A multiplication of $\mathbf V$ from the right by an orthogonal $N \times N$ matrix $\mathbf S$, i.e. a rotation in the space of observables $\mathbb O \cong \mathbb O'$, does not generate a new projection, because $\mathbf{(VS) (VS)}^T $ $ = \mathbf{V S S} ^T \mathbf{ V}^T = \mathbf{P}'$.}. Then the corresponding data matrix $\mathbf{\hat{V}} = \mathbf{SV}$ and projection matrix $\mathbf{\hat{P'}} = \mathbf{SP'S}^T$ still satisfy (\ref{0.-6}) and (\ref{0.-6a}):
\begin{align}
\mathbf {\hat{P'} X} = \mathbf{\hat{V} \hat{V}} ^T \mathbf{X} = \mathbf{X} , \qquad \mathbf{\hat{V}}^T \mathbf{\hat{V}} =\mathbf{I}_n.
\label{0.-6b}
\end{align}
In the general case, when  an arbitrary $P \times P$  orthogonal matrix $\mathbf S$ does not preserve the observation space $\mathbb O'$ and  $\mathbf X$, i.e. $\mathbf{S X} = \mathbf {S}^T\mathbf{X} $ $ \neq \mathbf X$, the corresponding data matrix  $\mathbf{\hat{V}} = \mathbf{SV}$ still has orthonormal observables i.e. $\mathbf{\hat{V}}^T \mathbf{\hat{V}} =\mathbf{I}_n$.  In the general case though, the Gram matrix $\mathbf{\hat{V} \hat{V}} ^T$ - while still a projection matrix - projects on a different space. It is  a projection on the rotated observation space,  spanned by $ \mathbf{SX}$:
\begin{align}
\mathbf {\hat{P'} (SX)} = \mathbf{\hat{V} \hat{V}} ^T \mathbf{(S X)} =  \mathbf{S P' X} = \mathbf{SX} , \qquad \mathbf{\hat{V}}^T \mathbf{\hat{V}} =\mathbf{I}_n.
\label{0.-6c}
\end{align}
The projection matrix $\mathbf{P'}$ defines a corresponding \emph{training projection} $\mathbb P'$ on $\mathbb R ^P$ in an obvious way :
\begin{align}
\mathbb{P'} &: \mathbb{R}^P  \rightarrow \mathbb{O} , \nonumber \\
\mathbb{P'}\mathbf{q} & := \mathbf{P}' \mathbf{q}  = \mathbf{X G}^{-1}\mathbf{X}^T \mathbf{q} .
\label{0.-5}
\end{align}
The training projection $\mathbb{P'}$ can be thought of as \emph{Dimension reduction} which is especially violent for the typical case $P \gg N$ (see Figure \ref{Fig3.-27}). It is shown  diagrammatically via the training mappings $\mathbb{T,T'}$ and the conjugation $\check{ }$ on the top of Figure \ref{Fig3.-24}. 

On the bottom of Figure \ref{Fig3.-24} is shown the action of $\mathbb P'$ on  the $\mathbb{R} ^P$-orthonormal basis of column-vectors $\{ \mathbf{e}_{\mu}\}_{{\mu}=1}^P,$  introduced in (\ref{0.-2a}). The training projection of $\mathbf{e}_{\mu}$ is the $\mu$-th row of the training projection matrix $\mathbf{P'}$. In coordinate terms,
\begin{align}
(\mathbb{P'}\mathbf{e}_{\mu})_{\nu} = (\mathbf{P'}\mathbf{e}_{\mu})_{\nu} = (\mathbb{T} \mathbf{\check{x}}_{\mu})_{\nu} = \mathbf{x}_{\mu} \mathbf{G}^{-1}\mathbf{x}_{\nu}^T = P'_{\mu \nu} ,
\label{0.-7}
\end{align}
  which is exactly the conjugate self-overlap of $\mathbf{x}_{\mu}$ - compare with the self-overlap $(\mathbb{T}\mathbf{x}_{\mu})_{\nu}$ $=G'_{\mu \nu}$ from (\ref{0.1.2c.1}). This means that the squared Euclidean norm of the training projection  $||\mathbb{P'}\mathbf{e}_{\mu} ||_{\mathbb E}^2 $ measures the sum of the squared overlaps, hence the proximity, of the conjugate training observation $\mathbf {\check{x}}_{\mu}$ with all other training observations. Because $\mathbf{P'}$ is a projection (property ii) in (\ref{0.-6})),
 \begin{align}
||\mathbb{P'}\mathbf{e}_{\mu} ||_{\mathbb E}^2 = \sum_{\nu} (P'_{\mu \nu})^2 = P'_{\mu \mu} ,
\label{0.-7.1}
\end{align}
the diagonal element $P'_{\mu \mu} $ $= \mathbf{x}_{\mu} \mathbf{G}^{-1}\mathbf{x}_{\mu}^T$ itself measures the overlap of the conjugate $\mathbf {\check{x}}_{\mu}$ with all other training observations - see for the MNIST example Figures \ref{Fig3.-26} and \ref{Fig3.-27}.

\begin{figure}[!ht]
\vskip 0.2in
\begin{center}
\centerline{\includegraphics[width=\columnwidth]{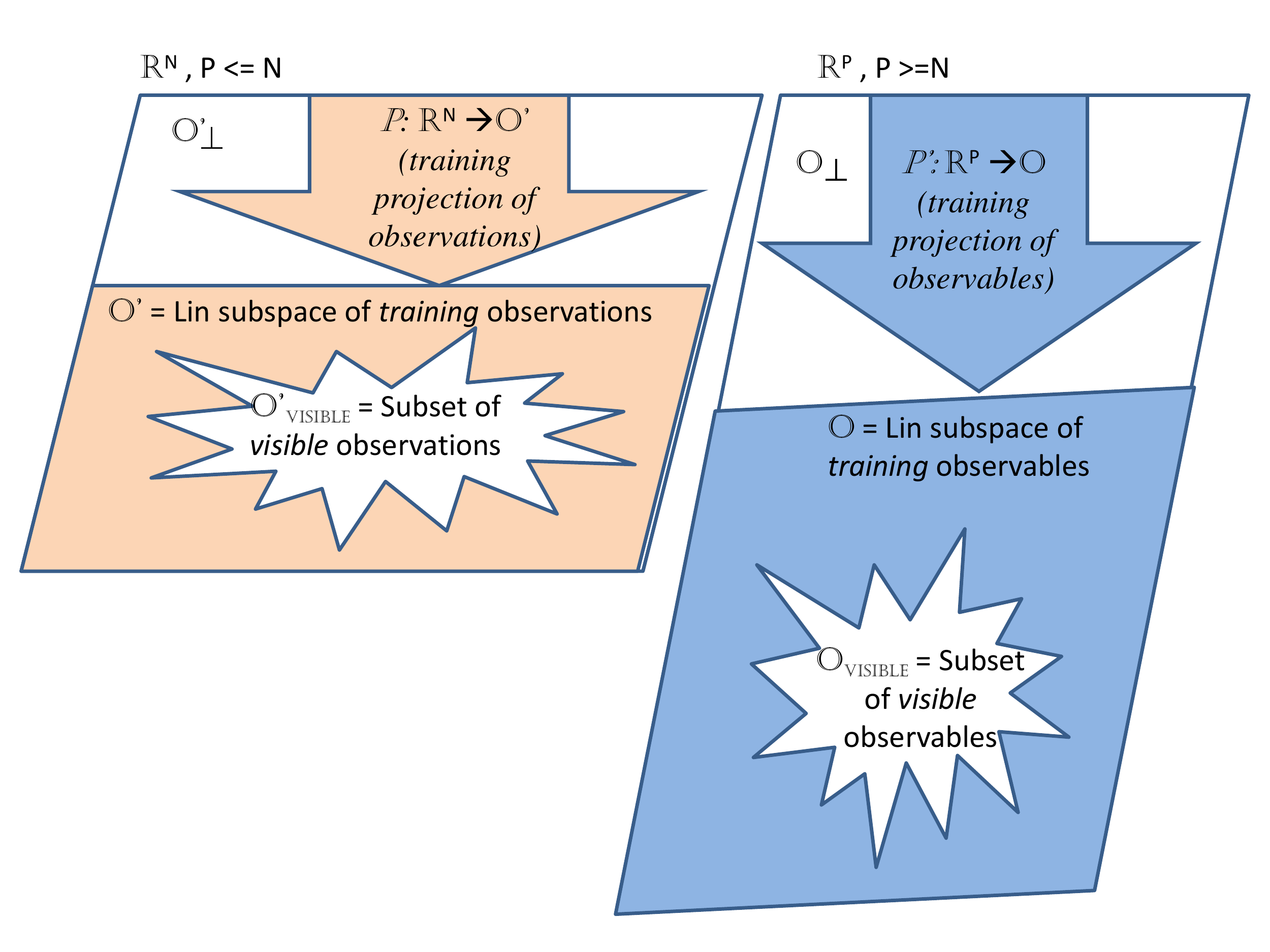}}
\caption{ In the framework of Figure \ref{Fig3.-21}, on the right is the  typical case $P \geq N$ and the training decomposition  of the observables space $\mathbb{R}^P = \mathbb{O} \oplus \mathbb{O}_{\perp}$  as in (\ref{0.-8}). On the left is the case $P \leq N$ and the training decomposition of the space  of observations  $\mathbb{R}^N = \mathbb{O'} \oplus \mathbb{O'}_{\perp}$ as in (\ref{0.-22}). }
\label{Fig3.-25}
\end{center}
\vskip -0.2in
\end{figure}

\begin{figure}
\vskip 0.2in
\begin{center}
\centerline{\includegraphics[width=\columnwidth]{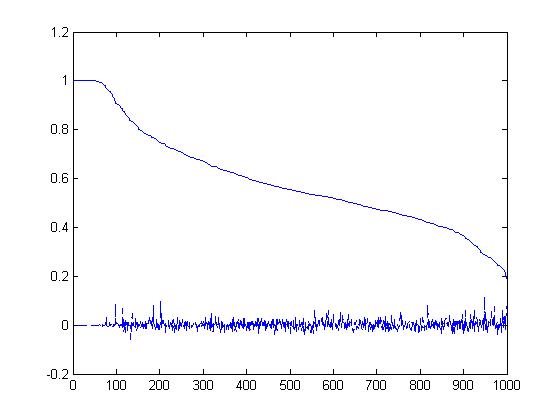}}
\vskip 0.2in
\centerline{\includegraphics[width=\columnwidth]{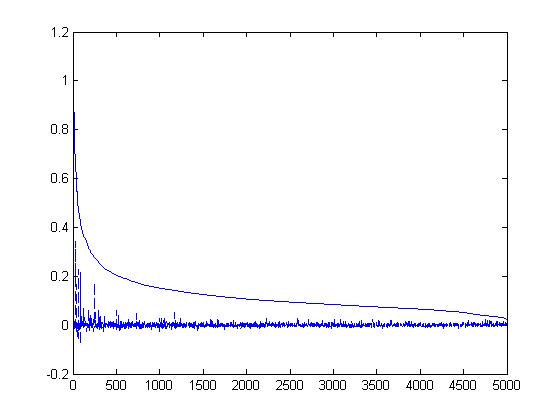}}
\caption{The  diagonal (solid line) $P'_{\mu \mu}$ and first off-diagonal (dashed line) elements $P'_{\mu, \mu+1}$  of the training projection matrix  $\mathbf{ P'}$   for $P = 1,000$ (top) and $P=5,000$ (bottom), sorted in descending order. We used respectively the first 1,000  MNIST images (top) and first 5,000 MNIST images (bottom). The residual sum of squares $RSS$ $=\sum_{\mu}  (1- P'_{\mu \mu})$ is 1 minus the solid line and clearly increases  as $P$ increases, for a fixed $N$. The bottom chart differs from the bottom chart in Figure \ref{Fig3.-26} because here the dimension was reduced in the space of observables from $N=784$ to $M=600$.}
\label{Fig3.-27}
\end{center}
\vskip-0.2in
\end{figure}

The orthogonality of the training projection (property iii) in (\ref{0.-6})), allows us to split the space of observables $\mathbb{R}^P$ into a direct sum of the training space $\mathbb O$ $=\mathbb {P'}\mathbb{R}^{P}$ and its orthogonal $\mathbb{O}_{\perp}$  = $(I_{P} - \mathbb {P'})\mathbb{R}^{P}$:
\begin{align}
\mathbb{R}^P = \mathbb{O} \oplus \mathbb{O}_{\perp} =  \mathbb {P'}\mathbb{R}^{P} \oplus (I_{P} - \mathbb {P'})\mathbb{R}^{P}.
\label{0.-8}
\end{align}
In the framework of Figure \ref{Fig3.-21}, we visualize on the right side of Figure \ref{Fig3.-25} this \emph{training decomposition} (cmp. against Figure \ref{Fig3.-23}). 

For the basis vectors $\{ \mathbf{e}_{\mu}\}_{{\mu}=1}^P$ introduced in (\ref{0.-2a}), this orthogonal decomposition becomes:
\begin{align}
\mathbf{e}_{\mu} =  \mathbb {P'}\mathbf{e}_{\mu} + \boldsymbol{\varepsilon}_{\mu},
\label{0.-9}
\end{align}
where $\boldsymbol{\varepsilon}_{\mu}$ $\in \mathbb{O}_{\perp}$ can be thought of as a residual, similarly to a linear regression model as in footnote \ref{MultLinReg}. Due to the orthogonality (\ref{0.-8}), the \emph{Pythagorean theorem} and the projection property (\ref{0.-7.1}) now imply:
\begin{align}
1 = ||\mathbf{e}_{\mu}||_{\mathbb E}^2 =  ||\mathbb {P'}\mathbf{e}_{\mu}||_{\mathbb E}^2 + ||\boldsymbol{\varepsilon}_{\mu}||_{\mathbb E}^2 =  P'_{\mu \mu} + ||\boldsymbol{\varepsilon}_{\mu}||_{\mathbb E}^2 .
\label{0.-10}
\end{align}
In particular $ P'_{\mu \mu} $ $= \mathbf{x}_{\mu} \mathbf{G}^{-1}\mathbf{x}_{\mu}^T$ $\leq 1$ and equality is reached only if the basis vector $\mathbf{e}_{\mu}$ is in the training space $\mathbb O$. In analogy with regressions, one can call the sum 
\begin{align}
RSS = \sum_{\mu} ||\boldsymbol{\varepsilon}_{\mu=1}^P||_{\mathbb E}^2 = \sum_{\mu=1}^P (1- P'_{\mu \mu})
\label{0.-11}
\end{align}
residual sum of squares (RSS). In a typical dataset, it is likely to increase as $P$ increases for a fixed $N$ - see Figure \ref{Fig3.-27}.

Switching for completeness to the case $P \leq N$, there is an obvious analogue of the training projection, using the right conjugate matrix $\mathbf{\check{X'}}$ from (\ref{0.1.1b.1a}): the projection matrix:
\begin{align}
 \mathbf{P} :=\mathbf{X}^T\mathbf{\check{X'}} =\mathbf{X}^T \mathbf{G'}^{-1}\mathbf{X} =\mathbf{X}^T\mathbf{ (X X}^T)^{-1}\mathbf{X}
 \label{0.-20}
 \end{align}
 induces the projection  $\mathbb {P}$:
\begin{align}
\mathbb{P} &: \mathbb{R}^N  \rightarrow \mathbb{O'} , \nonumber \\
\mathbb{P}\mathbf{q'} & := \mathbf{ q' P}  =  \mathbf{q} \mathbf{X}^T \mathbf{G'}^{-1}\mathbf{X}.
\label{0.-21}
\end{align}
 with the respective direct sum decomposition of the space of observations:
\begin{align}
\mathbb{R}^N = \mathbb{O'} \oplus \mathbb{O'}_{\perp},
\label{0.-22}
\end{align}
where $\mathbb{O'}_{\perp}$  = $(I_{N} - \mathbb {P})\mathbb{R}^{N}$.

\subsection{Training metrics and their conjugates.}
\label{Training metrics}

The plain Euclidean metric for observables $<\mathbf {p,q}>_{\mathbb E}$ $=\mathbf {p^T q}$  in the image space $\mathbb R^P$ of $\mathbb T$, induces  a  new metric in domain space $\mathbb R ^N$ of $\mathbb T$ : for two arbitrary observations $\mathbf{p'}$  , $\mathbf{q'}  \in \mathbb R^N$ :
\begin{align}
<\mathbf{ p',q'}>_{\mathbb T} &:= <\mathbb{T}\mathbf{q'}, \mathbb{T}\mathbf{q'}>_{\mathbb E} = (\mathbb{T}\mathbf{q'})^T (\mathbb{T}\mathbf{q'})  = \mathbf{p' G q'}^T , 
\label{0.1.2a}
\end{align}
where $\mathbf G$ is the  Gram matrix of the training observables (\ref{0.1a}). We will refer to (\ref{0.1.2a}) as the $\mathbb T$-\emph{training inner product} (or simply \emph{training metric}) for observations. By the definition (\ref{0.1.2a}), the training inner product is the plain Euclidean product of overlaps:
\begin{align}
<\mathbf{ p',q'}>_{\mathbb T} &=  \sum_{\mu=1}^P (\mathbb{T}\mathbf{p'})_{\mu} (\mathbb{T}\mathbf{q'})_{\mu}, 
\label{0.1.2c.0}
\end{align}
where the right-hand sight is the familiar \emph{Hebbian} metric for the  two observables $ \{p_{\mu} \}$, $ \{q_{\mu} \} \in \mathbb O$ , still widely used for learning algorithms in Neural Networks.  By mapping arbitrary observations $\mathbf{p'}$ , $\mathbf{q'} $ $ \in \mathbb R^N$ into the respective observables  via $\mathbb T$, we in other words ``flattened'' the highly non-trivial metric $\mathbf{p' G q'^T}$  into a plain Euclidean metric.

 The training inner product  of two training observations $<\mathbf{x_{\mu},x_{\nu}}>_{\mathbb T}$  is  from (\ref{0.1.2c.1}), (\ref{0.1.2a}), (\ref{0.1.2c.0}):
\begin{align}
<\mathbf{x_{\mu},x_{\nu}}>_{\mathbb {T}}  &:= \mathbf{x_{\mu}} \mathbf{G} \mathbf{x}^T_{\nu} = \sum_{\kappa=1}^P (\mathbb{T}\mathbf{x_{\mu}})_{\kappa} (\mathbb{T}\mathbf{x_{\mu}})_{\kappa} \nonumber \\
&=\sum_{\kappa=1}^P G'_{\mu \kappa} G'_{\nu \kappa} = (\mathbf{G'^2)_{\mu \nu}}.
\label{0.1.2c.3}
\end{align}
In particular, the squared \emph{training norm} $||\mathbf{x_{\mu}}||_{\mathbb T}^2$ $=\mathbf{x_{\mu}} \mathbf{G} \mathbf{x}^T_{\mu}$ is given by the respective diagonal element of $\mathbf {G'^2}$:
\begin{align}
||\mathbf{x_{\mu}}||_{\mathbb {T}}^2:= <\mathbf{x_{\mu},x_{\mu}}>_{\mathbb T}   =\mathbf{x_{\mu}} \mathbf{G} \mathbf{x}^T_{\mu} = \nonumber \\ = (\mathbf{G'^2)_{\mu \mu}} = \sum_{\kappa=1}^P (G'_{\mu \kappa})^2,
\label{0.1.2c.4}
\end{align}
and measures the sum of the squared overlaps, hence the proximity, of this particular training observation $\mathbf x_{\mu}$ with all other training observations. 

\begin{figure}[!ht]
\vskip 0.2in
\begin{center}
\includegraphics[width=\columnwidth]{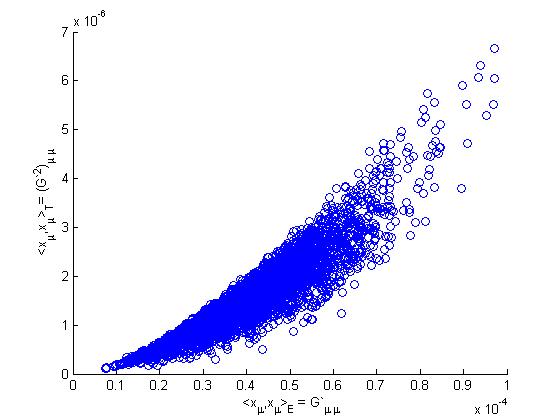}
\caption{ Scatter plot of the squared training norm $||\mathbf{x_{\mu}}||^2_{\mathbb T} $ $=(\mathbf{G'^2)_{\mu \mu}}$ (vertical dimension) against the squared Euclidean norm $ ||\mathbf{x_{\mu}}||^2_{\mathbb E} $ $=G'_{\mu \mu}$ (horizontal dimension), using the first 5,000 MNIST images ($P = 5,000; N= 784$).}
\label{Fig1.1}
\end{center}
\vskip-0.2in
\end{figure}

\begin{figure}[!ht]
\vskip 0.2in
\begin{center}
\includegraphics[width=\columnwidth]{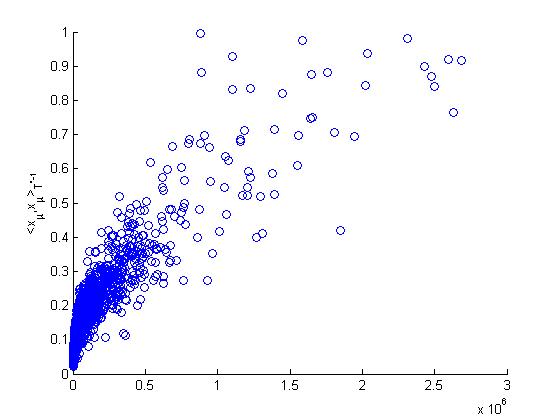}
\caption{ The equivalent of Figure \ref{Fig1.1} but for conjugates: the scatter plot of the squared conjugate training norm $ ||\mathbf{x_{\mu}}||^2_{\mathbb T'^{-1}} $ $=P'_{\mu \mu}$ $=||\mathbf{\check{x}_{\mu}}||^2_{\mathbb T} $ (vertical dimension) against the squared Euclidean norm of the conjugate observations $ ||\mathbf{\check{x}_{\mu}}||^2_{\mathbb E} $ (horizontal dimension), using the first 5,000 MNIST images (i.e., $P = 5,000$). Observables dimension was reduced from 784 to 600 i.e. $N=600$.}
\label{Fig1.1b}
\end{center}
\vskip-0.2in
\end{figure}

To visualize  the difference between the the training norm  $<,>_{\mathbb T}$ and the standard Euclidean norm $<,>_{\mathbb E}$ , we plot on Figure \ref{Fig1.1}, for  the first 5,000 MNIST observations,  the scatter plot of their squared training norm   $<\mathbf{x_{\mu}} , \mathbf{x}_{\mu} >_{\mathbb T}$  against their respective squared Euclidean norm $<\mathbf{x_{\mu}}, \mathbf{x}_{\mu}>_{\mathbb E} $ $=G'_{\mu \mu}$,  $\mu = 1,...,5000$. The two metrics are  different, albeit correlated. Similarly, the scatter plot between the squared Euclidean and squared training norm, but for invisible conjugate observations, is on Figure \ref{Fig1.1b}. One sees a again a very high correlation between the two metrics.

Switching to observables, in analogy with (\ref{0.1.2a}), the Euclidean metric in the image space of $\mathbb T'$ induces again a training metric in the domain space $\mathbb{R}^P$ of $\mathbb T'$:  for any two observables $\mathbf{p,q}$ $ \in \mathbb{R}^P$, one has:
\begin{align}
<\mathbf{ p,q}>_{\mathbb T'} &:= <\mathbb{T'}\mathbf{p}, \mathbb{T'}\mathbf{q}>_{\mathbb E} = (\mathbb{T'}\mathbf{p})^T (\mathbb{T'}\mathbf{q})  = \mathbf { p}^T \mathbf{G' q} , 
\label{0.1.2b}
\end{align}
where $\mathbf G'$ is the Gram matrix of  training observations (\ref{0.1.1b}). 

Equally important to the training metrics $<\mathbf{p',q'}>_{\mathbb{T}}$  and $<\mathbf{p,q}>_{\mathbb{T'}}$ are their  ``conjugate'' counterparts. The conjugates training metrics are simply the training metrics introduced above, but for the conjugates $<\mathbf{\check{p'},\check{q'}}>_{\mathbb{T}}$  and $<\mathbf{\check{p},\check{q}}>_{\mathbb{T'}}$, which were introduced in (\ref{0.0d}), (\ref{0.0d.1}).  The \emph{conjugate training metric} of  observations $\mathbf {p',q'} \in \mathbb R^N$ is defined:
 \begin{align}
<\mathbf{ p',q'}>_{\mathbb {T'}^{-1}} &:=    <\mathbf{ \check{p'},\check{q'}}>_{\mathbb T} = \mathbf{\check{p'} G \check{q'}}^T= \mathbf{p' G^{-1} q'}^T , 
\label{0.1.0.1a}
\end{align}
where the conjugate observation $\mathbf{\check{q'}}$ was defined in (\ref{0.0d}) (and assuming that $\mathbf{G}^{-1}$ is well-defined). The subscript $\mathbb{T'}^{-1}$ is used because it can be thought of as induced by the Euclidean metric for observables in the domain space $\mathbb R^P$ of $\mathbb{T'}$:
\begin{align}
<\mathbf{ p',q'}>_{\mathbb T'^{-1}} &=  <\mathbb{T'}^{-1}\mathbf{p'}, \mathbb{T'}^{-1}\mathbf{q'}>_{\mathbb E}.
\label{0.1.0.1}
\end{align}

There is an obvious connection with the training projection matrix $\mathbf{P'}$: the conjugate training inner product  of two training observations $<\mathbf{x_{\mu},x_{\nu}}>_{\mathbb T^{-1}}$ is from (\ref{0.-5.-1})  the corresponding matrix element $P'_{\mu \nu}$ of $\mathbf{P'}$ and the squared conjugate training norm is:
\begin{align}
||\mathbf{x_{\mu}}||_{\mathbb {T'}^{-1}}^2:= <\mathbf{x_{\mu},x_{\mu}}>_{\mathbb {T}^{-1}}  = \mathcal{L_{\mathcal G} (\mathbf{x}_{\mu})} = \nonumber \\
=  \mathbf{x_{\mu}} \mathbf{G}^{-1} \mathbf{x}^T_{\mu} = ||\mathbb{P'}\mathbf{e}_{\mu} ||_{\mathbb E}^2 = P'_{\mu \mu}  = ||\mathbf{\check{x}_{\mu}}||_{\mathbb {T}}^2,
\label{0.1.2c.10}
\end{align}
(cf. (\ref{0.-7.1})). Unlike the training norm (\ref{0.1.2c.4}), there is no matrix squaring here, due to the projection property $\mathbf{P' }^2 = \mathbf{P'}$. Similarly to the training norm,  the conjugate training norm $||\mathbf{x_{\mu}}||_{\mathbb {T'}^{-1}}^2$ measures the sum of the squared overlaps, hence the proximity, of the conjugate training observation $\mathbf {\check{x}}_{\mu}$ with all other training observations. 

\begin{figure}[!ht]
\vskip 0.2in
\begin{center}
\includegraphics[width=\columnwidth]{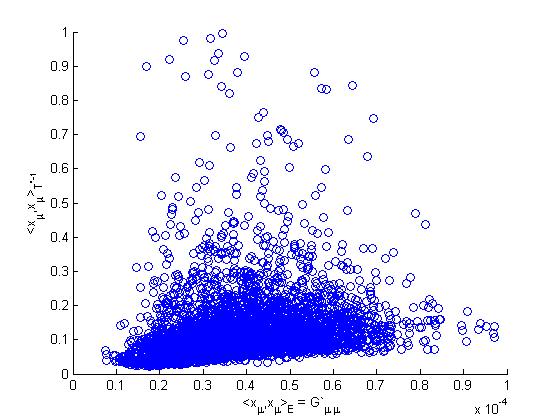}
\caption{ Scatter plot of the squared conjugate training norm $ ||\mathbf{x_{\mu}}||^2_{\mathbb T'^{-1}} $ $=P'_{\mu \mu}$ (vertical dimension) against the squared Euclidean norm $ ||\mathbf{x_{\mu}}||^2_{\mathbb E} $ $=G'_{\mu \mu}$ (horizontal  dimension), using the first 5,000 MNIST images (i.e., $P = 5,000$). Observables dimension was reduced from 784 to 600 i.e. $N=600$.}
\label{Fig1.1a}
\end{center}
\vskip-0.2in
\end{figure}

The squared conjugate training norms  for  the  MNIST dataset are plotted on Figure \ref{Fig3.-27} for two different number of observations $P$. We also plot on Figure \ref{Fig1.1a}  the scatter plot of their squared conjugate training norm   $||\mathbf{x_{\mu}}||^2_{\mathbb T'^{-1}}$ $=P'_{\mu \mu}$  against the respective squared Euclidean norm $||\mathbf{x_{\mu}}||^2_{\mathbb E} $ $=G'_{\mu \mu}$,  $\mu = 1,...,5000$. The two metrics are completely different! It turns out, the conjugate training metric is highly correlated with another Euclidean metric:  not of the original observations, but of their conjugates, as seen in Figure \ref{Fig1.1b}.

Switching again to observables,  for any two arbitrary observables $\mathbf{p} $ , $\mathbf{q}  \in \mathbb R^P$:
  \begin{align}
<\mathbf{ p,q}>_{\mathbb T^{-1}} &:=  <\mathbf{ \check{p},\check{q}}>_{\mathbb T} =   \mathbf{\check{p}}^T \mathbf{G' \check{q}} = \mathbf{p}^T \mathbf{G'^{-1} q} , 
\label{0.1.1b.3}
\end{align}
which is equivalent to:
\begin{align}
<\mathbf{ p,q}>_{\mathbb T^{-1}} &:=  <\mathbb{T}^{-1}\mathbf{p}, \mathbb{T}^{-1}\mathbf{q}>_{\mathbb E}  . 
\label{0.1.2a.1}
\end{align}

%

\subsection{Graph-theoretical view.}
\label{Graph-theoretical view.}

From  graph-theoretical point of view, we will think of training observations as the vertices of the \emph{ training graph} and will consider two training observations connected if the overlap $(\mathbf{x}_{\mu})_{\kappa} = (\mathbf{x}_{\kappa})_{\mu}$ $=G'_{\mu \nu}$  is greater than a pre-defined threshold. One can  think of the square of the overlap  $(G'_{\mu \nu})^2$  as being proportional to the number of edges between the two vertices $\mu$ and $\nu$ and hence the matrix $\mathbf G'$ can be thought of as the \emph{adjacency matrix} of the training graph. If one goes one step further and introduces ``transition probability'' between vertices in the quantum probability sense, the square of the overlap between two vertices will be proportional to the probability of ``direct'' transition between two vertices. Obviously, the one-step direct transition is not in general the most probable path between two vertices and there will be a miriad of more probable multi-step paths connecting them. 

Moreover, since $\mathbf G'^2$ is exactly the Gram matrix of the overlaps, the training norm (\ref{0.1.2c.4}) is the \emph{ vertex degree} or \emph{ vertex valency} of the vertex $\mu$ i.e. it is proportional to the ``number of edges'' incident to the vertex. In particular, the summand  $(G'_{\mu \mu})^2$ in (\ref{0.1.2c.4}) for $\kappa = \mu$ corresponds to the self-loop of vertex $\mu$. In order to measure the true inter-connectedness of vertex $\mu$, we need to compare our graph against a graph consisting of $N$ identical vertices $\mu$, each  with self-loops $(G'_{\mu \mu})^2$.  

\subsection{Summary of metrics.}
\label{Summary of}

Let us finally summarize the different metrics introduced so far for observations and observables. For arbitrary observations $\mathbf{p',q'} \in \mathbb R^N$ and observables $\mathbf{p,q} \in \mathbb R^P$, we have:

i) the plain Euclidean metric
\begin{align}
 \mathbf {<p',q'>_{\mathbb E} } &= \mathbf {p' q'}^T  ,\nonumber \\ 
  \mathbf {<p,q>_{\mathbb E} } & = \mathbf {p}^T \mathbf{q}.
\label{0.1.10}
\end{align}

ii) the training metric (\ref{0.1.2a}) for observations and (\ref{0.1.2b}) for observables:
\begin{align}
 \mathbf {<p',q'>_{\mathbb T} } &=\mathbf {<\mathbb{T}p',\mathbb{T}q'>_{\mathbb E} } =  \mathbf {p'G q'}^T  , \nonumber \\
  \mathbf {<p,q>_{\mathbb T'} } & =\mathbf {<\mathbb{T'}p,\mathbb{T'}q>_{\mathbb E} } =  \mathbf {p}^T\mathbf{G' q}.
\label{0.1.11}
 \end{align}

iii) the conjugate training metric  (\ref{0.1.0.1}) for observations and (\ref{0.1.2a.1}) for observables:
\begin{align}
 \mathbf {<p',q'>}_{\mathbb {T'}^{-1}}  &=\mathbf {<p'\mathbb{T'}^{-1},q'\mathbb{T'}^{-1}>_{\mathbb E} } =  \mathbf {p'G^{-1} q'}^T  , \nonumber \\
  \mathbf {<p,q>}_{\mathbb {T}^{-1}}  & =\mathbf {<\mathbb{T}^{-1}p,\mathbb{T}^{-1}q>_{\mathbb E} } =  \mathbf {p}^T \mathbf{G'^{-1} q}.
\label{0.1.12}
\end{align}
Let us stress that the conjugate training metric for observations (resp. observables) is applicable only when $\mathbf{G}^{-1}$ (resp. $\mathbf{G'}^{-1}$) exist.

\section{Training probability distributions.}
\label{Training probability}

Let us now look at the training set from probability-theoretical point of view.
 
\subsection{Eigen-observations.}
\label{Eigen-observations}

From probabilistic point view, one is always better off if some form of factorization can be achieved i.e. deal with random variables which are  independent. Independence typically implies orthogonality in some natural metric. In the context of the linear algebraic picture of  training set developed in Section \ref{The training set}, we can therefore first address the simpler problem of orthogonal zing observations and observables. 

\begin{figure}[!ht]
\vskip 0.2in
\begin{center}
\includegraphics[width=\columnwidth]{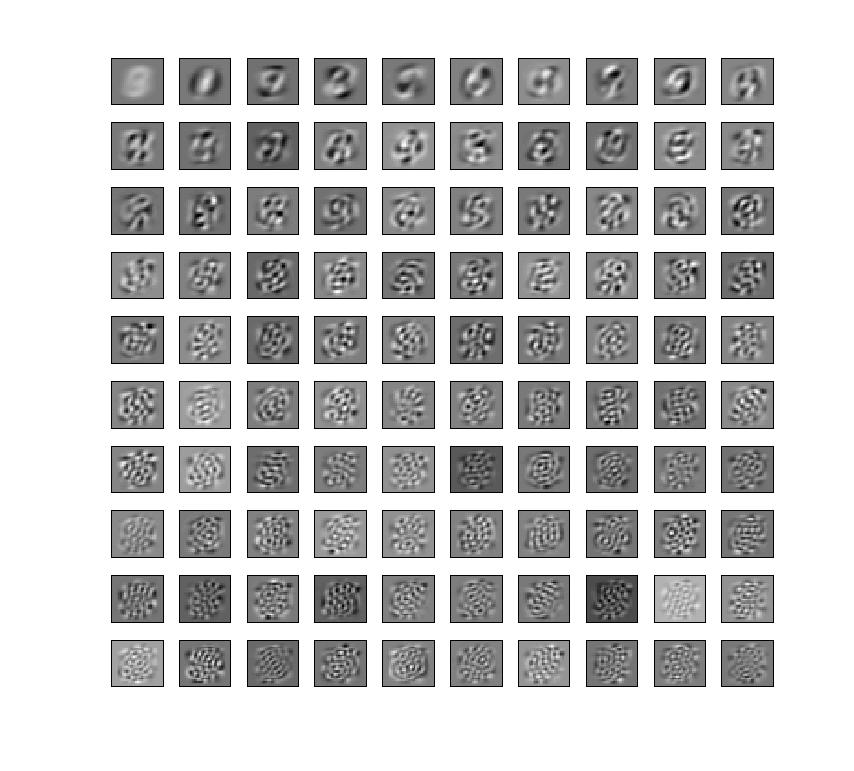}
\caption{Top one hundred eigen-observations $\{\mathbf{w}_i\}_{i=1}^{10},$ in decreasing order of the respective eigen-values, using the first 5,000 MNIST images (i.e., $P = 5,000$). }
\label{Fig1.2.0}
\end{center}
\vskip -0.2in
\end{figure}

\begin{figure}[!ht]
\vskip 0.2in
\begin{center}
\includegraphics[width=\columnwidth]{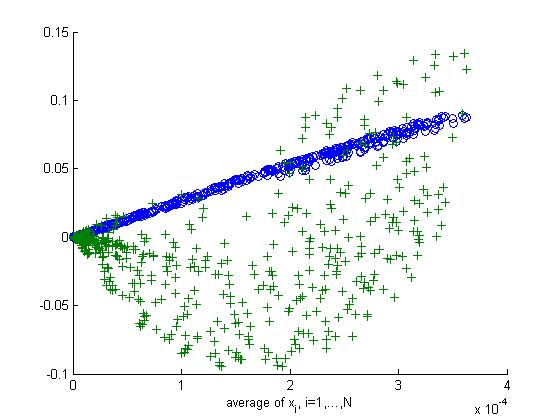}
\caption{Scatter plot of the values of  the first eigen-observation $\mathbf{w}_1$ (circles) and the second eigen-observation $\mathbf{w}_2$ (crosses)  against  $\{\bar{\mathbf{x}}_i \}_{i=1}^N$, where $\bar{\mathbf{x}}_i= \sum_{\mu=1}^P X_{\mu i} $, using the first 5,000 MNIST images (i.e., $P = 5,000$). The first eigen-observation is virtually identical to the average value of the respective pixel, which is not the case if the data is de-meaned.}
\label{Fig1.2.1}
\end{center}
\vskip -0.2in
\end{figure}

Training observables and observations are in general highly correlated, so we need to consider  in the spirit of (\ref{0.-1}) \emph{hidden} variables  which are linear combinations of training observations i.e. points in the training space $\mathbb O'$ which  are not in the training set $\mathbb{O'}_{visible}$. Natural candidates are the orthonormal eigen-vectors $\{\mathbf{w}_i\}_{i=1}^M$  of the  training Gram matrix $\mathbf G$ (cf. (\ref{0.1a})) which are covariant column-vectors in $ \mathbb{R} ^N$. They are in the training space $\mathbb O'$ (cf. (\ref{1.1.6})) but typically hidden  and we refer to them, as is common,  as \emph{eigen-observations}. The eigen-observations satisfy:
\begin{align}
\mathbf  {G w}_i= \lambda_i^2 \mathbf{w}_i , \quad i=1,...,M. 
\label{1.1.1}
\end{align}
They are orthogonal in the training metrics and orthonormal in Euclidean metrics,
\begin{align}
<\mathbf{w}^T_i,\mathbf{w}^T_j>_{\mathbb T}& =  \mathbf{w}^T_i\mathbf{G} \mathbf{w}_j = \lambda_i^2 \delta_{i j}   , 
\label{1.1.2} \\
<\mathbf{w}^T_i,\mathbf{w}^T_j>_{\mathbb E} &=  \mathbf{w}^T_i\mathbf{w}_j = \delta_{i j}.
\label{1.1.2a}
\end{align}
We will order the eigen-observations  $\{\mathbf{w}_i\}_{i=1}^M$ in decreasing order of the respective \emph{eigen-values} $\{\lambda_i\}_{i=1}^M$. 
For the first 5,000 MNIST images, the top one hundred eigen-observations in this order are plotted on Figure \ref{Fig1.2.0}. The scatter plot of the top two eigen-observations against the averaged pixel intensities $\{\bar{\mathbf{x}}_i \}_{i=1}^N$  are shown in Figure \ref{Fig1.2.1}. 
 
\begin{figure}[!ht]
\vskip 0.2in
\begin{center}
\includegraphics[width=\columnwidth]{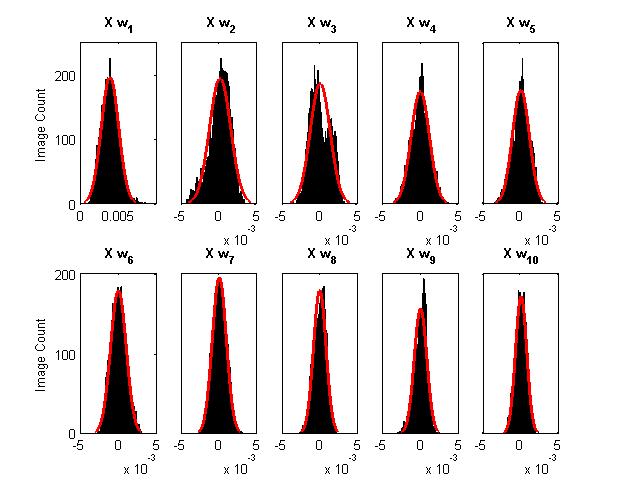}
\caption{Probability distributions of the top ten eigen-mappings $\{\mathbb{T}\mathbf{ w}^T_i\}_{i=1}^{10}$ $=\{\mathbf{X w}_i\}_{i=1}^{10}$, using the first 5,000 MNIST images (i.e., $P = 5,000$). The aggregate image count for each of the 10 histograms is hence 5,000 . These are also the histograms of the top ten eigen-observables $\{\mathbf{v}_i\}_{i=1}^{10}$ (cf. (\ref{1.1.6.1.-1})).}
\label{Fig1.2.2}
\end{center}
\vskip -0.2in
\end{figure}

\begin{figure}[!ht]
\vskip 0.2in
\begin{center}
\includegraphics[width=\columnwidth]{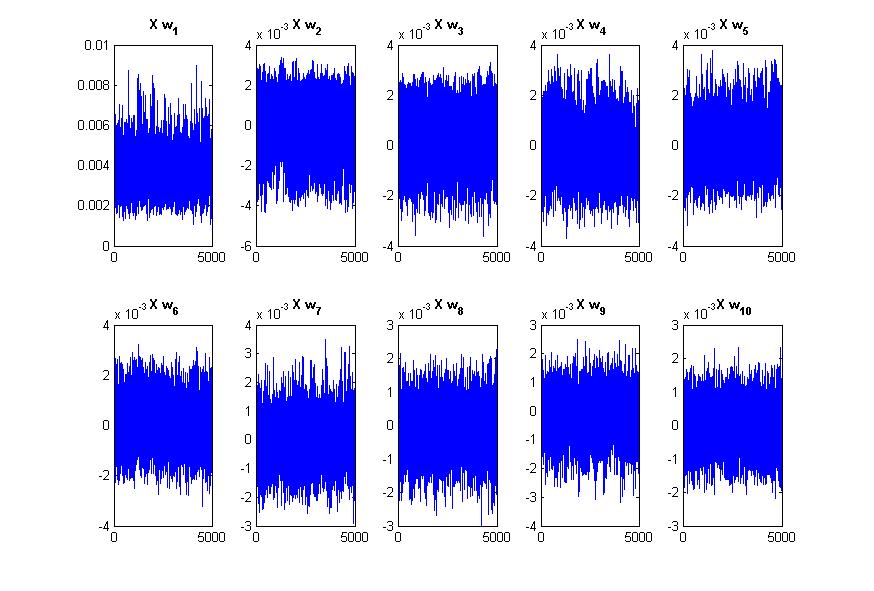}
\caption{Time series plots of the top ten eigen-mappings $\{\mathbb{T}\mathbf{ w}^T_i\}_{i=1}^{10}$ ($P = 5,000$).}
\label{Fig1.2.2a}
\end{center}
\vskip -0.2in
\end{figure}

We will refer to the training mappings $\mathbb{T} \mathbf{w}_i^T$ $= \mathbf {Xw}_i$ of eigen-observations (cf. (\ref{0.0a})) as \emph{eigen-mappings}. We plot on Figure \ref{Fig1.2.2} the histogram and  on  Figure \ref{Fig1.2.2a} the quantile-quantile plots of top ten eigen-mappings $\{\mathbf{X w}_i\}_{i=1}^{10}$ for the first 5,000 MNIST images.
 
Due to (\ref{0.1.0.1b}), and the definition (\ref{1.1.1}), eigen-observations with non-zero eigenvalues equal, up to a scaling factor, their conjugates:
\begin{align}
\mathbf  {\check{w}}_i= \frac{1}{ \lambda_i^2}\mathbf{w}_i, \quad i=1,...,M, 
\label{1.1.1a}
\end{align}
see (\ref{1.1.6.1.-1}) and (\ref{1.1.6.1.-2}) for more color on that.

\subsection{Distributions of Eigen-Mappings.}
\label{Mapping Pursuit}
\begin{figure}[!ht]
\vskip 0.2in
\begin{center}
\includegraphics[width=\columnwidth]{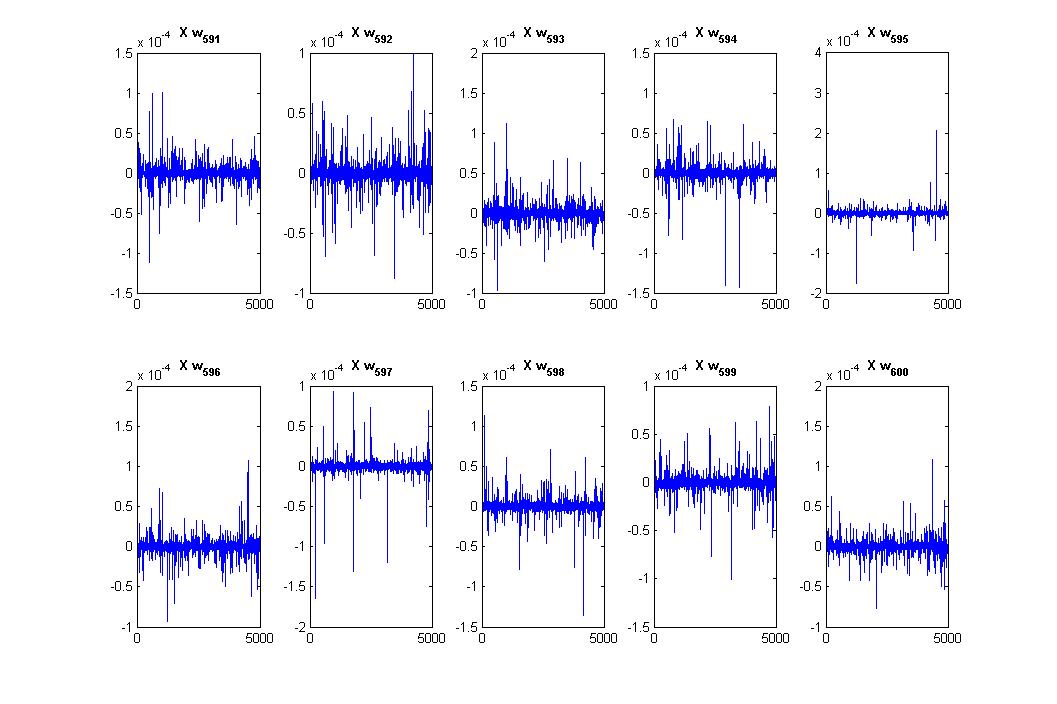}
\caption{Time series plots of the bottom ten eigen-mappings $\{\mathbb{T}\mathbf{ w}^T_i\}_{i=591}^{600}$ ($P = 5,000$). In order to remove unimportant noise, we have removed the smallest eigenvectors and have retained only the first $n=600$ of them (out of $N=784).$}
\label{Fig1.2.2b}
\end{center}
\vskip -0.2in
\end{figure}

\begin{figure}[!ht]
\vskip 0.2in
\begin{center}
\includegraphics[width=\columnwidth]{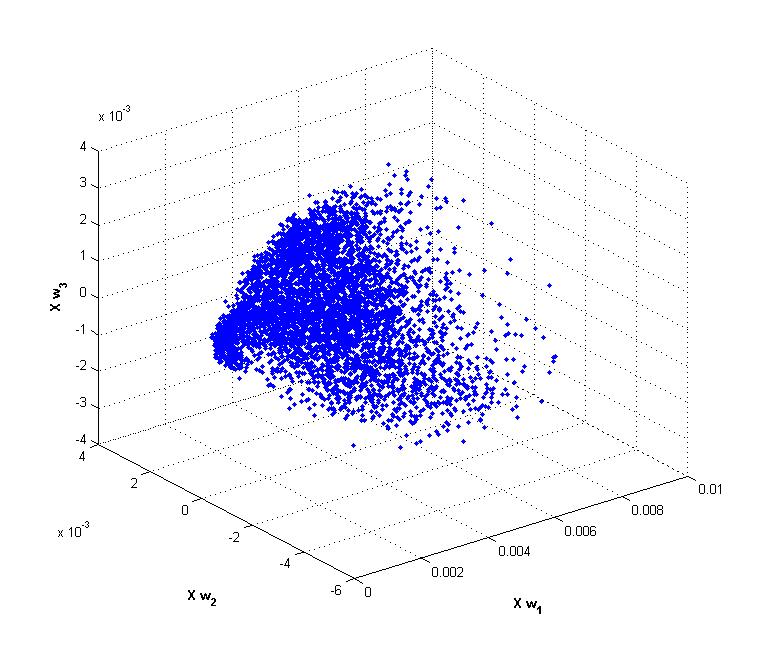}
\caption{Scatter plot of the top three eigen-mappings $\mathbb{T}\mathbf{w}^T_1$ , $\mathbb{T}\mathbf{w}^T_2$ , $\mathbb{T}\mathbf{w}^T_3$ , using the first 5,000 MNIST images (i.e., $P = 5,000$).}
\label{Fig1.2.3}
\end{center}
\vskip -0.2in
\end{figure}

 One has to ask to what extent training mappings on the eigen-observations capture important structures of the dataset. In the MNIST dataset e.g, there are ten distinguished clusters, namely the ten digits, and one would want to find one- or higher-dimensional mappings along which the ten clusters are visibly separated. Unfortunately, as Figure \ref{Fig1.2.2} shows, except for the 3rd eigen-mapping, there is no sign of multi-modal distributions which would allow us to separate the main clusters in the dataset. Even worse, except for the 3rd eigen-mapping which is bi-modal, the rest of the eigen-mappings are uni-modal and very close to Gaussian - cf. Figure \ref{Fig1.2.2a} to confirm that they do look like ``noise''. This is in stark contrast with the time series plot of the bottom ten eigen-mappings on Figure \ref{Fig1.2.2b}, which are visibly spiky and hence, highly non-Gaussian. Since the clusters may be ``squashed'' and invisible in one-dimensional mappings, one could look for them in higher-dimensional eigen-mappings space. We plot the 3-dim scatter plot of the top three eigen-mappings on Figure \ref{Fig1.2.3}. Again, there is no sign of any clustering.
 
\subsection{Singular value decomposition.} 
 \label{Singular value}
 
 We will assume  for simplicity of presentation that the  eigenvalues $\{\lambda^2_i\}$ of $\mathbf G$ are different and $\mathbf G$ is of full rank $M = N <= P$ (this assumption is not critical). The eigen-observations are  the columns $\{ \mathbf{w}_i\}_{i=1}^M$ of the orthonormal matrix $\mathbf W$ in the \emph{singular value decomposition} of $\mathbf X$:
\begin{equation}
\mathbf{X = V \Lambda W^T}
\label{1.1.3}
\end{equation}
where $\mathbf V$ is $P\times P$ \emph{left singular } orthogonal matrix ($\mathbf {V V^T = V^T V = I}_P$) , $\boldsymbol \Lambda$  is $P\times N$ diagonal matrix, with diagonal elements - the eigenvalues $ \{\lambda_i \}$, and $\mathbf W$ is $N \times N$ orthonormal matrix ($\mathbf {W W^T = W^T W = I}_N$), using the notation $\mathbf {I}$ for the identity matrix in the respective dimension.  The matrix $\mathbf W$ is the same matrix as in (\ref{2.0.6}). 

One can now easily check that the eigen-observations are in the training observation space $\mathbb O'$, albeit hidden in the general case (cf. Section \ref{The training set}): Using the  orthogonality of $\mathbf V$, it follows from (\ref{1.1.3}) that:
\begin{equation}
W_{ji}  = W^T_{ij} = \sum_{\mu} \lambda_i^{-1} V^T_{i\mu} X_{\mu j},
\label{1.1.5}
\end{equation}
and hence
\begin{align}
\mathbf{w}^T_{i} = \sum_{\mu} \left(\lambda_i^{-1} V_{\mu i}\right) \mathbf{x}_{\mu} .
\label{1.1.6}
\end{align}

The singular value decomposition of $\mathbf X$  has as an obvious by-product the singular value decomposition of $\mathbf {G, G'}$: 
\begin{align}
\mathbf {G} =\mathbf{X^T X } = \mathbf{W}\mathbf{ \Lambda}^2 \mathbf{W}^T, \nonumber \\
\mathbf {G'} =\mathbf{X X^T } = \mathbf{V}\mathbf{ \Lambda}^2 \mathbf{V}^T,
\label{1.1.4}
\end{align}
where $\mathbf{\Lambda}^2$ is a diagonal matrix with positive diagonal elements the squared eigenvalues  $\{\lambda^2_i\}$ $\in \mathbb R$.

\subsection{Whitening.} 
\label{Whitening}

In  typical real-life datasets, one can add more observations while number of observables is fixed i.e. $P \gg N$ and let us assume again for simplicity that $rank(\mathbf{X}) = M =N$. In that case, one can consider instead of the left singular matrix $\mathbf V$ its  truncated $P \times N$ sub-matrix, comprised of the first $N$ columns (abusing notations, we will continue to use the same letter $\mathbf V$ but will refer to it as \emph{truncated} left singular matrix).  The truncated $\mathbf V$ is orthogonal in one direction only i.e. $\mathbf {V^T V = I}_N$ but $\mathbf {V V^T \neq I}_P$. In fact, one can easily check from the definitions (\ref{0.-5.-2}), (\ref{1.1.3}), (\ref{1.1.4}) that the Gram matrix $\mathbf {V V}^T$ is exactly the training projection matrix $\mathbf P'$:
\begin{align}
\mathbf {V V}^T = \mathbf{P}', \quad (\mathbf {V V}^T)_{\mu \nu} = P'_{\mu \nu} =  \mathbf{x}_{\mu}\mathbf{G}^{-1}\mathbf{x}_{\nu}^T
\label{1.1.4.0}
\end{align}
 and hence its elements yield the conjugate training metric (\ref{0.1.0.1a}), (\ref{0.1.2c.10}) for the training observations - see for the MNIST example Figures \ref{Fig3.-26} and \ref{Fig3.-27}. Because the Gram matrix  $\mathbf {V^T V}$ of the truncated $\mathbf V$ is the identity matrix $\mathbf {I}_N$, if the columns of $\mathbf V$  have in addition zero means, the truncated $\mathbf V$ is often referred to as the \emph{whitened} data matrix.  The rows $\{ \mathbf{v}_{\mu} \}$ of the truncated $\mathbf V$ can be used to express the training observations: from (\ref{1.1.3}),
\begin{align}
\mathbf  {x}_{\mu}= \mathbf{v}_{\mu} \mathbf{\Lambda} \mathbf{W}^T, \quad \mu=1,...,M,
\label{1.1.4.1}
\end{align}
and this is why the matrix $\mathbf{\Lambda} \mathbf{W}^T$ is called \emph{de-whitening transformation}. The inverse transformation $ \mathbf{W} \mathbf{\Lambda}^{-1}$ , when well-defined, is called \emph{whitening transformation}:
\begin{align}
\mathbf  {v}_{\mu}= \mathbf{x}_{\mu} \mathbf{W}  \mathbf{\Lambda}^{-1}, \quad \mu=1,...,M.
\label{1.1.4.1a}
\end{align}
The conjugate analog of $<\mathbf{ x_{\mu},x_{\nu}}>_{\mathbb {T}}$ is from(\ref{1.1.4.0}): 
\begin{align}
<\mathbf{ x_{\mu},x_{\nu}}>_{\mathbb {T'}^{-1}} =  \mathbf{x}_{\mu}\mathbf{G}^{-1} \mathbf{x}_{\nu}^T = \nonumber \\ = <\mathbf{v}_{\mu}, \mathbf{v}_{\nu}>_{\mathbb E}= (\mathbf{VV}^T)_{\mu \nu} ,
\label{1.1.4.5}
\end{align}
 (cf. (\ref{0.1.0.1})). Unlike (\ref{0.1.2c.3}), there is no squared Gram matrix here.

\subsection{Eigen-observables. }
\label{Eigen-observables} 
 
\begin{figure}[!ht]
\vskip 0.2in
\begin{center}
\includegraphics[width=\columnwidth]{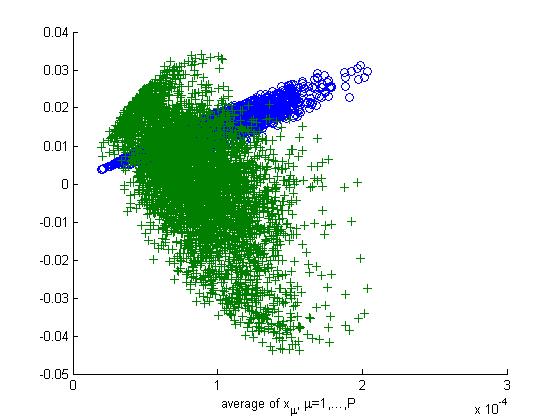}
\caption{Scatter plot of the values of the first eigen-observable $\mathbf{v}_1$ (circles) and the second eigen-observable $\mathbf{v}_2$ (crosses)  against  $\{\bar{\mathbf{x}}_{\mu} \}_{\mu=1}^P$, where $\bar{\mathbf{x}}_{\mu}= \sum_{i=1}^N X_{\mu i} $, using the first 5,000 MNIST images (i.e., $P = 5,000$).}
\label{Fig1.2.4}
\end{center}
\vskip -0.2in
\end{figure}

The columns of the truncated $\mathbf V$  form an orthonormal basis in the observable space. They are  called  \emph{eigen-observables}  $\{\mathbf{v}_{i}\}_{i=1}^M$  because they are the eigen-vectors of the other training Gram matrix $\mathbf G'$ (cf. (\ref{0.1.1b})), as column-vectors in $ \mathbb{R} ^P$. Because the eigen-observables are orthonormal, when their means are zero, they can also be referred to as  \emph{whitened observables}. The eigen-observables satisfy:
\begin{align}
\mathbf  {G' v}_{i}= \lambda_{i}^2 \mathbf{v}_{i} , \quad \mu=1,...,M.
\label{1.1.1b}
\end{align}
From the singular value decomposition (\ref{1.1.3}) and (\ref{1.1.4}), eigen-observables equal, up to a constant, the training mappings (\ref{0.0a}) of the transposed eigen-observations:
\begin{align}
\mathbf{v}_{i}  = \frac{1}{\lambda_i} \mathbb{T} \mathbf{ w}^T_i := \frac{1}{\lambda_i} \mathbf{X w}_i,
\label{1.1.6.1.-1}
\end{align}
and vice versa, from (\ref{0.0b}),
\begin{align}
\mathbf{w}^T_{i}  = \frac{1}{\lambda_i}   \mathbb{T'} \mathbf{v}_i := \frac{1}{\lambda_i} \mathbf{v}^T_i \mathbf{X}.
\label{1.1.6.1.-2}
\end{align}
This explains why eigen-observations and eigen-observables are self-conjugate, up to a constant:
\begin{align}
\mathbf  {\check{v}}_i= \frac{1}{ \lambda_i^2}\mathbf{v}_i, \quad i=1,...,M, 
\label{1.1.1a}
\end{align}
In addition, one can express training observables $\mathbf {x}_i$ via the orthonormal $\mathbb O$-basis of eigen-observables:
\begin{align}
\mathbf{x}_{i}  = \sum_{j} \left(\lambda_j W_{ ij }\right) \mathbf{v}_{j}.
\label{1.1.6.1}
\end{align}
Due to the relations (\ref{1.1.6.1.-1}), (\ref{1.1.6.1.-2}), the plots on  Figures (\ref{Fig1.2.2}) - (\ref{Fig1.2.3}) represent the eigen-observables as well. The scatter plot of the values of the top two eigen-observables against the averaged observation  intensities $\{\bar{\mathbf{x}}_{\mu} \}_{\mu=1}^P$  are shown in Figure \ref{Fig1.2.4} using the first 5,000 MNIST images. The first eigen-observable is not identical to the average observation intensity but appears to be highly correlated to it. Surprisingly, the second eigen-observable is negatively correlated to the average observation intensity.

\subsection{Observation probabilities. Non-Gaussianity.} 
 \label{Observation probabilities}
 
When asking the question of probability distributions in the training set, it is convenient to work with the whitened data matrix $\mathbf V$, instead of the original $\mathbf X$. Because of the linear relationships (\ref{1.1.6.1}) between the two, testing for Gaussianity e.g. is easier done on the whitened data matrix. 
 
 We did some preliminary analysis of the top ten eigen-observables in Figures \ref{Fig1.2.2} and  \ref{Fig1.2.2a} and did not find evidence of major deviation from Gaussianity for them. They represent only a very special small subset of mappings of the training set and the cited evidence is by no means representative of the overall distribution. 
 
\begin{figure}[!ht]
\vskip 0.2in
\begin{center}
\includegraphics[width=\columnwidth]{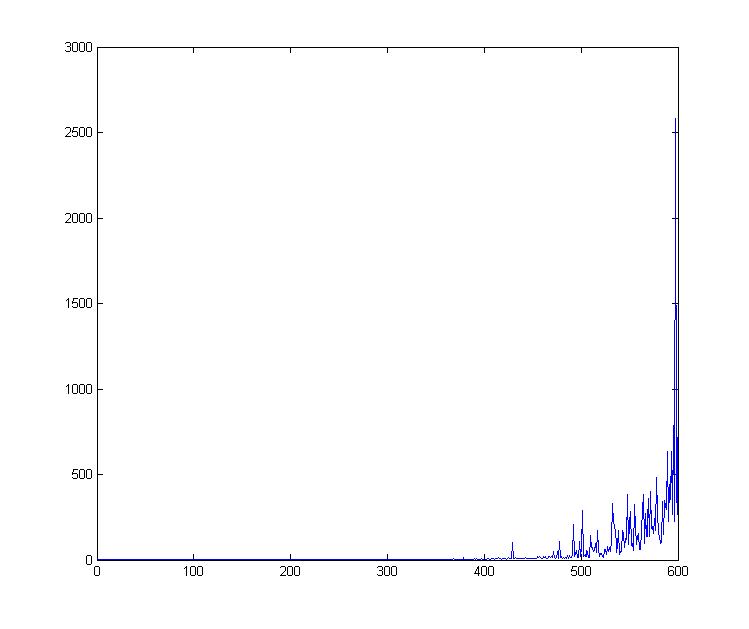}
\caption{The fourth moment $\mathbf{E(v}^4_i) $ $ = \frac{1}{N} \sum_{i=1}^N V_{\mu i}^4$ of the top $600$ eigen-observables in the MNIST dataset (using all $P=60,000$ observations, i.e. $60,000 = P \gg N=784$). All data was scaled up by a factor of $10^4$ , in order to ensure that the minimum eigenvalue exceeds $10^{-14}$, which is dangerously close to the \textbf{double} machine precision. The fourth moment of a Gaussian distribution with unit variance is $3$. Eigen-observables are ordered in descending order of their eigenvalues: clearly there is  a major break-down of Gaussianity for eigen-observables from the bottom half (to the right) which have super ``fat tails'' . }
\label{Fig1.2.5}
\end{center}
\vskip -0.2in
\end{figure}

We need a more quantitative measure of deviations from Gaussianity. A common metric for non-Gaussianity of, say the  whitened eigen-observables $\mathbf{v}_i$ $=\{\mathbf{v}_{\mu i}\}_{\mu=1} ^P$, is their fourth moment $\mathbf{E(v}^4_i) $, where $\mathbf{E}(.) $ signifies expected value. The fourth moments of the eigen-observables $\{\mathbf{v}_{ i}\}_{i=1} ^M$ for the  MNIST dataset are plotted on Figure \ref{Fig1.2.5} in three different scales. With the exception of the first few eigen-observables, the rest are \emph{super-Gaussian} i.e. have fatter tails than a Gaussian distribution. 
 
\begin{figure}[!ht]
\vskip 0.2in
\begin{center}
\includegraphics[width=\columnwidth]{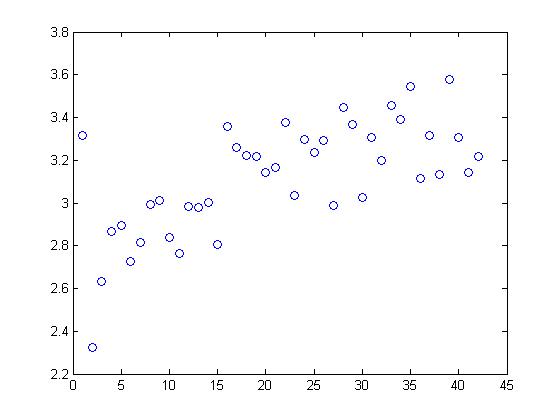}
\vskip 0.2in
\includegraphics[width=\columnwidth]{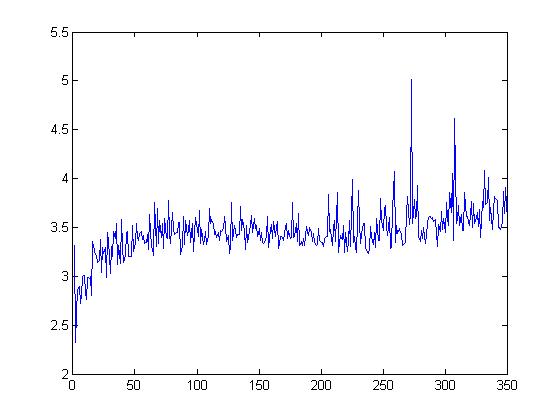}
\caption{ In order to appreciate the behavior of the leading  eigen-observables, we zoom in Figure \ref{Fig1.2.5} and plot separately the fourth moment of the leading $40$ (top) and the leading $350$ (bottom) eigen-observables. With the exception of the first few eigen-observables, the rest are \emph{super-Gaussian} i.e. have fourth moments significantly exceeding $3$ and  hence fatter tails than a Gaussian.}
\label{Fig1.2.5a}
\end{center}
\vskip -0.2in
\end{figure}

\begin{figure}[!ht]
\vskip 0.2in
\begin{center}
\includegraphics[width=\columnwidth]{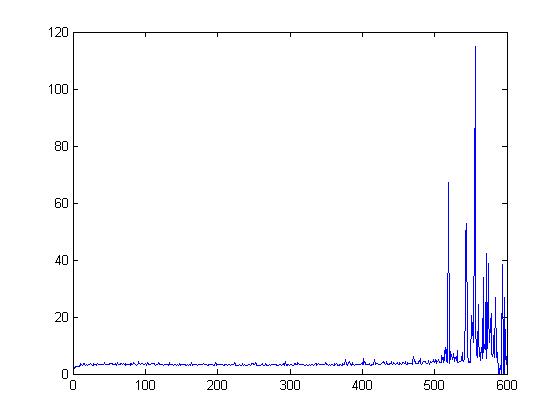}
\caption{The same as Figure \ref{Fig1.2.5} (the rescaled fourth moment of the top $600$ eigen-observables) but using only the first $P=1,000$ MNIST observations, i.e. $1,000 = P \approx N=784$. While there are some extreme fourth moments on the right, they are much more subdued than in the case  $P \gg N$ in Figure \ref{Fig1.2.5}. }
\label{Fig1.2.6}
\end{center}
\vskip -0.2in
\end{figure}

 This super-Gaussian behavior takes extreme proportions for the bottom half of the eigen-observables.  Let us dig a little deeper into this. Recall that the eigen-observable $\mathbf{v}_i$ is obtained by ``normalizing'' the eigen-mapping $\mathbf{\mathbf{X w}}_i$,i.e. dividing it by the respective eigenvalue $\lambda_i$.  The \emph{kurtosis} i.e. the fourth \emph{cumulant} 
 \begin{align}
 \kappa(\mathbf{\mathbf{X w}}_i) &= \mathbf{E( (\mathbf{X w}_i})^4) - 3(\mathbf{E(X w_i})^2))^2 
 \label{1.1.6.1a}
 \end{align}
of the eigen-mappings $\{\mathbf{X w}_i\}_{i=1}^{M}$ is the correct \footnote{If the data is de-meaned i.e. $\mathbf{E(X w_i}) = 0$.} ``un-whitened'' scale-dependent generalization of $\mathbf{E(v}^4_i) $.  When $P \gg N$ , as eigenvalues decrease, somewhere half-way through, the fourth moment $\mathbf{E( (\mathbf{X w}_i})^4)$ does not decrease as fast as $\lambda^4_i$, which gives rise to the extreme kurtosis on the right-hand side of Figure \ref{Fig1.2.5} (Figure \ref{Fig1.2.5a} zooms in on different parts of Figure \ref{Fig1.2.5}). As Figure \ref{Fig1.2.6} shows, this is not the case when $P \approx N$. While there is still visible non-Gaussianity for the eigen-observables at the bottom end, this non-Gaussian behavior starts a lot closer to the end and as the vertical scale shows, is not nearly as extreme as when $P \gg N$. This is not a priori obvious since the eigenvalues and the second moments are the same, irrespective of whether $P \approx N$ or $P \gg N$.

\section{Statistics and hierarchy.}
\label{Statistics and hierarchy}

Increasing the training paths length to an arbitrary large number naturally brings about the need for probabilistic and statistical considerations. We will present here combinatorial heuristics for the training graph which will lead to natural connections with Classical and Quantum Non-equilibrium Thermodynamics. In particular, we will see how  non-linear, bounded, monotonic, $C^{\infty}$ \emph{activation functions},  widely used  to connect layers in Neural Networks, arise naturally.

In the following sections, we will leverage off this intuition to develop from first principles both the statistics and the kinetics of the training set. We will first show how the classical Boltzmann statistics naturally arises in the equilibrium training graph and argue that it is inadequate in the general case because it describes only \emph{factorisable correlations} between observations or observables. In a geometric sense, this is equivalent to a global flattening of the metric in the training space. The non-equilibrium nature of the real world on the other hand  demands  breaking the reversibility in time i.e. the rise of \emph{arrow of time}. Markov processes irreversible in time violate the Principle of Detailed Balancing (\cite{Lifshitz81}, Ch.2) and hence have non-symmetric in the real domain Hamiltonians i.e. transition matrix.

The vast majority of Machine Learning procedures either assume a priori independence of training observations or ``force it'' when taking a \emph{Thermodynamic Limit} $N \rightarrow \infty$ (see e.g. \cite{MacKay98}, 11.1 for the case of Gaussian Mixture Models or \cite{Kingma14-1}, \cite{Rezende14} for the so-called Bayesian Variational models). Ordered observations (respectively observables) which are independent or have dependencies which are of short-term (resp. short-range) nature i.e. ``strongly mixing'', have nice asymptotic properties when taken to the Thermodynamic Limit: they satisfy a classical Central Limit Theorem i.e. the distribution of their mean converges to a Gaussian which corresponds to the familiar \emph{Boltzmann statistics} in Statistical Physics, \cite{Landau80}, Sec 40.

When on the other hand the dependencies or interactions are long-term (resp. long-range), more general \emph{super-statistics} come in play, \cite{Beck09}. A subset of those are the so-called \emph{Tsallis statistics} for which a  q-Central Limit Theorem holds \cite{Tsallis09}.  The limit case $q \rightarrow 1$ corresponds to the classical Boltzmann statistics. For the  cases of $q > 1$, the asymptotic limit is a distribution with fatter polynomial tails than the Gaussian exponential and the classical exponentials from the Boltzmann statistics are replaced by the so-called \emph{Tsallis q-exponential}.   

In physics terms, the Boltzmann statistics corresponds to low occupational densities (see below) i.e. to very weak statistical ``interactions'' or dependencies. For stronger dependencies, there is a rich formalism developed for the needs of Quantum Field Theory and Statistical Physics: one now considers the tensor product of  replicas of our original space of observations (resp. observables) and new statistics emerge.  The classical \emph{Bose-Einstein} Statistics described by \emph{bosons} is the case of symmetric tensor products and \emph{Fermi} statistics described by  \emph{fermions} is the case of skew-symmetric tensor products.  More generally, if the statistics changes as the size of the tensor product of replicas increases, one arrives at the so-called \emph{exclusion statistics} introduced relatively recently in physics by \cite{Haldane91} (see \cite{Murthy09} for more recent review). It has an intimate relation with \emph{Quantum} or \emph{q-Groups} (\cite{Lusztig94}),  Vertex Operator Algebras, \cite{Lepowsky04}, representations of Infinite-Dimensional Lie Algebras, \cite{Georgiev96}, and many other seemingly unrelated areas in physics and mathematics. The exclusion statistics have similar appearance to the Boltzmann statistics but with the classical exponential replaced by another \emph{q-exponential} which differs from the Tsallis q-exponential. For a generic training set, both the Tsallis q-statistics and the exclusion statistics will come into play.

Let us go back to our training graph (Sub-section \ref{Graph-theoretical view.}). Inspired by quantum-mechanical analogies, the inner products above and their derivatives can be thought of as being proportional to transition probabilities between observations (resp. observables in the dual picture). A path is then a  monomial of observations (resp. observables) and ``lives'' in the tensor products of $\mathbb {O}$ (resp. $\mathbb { O'}$). In this interpretation,  the path-sums quantify  the \emph{interactions} between the starting and ending observation, in particular, a path of length $K$ accounts for a specific $K$ -observation \emph{interaction}. There is no reason to choose a priori a specific path-length, so one has to consider sums over paths of all possible lengths. 
In practice, the length of the training paths will be limited because many inner products  $\mathbf{<x_{\mu}, x_{\nu}>}_{\mathbb E}$ will be ``too small'' and can be discarded i.e. we have \emph{sparsity}. As discussed above, in this case, we will not consider the two vertices $\mathbf{x_{\mu}, x_{\nu}}$ connected and hence the training graph will not be fully-connected but instead broken into \emph{irreducible} sub-graphs . 

There are a number of different ways to to consider arbitrary long paths and ensure convergence in the limit to infinitely long paths.  In the spirit of social network graphs, \cite{Barabasi01}, we will associate every vertex in the factorized graph with an \emph{energy level} $\varepsilon_i$ and a link between two vertices with a \emph{quasi-particle} $i$. We will assume at first that interactions between vertices are \emph{factorisable} i.e. we only have paths between identical vertices and thus our graph  can be decomposed into mutually unconnected sub-graphs. The edges in these sub-graphs correspond to the  training mappings defined in Section \ref{The training set}  and thus a $K_i$-step path is a collection of $K_i$ identical quasi-particles of type $i$. The path-length (number of quasi-particles) $K_i$ is commonly referred to in physics as \emph{occupation number}.

\subsection{Equilibrium, noise and hierarchy}
\label{Equilibrium, noise}

Our goal here is, roughly speaking, to find from first principles the  most likely value of the path-lengths (occupation numbers) $\{K_i\}$ in \emph{ equilibrium}. In order to explain what that means, let us note that in a general (\emph{non-equilibrium}) condition, the state of a   system  with variable occupation numbers, like ours, is determined by both the energy $\varepsilon_i$ and the occupation number $K_i$ of the state, among possibly other \emph{macro variables}. Also, except  for zero temperature $T=0$ (see below), there is an ambient uncertainty i.e. \emph{noise} because of the large number of \emph{micro variables} $\{\xi_{\alpha}\}$ whose dynamics is too complicated to quantify. In other words, macro variables like $\varepsilon_i$ and $K_i$ can not be calculated exactly but are \emph{ensemble averages} over the complicated and generally unknown probability distribution of the micro variables $\mathcal{P}^{(0)}(\xi_{\alpha})$ :
\begin{align}
\varepsilon_i = \bar{\varepsilon_i}(\xi_{\alpha}), \quad K_i = \bar{K_i}(\xi_{\alpha}),
\label{1.1.-3}
\end{align}
subject to certain constraints on the micro variables. For systems which can be broken into a large enough sub-systems, one can consider higher levels of \emph{hierarchy} where the former macro-variables become micro-variable and new class of macro-variables emerge, which are averages of the former ones. For example, the macro variable occupation numbers of the so-called \emph{Boltzmann} statistics (\ref{1.1.0.4}) become micro variables at the higher hierarchical \emph{Fermi} and \emph{Bose-Einstein} Statistics (\ref{1.1.0.5}), (\ref{1.1.0.6}). As is common in Statistical Physics, in order to simplify notations, for a given level of hierarchy, we will omit the bar signs, signifying averages. 

A full set of macro variables defines a \emph{state} of our system and Statistical Physics was built as an attempt to ``cancel out'' the micro variables completely and conjure up laws i.e. equations between measurable macro variable alone. If one succeeds, the frequency of occurrence or likelihood of a state is a  higher-hierarchical probability distribution $\mathcal{P}^{(1)}$ $=\mathcal{P}^{(1)}(\varepsilon_i, K_i)$ which is a function of the macro variables   alone (will skip for simplicity the full list of macro variables here). This distribution is \emph{the noise at the new higher hierarchical level}. There are situations where the energy  $\varepsilon_i$ is independent of $K_i$ and cases where $\varepsilon_i(K_i)$ is a function of $K_i$ but in general, one can think of $\mathcal{P}^{(1)}$ as a function of $K_i$:

\begin{align}
\mathcal{P}^{(1)}=\mathcal{P}^{(1)}(\varepsilon_i(K_i), K_i) = \mathcal{P}^{(1)}( K_i)
\label{1.1.-2}
\end{align}
In this context, \emph{equilibrium}  is the state with maximum log likelihood i.e. the state which maximizes the \emph{entropy} $\mathcal{S} = ln(\mathcal{P})$, \cite{Landau80}, Ch.XII.  It is determined, if one finds the explicit functional dependence $K_i = \varphi_i(\varepsilon_i)$ which solves the optimization problem:
\begin{align}
\max_{K_i} \quad ln \mathcal{P}^{(1)}(\varepsilon_i, K_i). 
\label{1.1.-2.1}
\end{align}  
In probabilistic terms, in equilibrium, the probability distribution $\mathcal{P}^{(1)}$ has a \emph{mode} for $K^{Eq}_i = \varphi_i(\varepsilon_i)$. The averaging of $K_i$ over the higher-level distribution $\mathcal{P}^{(1)}$ yields same results as the averaging (\ref{1.1.-3}) over the lower-level distribution  $\mathcal{P}^{(0)}$ but with some of the constraints relinquished. Hence, like the Gaussian distribution, the distribution of $K_i$ has the  mode equal its mean  i.e.
\begin{align}
 K^{Eq}_i = \varphi(\varepsilon_i) = \bar{K}_i (\varepsilon_i).  
\label{1.1.-2.2}
\end{align}

In practice, in order to account for the scale of the system, one does not work directly with the occupation numbers $\{K_i\}$ but instead, with  the \emph{occupation densities} $\{k_i\}$,  obtained by dividing $K_i$ by the respective \emph{characteristic scale } or \emph{degeneracy} $L_i$ for that state i.e. $k_i = K_i/L_i$. In Quantum Statistical Mechanics, the role of $L_i$ is played by the number of quantum occupation states available to the respective quasi-particles e.g. the degeneracy at a given energy level $\varepsilon_i$. In Classical Statistical Mechanics, $L_i$ is the rescaled phase volume i.e. 
\begin{align}
L_i = \frac{1}{(2\pi \hbar)^r}\Delta p^{(i)} \Delta q^{(i)} 
\label{1.1.-1}
\end{align}
where $\Delta p^{(i)} \Delta q^{(i)}$ is a small region in the phase space ($p, q$ are respectively the momenta and coordinates) but large enough so as to contain a statistically large number of quasi-particles, $r$ is the dimension of the system and $\hbar$ is the Planck constant (cf. for example \cite{Landau80}, Sec. 38). The  phase space in the numerator has the dimension of action in the physics sense of the word i.e. ``energy x time''. 

In our context, the momenta are ``integrated out'' and squashed into ``noise'', and we will resort to the ``energy x time'' interpretation. We will define it as the number of training observations $\{\mathbf{x}_{\mu}\}$ with yet to be defined  energy  $\mathcal{H}(\mathbf{x}_{\mu})$ in the range $\varepsilon_i \pm \Delta \varepsilon_i$, multiplied by the number of steps needed  to update all observables $i$, $i=1,...,N$:
\begin{align}
L_i \sim \{\# \mu| \mathcal{H}(\mathbf{x}_{\mu}) \in \varepsilon_i \pm \Delta \varepsilon_i  \} \{\# i| X_{\mu i} \neq X_{\mu +1 ,i} \}.
\label{1.1.-1a}
\end{align}
In the so-called \emph{parallel dynamics}, all observables are updated simultaneously and the second multiplier is $1$. In the so-called \emph{sequential dynamics}, the observables are updated one by one by drawing randomly from the set $\{1,2,...,N\}$ (cf. \cite{Coolen02-1}) and hence  the second multiplier is $\sim N $ .

Let us summarize the task of finding equilibrium in the context of the training graph: We are looking for Maximum Likelihood of appropriate distributions for occupation densities (scaled path lengths) of observables, possibly at different hierarchy levels, i.e.
\begin{align}
\max_{k_i}  \quad ln \mathcal{P}(\varepsilon_1,k_1,\varepsilon_2,k_2,...,\varepsilon_M,k_M), 
\label{1.1.0}
\end{align}
subject to constraints:
\begin{align}
\sum_i L_i k_i = K , \quad \sum_i \varepsilon_i L_i k_i = E,
\label{1.1.0.1}
\end{align}
where $M = dim(\mathbb O)$, cf. Section \ref{The training set}, $K$ is the total number of quasi-particles (total sum of path-lengths) and $E$ is the total energy. The convention in Statistical Physics is to ``hide'' the characteristic scales in the constraints by summing in addition over a ``degeneracy'' index $l_i$ such that $\sum_{l_i=1}^{L_i} 1 = L_i$ which corresponds to summing over all ``sub-states'' corresponding to a given energy level $\varepsilon_i$. Then the constraints read  as:
\begin{align}
\sum_{i, l_i} k_i = K , \quad \sum_{i,l_i} \varepsilon_i k_i = E.
\label{1.1.0.1a}
\end{align}
 
 In the space of observations, the characteristic lengths are the same (because training parallelogram has the same characteristic size) but the occupation densities are in general different and could have a very different probability density distribution:
\begin{align}
\max_{k'_i}  \quad ln \mathcal{P}(\varepsilon_1, k'_1,\varepsilon_2,k'_2,...,\varepsilon_M,k'_M), 
\label{1.1.0.2}
\end{align}
subject to constraints:
\begin{align}
\sum_{i,l_i}  k'_i = K',\quad \sum_{i,l_i} \varepsilon_i k'_i = E'.
\label{1.1.0.3}
\end{align}

\subsection{Fermionization and Bozonization}
\label{Fermionization and}

The three statistics most often arising in Statistical Mechanics are the familiar \emph{Boltzmann, Fermi} and \emph{ Bose-Einstein}  Statistics. Let us discuss them in our context, using notations for  occupation densities $k_i$ and energies $\varepsilon_i$  (cf. \cite{Landau80}, Sec. 40, 55):

\bigskip
i) \emph{Boltzmann Statistics}, valid when $k_i \ll 1$ and:
 \begin{align}
k^{\mathcal B}_i   = \bar{k}_i =  e^{\alpha - \beta \varepsilon_i}.
\label{1.1.0.4}
\end{align}
 where averaging is over micro variables (subject to constraints) as in (\ref{1.1.-3}), $\alpha = \mu /T$ and $\beta =  1/T$, $\mu$ is the so-called \emph{chemical potential} and $T$ is the \emph{temperature} (we assume the Boltzmann constant to be 1). The occupation densities are subject to the constraints (\ref{1.1.0.1a}) for the aggregate number of quasi particles and energy, in particular, 
\begin{align}
\sum_{i, l_i} p^{\mathcal B}_i = \sum_{i, l_i} \frac{k^{\mathcal B}_i}{K} = 1 ,
\label{1.1.0.4a}
\end{align}
and hence $p^{\mathcal B}_i$ $=k^{\mathcal B}_i/K$ can be interpreted as \emph{weights} or probability densities over all states, including degeneracies. 

One should stress that the Boltzmann statistics is meant for \emph{ideal gases} i.e. systems with negligible interactions between their components. It is inapplicable (i.e. its validity constraint  $k_i \ll 1$ is violated), if the chemical potential is fixed  and the temperature $T$ drops to zero, (\ref{1.1.0.4}): 
\begin{align}
\lim_{T \rightarrow 0}{k^{\mathcal B}_i} = \infty
\label{1.1.0.4b}
\end{align}
In other words, at low temperatures, ideal Boltzmann gases ``condense'' and the interactions between their components can not be neglected anymore. As a result, non-trivial higher hierarchies are created, as discussed in the text between (\ref{1.1.-3}) and (\ref{1.1.-2}). Examples of such higher hierarchy statistics are the Fermi and Bose-Einstein statistics introduced below.

Conversely, for fixed chemical potential and high temperatures $T \gg 0$, interactions fade and any probability distribution, including the Fermi and Bose-Einstein distributions, will eventually converge towards the Boltzmann distribution, ((\ref{1.1.0.5.1}), (\ref{1.1.0.6b}) below).

\bigskip
ii) \emph{Fermi Statistics}, valid when $k_i  \sim k^{\mathcal B}_i L_i \sim 1$:
 \begin{align}
k^{\mathcal F}_i  = \bar{k}^{\mathcal B}_i = \frac{1}{e^{-\alpha + \beta \varepsilon_i} + 1},
\label{1.1.0.5}
\end{align}
where averaging is over the Boltzmann distribution (\ref{1.1.0.4}), subject to the the constraints that $k_i = 0$ or $k_i = 1 $. In other words, it is obtained by mixing together a large  (of the order of magnitude of the characteristic scale $L_i$)  number of independent Boltzmann sub-systems (quasi-particles) from (\ref{1.1.0.4}) so that $k_i \sim k^{\mathcal B}_i L_i \sim 1$, subject to the the constraints that $k_i = 0$ or $k_i = 1 $. This process is called \emph{fermionization} and the resulting system (quasi-particle) is a \emph{fermion}. As mentioned above, for high enough temperatures, the Fermi distribution morphs back into the Boltzmann distribution:
 \begin{align}
\lim_{T \rightarrow \infty}{k^{\mathcal F}_i} = {k^{\mathcal B}_i}.
\label{1.1.0.5.1}
\end{align}

The Fermi distribution (\ref{1.1.0.5}) is easily derived from its interpretation as a mixture of independent Boltzmann sub-systems, subject to the above constraints. Because of independence, the probability of mixing $r k^{\mathcal B}_i \sim L_i k^{\mathcal B}_i$ Boltzmann sub-systems of type $i$ is $(p^{\mathcal B}_i)^{rk^{\mathcal B}_i}$ and hence, after canceling out $K$, we obtain (\ref{1.1.0.5}) from  
 \begin{align}
k^{\mathcal F}_i   = \bar{k}^{\mathcal B}_i = \frac{0 (p^{\mathcal B}_i)^0  + 1 (p^{\mathcal B}_i)^1 }{  (p^{\mathcal B}_i)^0 + ( p^{\mathcal B}_i)^1 }   .
\label{1.1.0.5a}
\end{align}
In the context of the earlier discussion, this is an example of a higher hierarchy distribution $\mathcal P^{(1)}$ built on top of the lower hierarchy Boltzmann distributions $\mathcal P^{(0)}$.
Note that the probability for the fermion to be in either of its two states is given by 
 \begin{align}
p^{\mathcal F}_i :=Prob(k^{\mathcal B}_i = 1)   &= \frac{ (p^{\mathcal B}_i)^1 }{  (p^{\mathcal B}_i)^0 + ( p^{\mathcal B}_i)^1 } =\frac{1}{e^{-\alpha + \beta \varepsilon_i} + 1},  \nonumber \\
Prob(k^{\mathcal B}_i = 0)   &= 1 - Prob(k^{\mathcal B}_i = 1),   
\label{1.1.0.5b}
\end{align}
i.e. it is given by the \emph{logistic function}. The  probability  can also be expressed via a trigonometric hyperbolic function as follows:
 \begin{align}
p^{\mathcal F}_i = \frac{1}{2} \left(1 -  tanh\left(\frac{\alpha - \beta \varepsilon_i}{2}\right)  \right),   
\label{1.1.0.5c}
\end{align}
It is convenient to describe the two fermion states via  a new macro variable \emph{spin} which flips sign between the two states i.e.  $\sigma_i$ $=\pm 1$:
 \begin{align}
Prob \left(\sigma_i = 1\right)  & := p^{\mathcal F}_i =\frac{1}{e^{-\alpha + \beta \varepsilon_i} + 1},  \nonumber \\
Prob \left(\sigma_i = -1\right)   &= 1 - p^{\mathcal F}_i.   
\label{1.1.0.5d}
\end{align}
With the standard parametrization $\alpha = \mu /T$ and $\beta =  1/T$ from (\ref{1.1.0.4}), noise magnitude is determined by the temperature $T$ and the noise disappears at $T=0$. This is the familiar non-linear, bounded, monotonic, $C^{\infty}$ activation function in \emph{recurrent} (stochastic) Neural Networks: for  given spins in all states $\{\sigma_j(t)\}$ at time $t$ and some typically linear function $\varepsilon_i$ $=  - h_i(\{\sigma_j(t)\}) + \mu$ of the spins in all states, the spin $\sigma_i(t+1)$ at time $t + 1$ is randomly drawn from the above distribution i.e. with probability:
 \begin{align}
Prob \left(\sigma_i(t+1) = 1\right)  = \frac{1}{e^{-\frac{1}{T} h_i(\{\sigma_j(t)\})} + 1}.  
\label{1.1.0.5e}
\end{align}
The average of the spin in  this higher hierarchy noise distribution is  easily computed to be:
 \begin{align}
\bar{\sigma}_i =   tanh\left(\frac{\alpha - \beta \varepsilon_i}{2}\right).
\label{1.1.0.5f}
\end{align}
 This is one of the \emph{activation functions} in recurrent Neural Networks, where instead of random sampling as above, one takes averages as time progresses: 
 \begin{align}
\sigma_i(t+1)  =  tanh \left( \frac{1}{2T} h_i(\{\sigma_j(t)\}) \right).  
\label{1.1.0.5g}
\end{align}

\bigskip
iii) \emph{Bose-Einstein Statistics}, valid when $k_i \sim 1 $ or $k_i \gg 1$ :
 \begin{align}
k^{\mathcal BE}_i  = \bar{k}^{\mathcal B}_i = \frac{1}{e^{-\alpha + \beta \varepsilon_i} - 1},
\label{1.1.0.6}
\end{align}
without any constraints for $k_i$. The Bose-Einstein distribution is derived from Boltzmann distributions, using:

 \begin{align}
k^{\mathcal BE}_i   = \bar{k}^{\mathcal B}_i = \frac{0 (p^{\mathcal B}_i)^0  +  1 (p^{\mathcal B}_i)^1 + \dots +  k (p^{\mathcal B}_i)^k + \dots}{  (p^{\mathcal B}_i)^0 + ( p^{\mathcal B}_i)^1 + \dots + (p^{\mathcal B}_i)^k + \dots }   .
\label{1.1.0.6a}
\end{align}
The only condition is that $p^{\mathcal B}_i < 1$ for all $\varepsilon_i,$ hence  $\mu < 0.$  This is anothern example of a higher hierarchy distribution $\mathcal P^{(1)}$ built on top of the lower hierarchy Boltzmann distributions $\mathcal P^{(0)}$. As mentioned above, for high enough temperatures, the Bose-Einstein distribution morphs back into the Boltzmann distribution:
 \begin{align}
\lim_{T \rightarrow \infty}{k^{\mathcal BE}_i} = {k^{\mathcal B}_i}.
\label{1.1.0.6b}
\end{align}

\section{Training Statistics.}
\label{Training statistics}

We will focus here on applying the concepts developed above to the training set.

\subsection{Characteristic training scales}
\label{Characteristic training}

Since we have all the machinery ready, lets introduce here the concept of \emph{characteristic scales} in the training space.  As mentioned above during the introduction of the physical concepts of occupation numbers and densities, we will need some notion of ``scale'' and ``volume'' in our training set. There are the analogues of phase space volume in Classical Statistical Mechanics  and the number of quantum states in Quantum Statistical Mechanics. 
As mentioned above, the \emph{characteristic scale} $L_i$ of the $i-$the eigen-observation (or eigen-observable) is expected to be of the same order of magnitude as the degeneracy or \emph{multiplicity} of the resp. energy eigenvalue $\varepsilon_i$ i.e.
  \begin{align}
L_i \sim mult(\varepsilon_i).
\label{1.1.6.4}
\end{align} 
Since $\varepsilon_i$ are related to the eigenvalues of the Gram matrices $\lambda_i^2$ and will change as time progresses and one adds more observations, let us discuss their asymptotic behavior. As the size of our training set grows, so do its characteristic scales. For random matrices consisting of independent identical Gaussian elements with unit variance, the asymptotic behavior (and even the distribution) of the largest eigenvalue $\lambda_1$ is well known (cf. \cite{Johnstone01}). It grows as the square root of the number of observations $P$ and the number of observables $N$:
   \begin{align}
 \lambda_1 \sim \sqrt{N} + \sqrt{P}, \quad P,N \rightarrow \infty.
\label{1.1.6.5}
\end{align} 
 The asymptotic distribution of the remaining eigenvalues is smaller than (\ref{1.1.6.5}) (\cite{Juhasz81}). One can therefore assume that there exists $n \in \mathbb N$, $n < N$, such that the largest $n$ eigenvalues increase in value according to (\ref{1.1.6.5}), as the system scales up  i.e. 
\begin{align}
\frac{\sum_{i=1}^n \lambda_i}{\sum_{i=1}^N \lambda_i} &\sim 1,\\
 P,N & \rightarrow \infty \qquad \forall i =1,..,n.
\label{1.1.6.6} 
  \end{align}

\subsection{Training Boltzmann statistics}
 \label{Training Boltzmann}
 
We are now ready to go back to the original goal of computing the most likely probability distribution (\ref{1.1.0}) of path-lengths (occupation numbers) in our training graph, subject to constraints (\ref{1.1.0.1}).  We will consider training graphs with eigen-observations as vertices and edges corresponding to training mappings defined in Section \ref{The training set}. 

Let's start with the sub-graph of eigen-observations of type $i$.
 As mentioned in the discussion leading to (\ref{1.1.0}) and (\ref{1.1.0.1}), every vertex will be associated with an energy level $\varepsilon_i$, yet to be determined. As discussed above, the equivalent of the number of available states for our training graph are the characteristic lengths $L_i \sim mult(\varepsilon_i)$ as in (\ref{1.1.6.4}). As mentioned above, the characteristic scales will change with time and as the training graph grows.  Paths between identical vertices are indistinguishable so our sub-graph is not ordered. The statistical weight of a $K_i$-step loop path  is hence the number of unordered sequences of length $K_i$, composed of the numbers $1,..., L_i$: 
\begin{align}
Prob(K_i)  \sim \frac{L_i^{ K_i}}{K_i!} \sim \frac{mult(\varepsilon_i)^{ K_i}}{K_i!}. 
\label{1.1.8}
\end{align}
The number of its permutations $K_i!$ in the denominator accounts for fact that the sub-graph is not ordered.

We can now move on to the full training graph. The orthogonality of the eigen-observations implies that the full graph will be an union of mutually unconnected sub-graphs, each corresponding to  distinct type $i$ of  eigen-observations. Therefore the full probability density (\ref{1.1.0}) is a product of the individual densities. Switching again to occupation densities $\{k_i\}$, $K_i = k_i L_i$, the equilibrium problem from (\ref{1.1.0}),(\ref{1.1.0.1})  i.e. finding Maximum Likelihood distribution of occupation densities is:
\begin{align}
\max_{k_i}  \quad S = ln \mathcal{P}(\varepsilon_1,k_1,\varepsilon_2,k_2,...,\varepsilon_M,k_M) = \nonumber \\
=\sum_i ln \frac{L_i^{ L_i k_i}}{(L_i k_i)!}, 
\label{1.1.9}
\end{align}
subject to constraints:
\begin{align}
\sum_i L_i k_i = K , \quad \sum_i \varepsilon_i L_i k_i = E,
\label{1.1.10}
\end{align}

We will show for completeness that the Boltzmann distribution is the solution of this optimization problem, following \cite{Landau80}, Section 40. Due to the Stirling approximation of log-factorial:
\begin{align}
ln K! \cong K ln(K/e), K \rightarrow \infty 
\label{1.1.20}
\end{align}
one has from (\ref{1.1.9}), after cancellation of $ln(L_i)$ terms,
\begin{align}
S \cong -\sum_i L_i k_i  ln(k_i/e).
\label{1.1.21}
\end{align}
Hence, after cancellation of $L_i$ term,
\begin{align}
\frac{\partial S}{\partial k_i}   \cong -L_i  ln(k_i). 
\label{1.1.22}
\end{align}
Using standard Lagrangian multipliers technique with Lagrangian coefficients $\alpha$ and $\beta$ for the constraints  (\ref{1.1.10}), 
\begin{align}
\frac{\partial }{\partial k_i} \left( S + \alpha K - \beta E\right)  =0, 
\label{1.1.23}
\end{align}
one arrives from (\ref{1.1.22}) at the Boltzmann distribution (\ref{1.1.0.4}):
\begin{align}
k_i   \sim e^{\alpha - \beta \varepsilon_i}. 
\label{1.1.24}
\end{align}

\section{Observations statistics: ferromagnetic and anti-ferromagnetic case.}
\label{Observation statistics}

We will focus here on the space of observations and look for appropriate statistical representation. As discussed in the introduction of Section \ref{Statistics and hierarchy}, most of modern Machine Learning assumes independence of observations while in real life they are clearly highly correlated (cf. Figures \ref{Fig3.0} and \ref{Fig3.-1} in Section \ref{The training set})

We will start from the Gram matrix for overlaps $\mathbf {G'}^2$ and the respective training graph, introduced in Section \ref{The training set}. Recall that according to (\ref{0.1.2c.1}), the $\nu$-th overlap $(\mathbf{x}_{\mu})_{\nu}$ of the $\mu$-th observation $\mathbf{x}_{\mu}$ is the matrix element $G'_{\mu \nu}$ of the observations Gram matrix $\mathbf G'$. Moreover, switching from observations to their overlaps ``flattens'' the highly non-trivial metric $\mathbf{x_{\mu}} \mathbf{G} \mathbf{x^T_{\nu}}$, defined by the metric tensor $\mathbf G$, into a plain Euclidean metric $(\mathbf {G'^{2}})_{\mu \nu}$ (cf. (\ref{0.1.2c.3})).

Recall that, in order to capture the ``pure'' overlap of an observation $\mu$ with the rest of the training observations, the squared self-overlap $(G'_{\mu \mu})^2$, which corresponds to a self-loop in the graph should be subtracted from the sum of all squared overlaps $(\mathbf{G'}^2)_{\mu \mu}$ (cf. (\ref{0.1.2c.4})). The resulting difference, taken  with negative sign, is smallest for those observations which have largest overlaps with rest of the training observations. It can therefore be interpreted as ``potential energy'' $ \varepsilon_{\mu}$ of the observation $\mu$:
\begin{align}
 \mathcal{H}(\mathbf{x}_{\mu}) = \varepsilon_{\mu} \sim -\frac{1}{2} \left((\mathbf {G'}^2)_{\mu \mu} - (G'_{\mu \mu})^2  \right) = \nonumber \\
 = -\frac{1}{2}\left(\mathbf{x_{\mu}} \mathbf{G} \mathbf{x_{\mu}}^T  - (\mathbf{x_{\mu}}  \mathbf{x_{\mu}}^T)^2 \right) < 0.
\label{0.1.2c.4a}
\end{align}
This is a common assumption in Neural Networks, including in the so-called \emph{Hopfield} model and in the \emph{Restricted Boltzmann Machines}   (see \cite{Coolen05} , \cite{Bengio12})\footnote{The self-interaction in the Hopfield model is strictly speaking $\sum_{i} X_{\mu i}^2 G_{ii}$ $=\sum_{\nu, i} X_{\mu i}^2  X_{\nu i}^2$ $\neq (G'_{\mu \mu})^2$ (cf. \cite{Coolen05}, (21.3)) but we don't expect the difference to be material for a typical dataset.}. The motivation for this form of the energy is the desire to conjure-up time evolution for the conjugate observation $\mathbf{\check{q}}$, given by
\begin{align}
 \Delta \mathbf{\check{q}}  = - \frac{\partial {\mathcal{H}(\mathbf{\check{q}}) }}{\partial {\mathbf{\check{q}}}} + noise,
\label{0.1.2c.4c}
\end{align}
which, if we ignore the smaller self-loop term, reads from (\ref{0.1.2c.4a}): 
\begin{align}
 \Delta \mathbf{\check{q}} \sim \mathbf{G}\mathbf{\check{q}} + noise =  \mathbf{q} + noise,
\label{0.1.2c.4d}
\end{align}
for the conjugate observation $\mathbf{\check{q}}$, introduced in (\ref{0.0c.-1}), (\ref{0.0d}). This is the discrete version of the \emph{ Langevin} equation - the fundamental equation of Non-Equilibrium Thermodynamics - in the so-called ``strong friction'' limit i.e. for times exceeding significantly the characteristic period of the noise and for unit \emph{mobility} (cf. \cite{Chavanis14} for a modern overview). When the noise is Gaussian, the time-continuous version of this equation describes in the time evolution of $N$ 1-dimensional Brownian particles. The Langevin equation is the stochastic generalization of the Hamiltonian equations of classical mechanics and the observation $\mathbf{q}$ plays the role of a ``force'' driving the time evolution of its conjugate $\mathbf{\check{q}}$.

Inspired by conjugate training metric (\ref{0.1.2c.10}), as opposed to the training metric, one can re-write Hamiltonian (\ref{0.1.2c.4a}) as follows:
\begin{align}
 \mathcal{H}(\mathbf{x}_{\mu}) =  -\frac{1}{2}\left(\mathbf{x_{\mu}} \mathbf{G}^{-1}\mathbf{x_{\mu}}^T  - (\mathbf{x_{\mu}}  \mathbf{x_{\mu}}^T)^2 \right) < 0,
\label{0.1.2c.4e}
\end{align}
and then (\ref{0.1.2c.4d}) is replaced by:
\begin{align}
 \Delta \mathbf{q} \sim \mathbf{G}^{-1} \mathbf{q}  + noise =  \mathbf{\check{q}} + noise,
\label{0.1.2c.4f}
\end{align}
 i.e. the conjugate observation $\mathbf{\check{q}}$ plays the role of a ``force'' driving the time evolution of $\mathbf{q}$.

In the presence of noise, which is proportional to the temperature $T$, we showed in Section \ref{Training statistics} that equilibrium  distribution is given the Boltzmann statistics (\ref{1.1.0.4}):
\begin{align}
 \mathcal{P^B}(\varepsilon_{\mu}) \sim e^{-{\frac{1}{T} \varepsilon_{\mu}}  }.
\label{0.1.2c.5}
\end{align}
After normalizing (\ref{0.1.2c.5}) to ensure that $\sum \mathcal{P}(\mu) = 1$, one has:
  \begin{align}
 \mathcal{P^B}(\varepsilon_{\mu}) = \frac{e^{-{\frac{1}{T}  \varepsilon_{\mu} }}}{\mathcal{Z}}, \qquad \mathcal{Z} = \sum_{\mu} e^{-{\frac{1}{T}  \varepsilon_{\mu} } },
 \label{0.1.2c.6}
\end{align}
  where $\mathcal{Z}$ is the so-called \emph{training partition function}. We inserted ``training'' in its name to stress the fact that we are summing up over visible training observations only. In this sense, it is constrained, i.e. it can be thought of as integral over all arbitrary observations of $\mathbf q \in \mathbb{R}^N$, with Dirac delta functions inserted in it:  
 \begin{align}
 \mathcal{Z} = \idotsint \prod_{\mu} \delta(\mathbf{q} - \mathbf{x}_{\mu})  e^{-{\frac{1}{T}  \mathcal{H}(\mathbf{q}) }}dq_1 ...dq_N .
 \label{0.1.2c.6a}
\end{align}

Going back to the energies definition (\ref{0.1.2c.4a}),(\ref{0.1.2c.4e}),  let us note that models where higher overlaps are favored and better alignment of observables is more probable, are called \emph{ferromagnetic} in physics. The energies in  (\ref{0.1.2c.4a}), (\ref{0.1.2c.4e}) are ferromagnetic because they favor alignment, for example in the pixels in the MNIST dataset. If  images of uniformly lit screens i.e. same pixel intensity were  part of the MNIST dataset, they would have been the most probable in these ferromagnetic models.

Conversely, a model is called \emph{anti-ferromagnetic}, if it favors maximum local misalignments between observables. In the case of images for example, the most probable images in an anti-ferromagnetic model would have been those with alternating intensities for neighboring pixels. Flipping the sign of the energy definition (\ref{0.1.2c.4a}) obviously turns our original ferro-magnetic model into an anti-ferromagnetic model:
  \begin{align}
 \varepsilon_{\mu} \sim \frac{1}{2} \left((\mathbf {G'}^2)_{\mu \mu} - (G'_{\mu \mu})^2  \right) = \nonumber \\
 =\frac{1}{2}\left(\mathbf{x_{\mu}} \mathbf{G} \mathbf{x_{\mu}}^T  - (\mathbf{x_{\mu}}  \mathbf{x_{\mu}}^T)^2 \right) > 0,
\label{0.1.2c.4b}
\end{align}
and (\ref{0.1.2c.4e}) turns into:
  \begin{align}
 \varepsilon_{\mu} \sim \frac{1}{2}\left(\mathbf{x_{\mu}} \mathbf{G}^{-1} \mathbf{x_{\mu}}^T  - (\mathbf{x_{\mu}}  \mathbf{x_{\mu}}^T)^2 \right) > 0,
\label{0.1.2c.4g}
\end{align}

\begin{figure}[!ht]
\vskip 0.2in
\begin{center}
\includegraphics[width=\columnwidth]{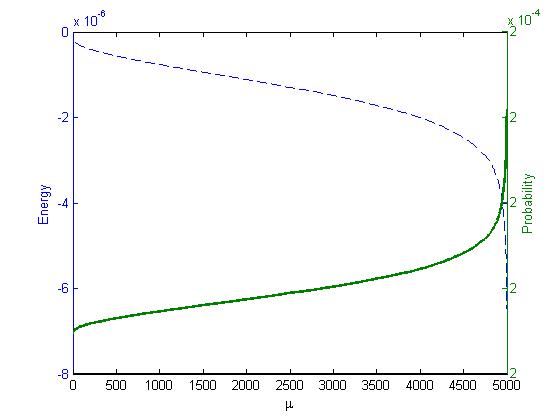}
\vskip 0.2in
\includegraphics[width=\columnwidth]{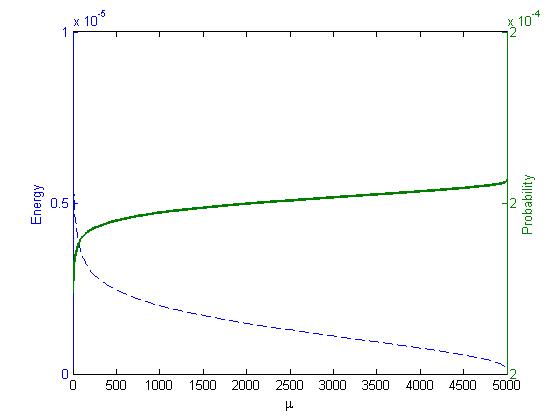}
\caption{ The equilibrium energies  in descending order (dashed line), and the respective probability  $\mathcal{P}(\varepsilon_{\mu})$   (solid line),  for the first 5,000 images from  MNIST dataset $\mu = 1,...,5000$ $(P = 5,000$). The ferromagnetic case (\ref{0.1.2c.4a}) is at the top and the anti-ferromagnetic case  (\ref{0.1.2c.4b}) is at the bottom.}
\label{Fig3.1}
\end{center}
\vskip -0.2in
\end{figure}

We visualize in Figure \ref{Fig3.1} the sorted energy $\varepsilon_{\mu}$ and respective probability  $ \mathcal{P}(\varepsilon_{\mu})$ for the first 5,000 MNIST observations for both the ferromagnetic case (\ref{0.1.2c.4a}) (top) and the anti-ferromagnetic case (\ref{0.1.2c.4b}) (bottom). \footnote{We assumed $T=1$.} There is clearly something wrong in the ferromagnetic case: the low energy states which correspond to stationarity and equilibrium states do not look stationary at all - there is no flattening associated with a convergence behavior near  a stationary point! 
            
\begin{figure}[!ht]
\vskip 0.2in
\begin{center}
\includegraphics[width=\columnwidth]{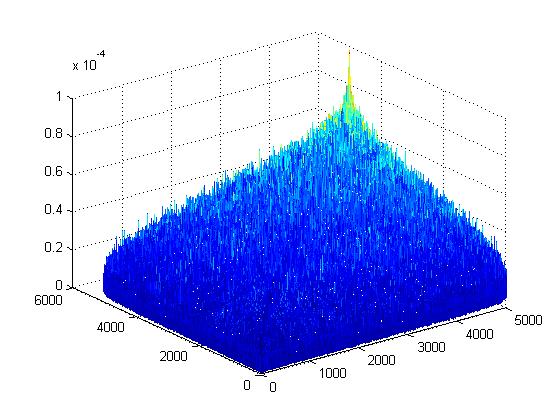}
\vskip 0.2in
\includegraphics[width=\columnwidth]{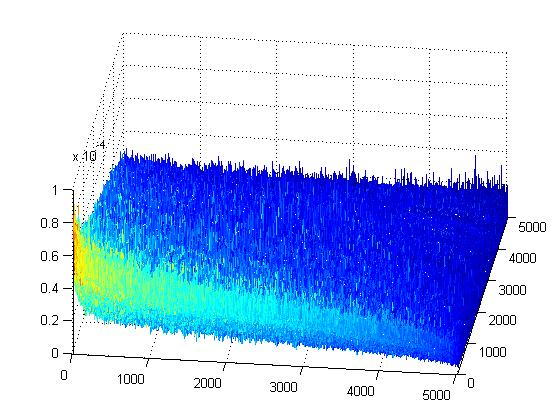}
\caption{$\mathbf{G'}$ with all observations sorted in ascending order of $\mathcal{P}(\varepsilon_{\mu})$ for the first 5,000 images from  MNIST dataset (i.e., $P = 5,000$). The ferromagnetic case (\ref{0.1.2c.4a}) is at the top  and the anti-ferromagnetic case  (\ref{0.1.2c.4b}) is at the bottom.}
\label{Fig3.0a}
\end{center}
\vskip -0.2in
\end{figure} 

To drive that point further, we plot in Figure \ref{Fig3.0a} $\mathbf G'$ for both the ferromagnetic and anti-ferromagnetic case, with the observations re-ordered in the new order - from lowest to highest probability (compare against the original Gram matrix $\mathbf G'$ from Figure \ref{Fig3.0}). After the initial spike, $\mathbf G'$ in the ferromagnetic case  has the recognizable look of the  non-stationary auto-covariance matrix of a Gaussian random walk. \footnote{Recall that the auto-covariance matrix $\mathbf{E}(Z_{\mu} Z_{\nu})$ of a 1-dimensional Gaussian process $\{Z_{\mu} \}$ with Gaussian increments $Z_{\mu+1} - Z_{\mu}$ $\sim \mathcal{N}(0,1)$ is given by $min(\mu, \nu)$.  This ``Gaussian'' auto-covariance shapes under the new order are surprising and certainly not a priori obvious, especially for $\mathbf {G'}^2$ : we created the order by merely sorting the diagonals of $\mathbf {G', G'}^2$, without any awareness of the rest of the matrix elements!}
 
One of the problems in the ferromagnetic case stems from the inapplicability of the Boltzmann statistics for low temperatures i.e. for a low or absent noise and negative energies. In other words, when the condition $e^{-{\frac{1}{T} \varepsilon_{\mu}}} \ll 1$, needed for the applicability of the Boltzmann statistics, is violated, we have the wrong distribution!\footnote{See for more details the discussion leading to  (\ref{1.1.0.4b}) .} This could have been expected: in the exponential defining $\mathcal{P}(\varepsilon_{\mu})$, we have a quadratic function with a positive sign, which dominates the other term. So we have a Gaussian distribution but with the wrong sign in the exponential!
  
\begin{figure}[!ht]
\vskip 0.2in
\begin{center}
\includegraphics[width=\columnwidth]{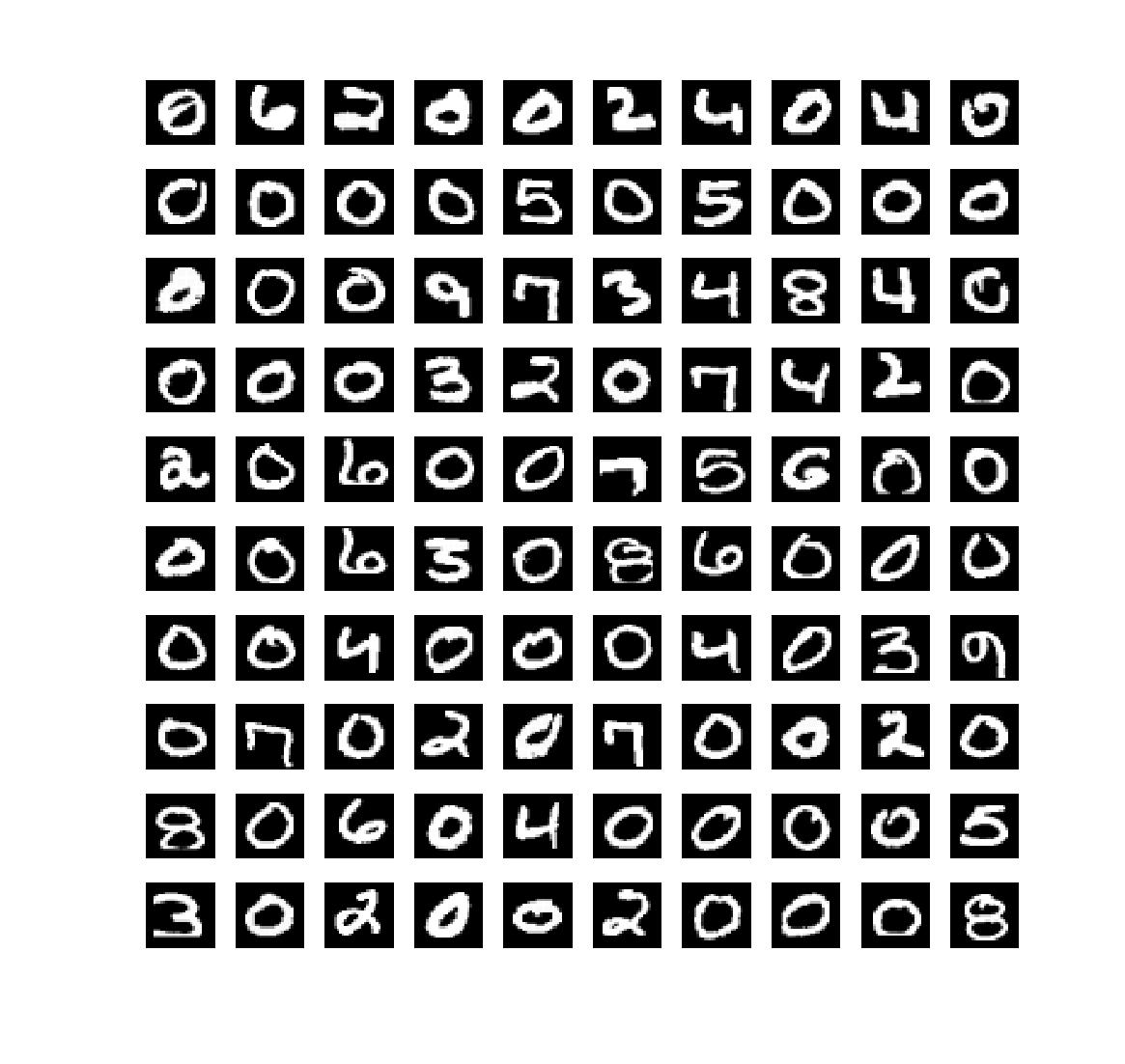}
\caption{The 100 most likely images in the ferromagnetic case (\ref{0.1.2c.4a}) from  the first 5,000 images from  MNIST dataset (i.e., $P = 5,000$).}
\label{Fig3.3}
\end{center}
\vskip -0.2in
\end{figure}

To get a visual intuition for the overlaps and the related probabilities, we plot  in Figure \ref{Fig3.3} the one hundred most probable observations for  the ferromagnetic case (\ref{0.1.2c.4a}). As expected, the very likely observations in the ferromagnetic case (\ref{0.1.2c.4a}) are very ``bloated'' because  $ \mathcal{P}(\varepsilon_{\mu})$ is proportional to the average overlapping of observation $\mu$ with the rest of the dataset.

\section{Neural network architecture.}
\label{Neural network}

Generally speaking, the typical problem of Machine Learning is inferring a structure in the training matrix $\mathbf X$ i.e. finding an explicit function $\mathbf{Y}: \mathbf{X \rightarrow Y(X)} $ which in turn  allows to generate ''typical values'' either for observables or observations or both. In the most ambitious scenario, $\mathbf Y$ is the probability density itself. Less taxing methods like Regression, Classification, Dimension Reduction etc, are all special cases of this general learning problem, the differences between them stemming mostly from the nature of the ''noise'' and the ways it is introduced to the system. Neural Networks offer one of the few generic methods which seem to be applicable to most learning problems. 

\subsection{Core architecture.}
\label{Core architecture}

\begin{figure}[!ht]
\vskip 0.2in
\begin{center}
\begin{tikzpicture}

\node[rectangle, text width = 0.6\columnwidth, text  centered,  draw ] 
	(Input) at (5,0) {\underline{1. \textbf{Input layer} }:\\Size = \# observables  N};
\node[rectangle,  text width = 0.6\columnwidth,  text  centered, below=0.5in of Input, draw] 
	(EncoderHidden)  {\underline{2. \textbf{Encoder hidden layer(s)} }: \\Size = $H_{enc}$};
\node[rectangle,  text width = 0.6\columnwidth,  text  centered, below=1in of EncoderHidden, draw] 
	(Latent/Feature)  {\underline{3. \textbf{Latent hidden layer(s)}}: \\Size = $H_{lat}$ \\ $\begin{array}{ll} \mathbf{AE}:&\text{ambient manifold} \\ &\text{+ latent random  processes} \\ \mathbf{C}:&\text{contra-variant features}\end{array} $ };
\node[rectangle,  text width = 0.6\columnwidth,  text  centered, below=1in of Latent/Feature, draw] 
	(DecoderHidden)  {\underline{4. \textbf{Decoder hidden layer(s)}}:  \\Size = $H_{dec}$, \text{ Spanned by:} \\  $\begin{array} {ll} \mathbf{AE}:&\text{transformed random}  \\ &\text{processes} \\ \mathbf{C}:&\text{feature basis vectors} \end{array} $  };
\node[rectangle,  text width = 0.6\columnwidth,  text  centered, below=0.5in of DecoderHidden, draw] 
	(Output)  {\underline{5. \textbf{Output layer}}:\\Size = \\ $\begin{array}{ll}\mathbf{AE}: &\text{\# observables } N \\ \mathbf{C}: &\text{\# classes } C \end{array} $};

\path [->](Input) edge node {$\begin{array}{l} \text{Affine mapping} \\ \& \quad \text{non-linear mapping} \end{array}$} (EncoderHidden); 
\path [->](EncoderHidden) edge node {$\begin{array}{l} \text{Affine projection made out of:} \\ \begin{cases} \mathbf{AE}:\text{encoding weights \& biases} \\ \mathbf{C}: \text{filters a.k.a. ``receptive fields''} \end{cases}\\ \& \quad \text{non-linear mapping} \end{array} $  } (Latent/Feature);
\path [->](Latent/Feature) edge node {$\begin{array}{l} \text{Affine mapping made out of:} \\ \begin{cases}\mathbf{AE}:\text{decoding weights \& biases} \\ \mathbf{C}: \text{covariant feature basis vectors} \end{cases}\\ \& \quad \text{non-linear mapping} \end{array} $} (DecoderHidden); 
\path [->](DecoderHidden) edge node {$\begin{array}{l} \text{Affine mapping}  \\ \& \quad \text{non-linear mapping}  \end{array}$} (Output); 

\draw [->]  ([xshift=0.2in]Output.east) edge[thick]  node [rotate=270,  midway, yshift = 0.3in] { $\begin{array}{l} \textbf{Back-propagation } of  \min \left(- log \mathcal{L} + reg~constraints\right),    \\ \begin{cases} \mathbf{AE}:- log \mathcal{L}_{AE} = reconstruction~error \\ \mathbf{C}: - log \mathcal{L}_{C} = classification~error \end{cases} \end{array} $} ([xshift=0.2in]Input.east);
\draw [decorate, decoration={brace}]  ([xshift=-0.2in, yshift=0.2in]Latent/Feature.west) -- ([xshift=-0.2in]Input.west) node [midway,rotate=90,yshift=0.2in] {Encoder};
\draw [decorate, decoration={brace}]  ([xshift=-0.2in]Output.west) -- ([xshift=-0.2in, yshift=-0.2in]Latent/Feature.west) node [midway,rotate=90,yshift=0.2in] {Decoder}; 
         
\end{tikzpicture}
\caption{Core Auto-Encoder / Classifier architecture . $\mathbf{AE}$ stands for ``Auto-Encoder'', $\mathbf{C}$ stands for ``Classifier''. }

\label{Fig2.0}
\end{center}
\vskip -0.2in
\end{figure}
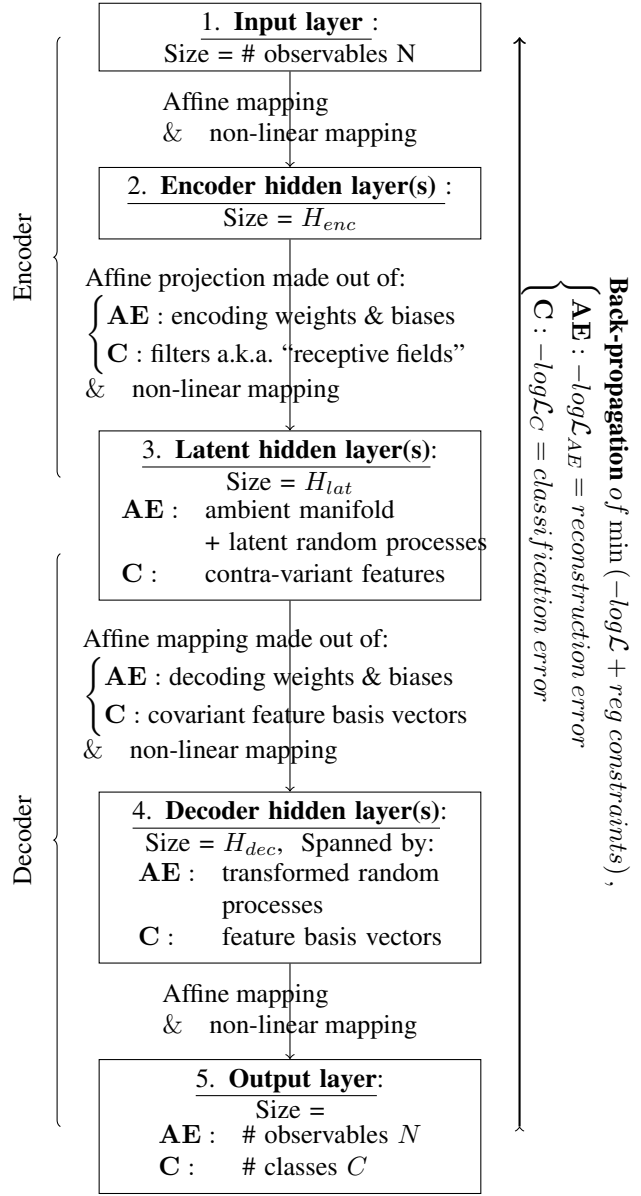

The core architecture of Auto-Encoder and Classifier is on Figure \ref{Fig2.0}. There are three main components: an \emph{Encoder}, a \emph{Latent} hidden layer(s) in the middle and a \emph{Decoder}. The latent layer encapsulates the ``coded'' input. For Classifiers, this is where the \emph{``features''} of the data set are. For probability density-describing nets like Variational Auto-Encoders, Restricted Boltzmann Machines, etc, this is where an ambient lower-dimensional manifold of the ``coded input'' is. Naturally, if the model is stochastic and/or generative, this is where the random number generation takes place. 

Subsequent layers are connected to one another by a composition of: 

i) affine mappings, with the tensors/matrices in the linear part usually referred to as \emph{``weights''} and the translation vectors referred to as \emph{``biases''} and  

ii) non-linear, so-called \emph{``activation''} mappings, which we will refer to as \emph{non-linearities}. 

The weights/biases are determined from a training data set via \emph{Back-propagation}, minimizing typically a negative log-likelihood function $- log \mathcal{L},$ where $\mathcal{L}$ is typically \emph{cross-entropy} between data- and model- distributions, plus additional regularization terms. The log-likelihood has a mandatory \emph{reconstruction error} component for Auto-Encoders and \emph{classification error} component for Classifiers. 

The Universal Approximator theorems for Neural Networks (\cite{Cybenko89}, \cite{Hornik90}) imply that the Encoder needs at least two hidden layers (including the latent) in order to approximate arbitrary well any given continuous function. The reason is that the non-linearities are chosen a priori and fixed thereafter, so one such non-linearity is generally not enough to approximate an arbitrary non-linear function. By considerations of symmetry, the Decoder also needs at least one additional hidden layer, hence the minimum five-layer architecture presented on Figure \ref{Fig2.0}.

Special cases:

1. \emph{Shallow} Auto-Encoder: having one  hidden (latent) layer only. It has two important special cases:
\begin{itemize}
\item \emph{Tied weights}: the weight matrix between the input and the latent layer is the transpose of the weight matrix between the latent and the output layer

\item \emph{Tied layers}: iterations are performed where the output layer is fed into the input layer and the reconstruction error and the respective gradients are calculated after the last iteration only.  \footnote{Note that the iterative step $j$ in a  tied-layer net is not to be confused with the iterative step $m$  in generic numerical optimization procedures (Sections \ref{Dimension reduction}, \ref{Numerical optimization}).} For generative nets, the random sampling in each of the iterative steps is referred to as \emph{Gibbs sampling}.
\end{itemize}

2. \emph{Denoising} Auto-Encoder : random sampling takes place at the input layer (\cite{Vincent08}).

An  example of a generative Shallow auto-encoder with tied weights of significance is the Restricted Boltzmann Machines (presented below in Sub-section \ref{Restricted Boltzmann}). It appears to have been historically  the first universal scalable neural net and influenced all subsequent developments.

\subsection{Shallow auto-encoder and its dual.}
\label{Shallow auto-encoder} 

We will present here the mathematical formulation of the special case of a Shallow auto-encoder i.e. with one hidden layer only. We will work in parallel in the observables and observations spaces and assume for simplicity that there is no random sampling and the bias vectors are all zero.

\begin{figure*}[!ht]
\vskip 0.2in
\begin{minipage}[]{\columnwidth}
\begin{tikzpicture}
\node[rectangle, text width =0.6 \columnwidth,  text  centered,  draw ] 
	(Input) at (5,0) {\underline{1. \textbf{Input layer}}: \\$\mathbf{x}_{\mu}$ \\Size = \# observables $N$ } ;
\node[rectangle,  text width = 0.5 \columnwidth,  text  centered, below=0.5in of Input, draw] 
	(Latent/Feature)  {\underline{2. \textbf{Latent hidden layer}}: \\ $\mathbf{y}_{\mu}=\varphi_{enc}(\mathbf{x}_{\mu} \mathbf{W^{(1)}} + \mathbf{b}^{(1)}) $ };
\node[rectangle,  text width = 0.6 \columnwidth,  text  centered, below=0.5inof Latent/Feature, draw] 
	(Output)  {\underline{3. \textbf{Output layer}}: \\ $\mathbf{\hat{x}_{\mu}}=\varphi_{dec}(\mathbf{y}_{\mu} \mathbf{W^{(2)}} + \mathbf{b}^{(2)}) $ 
	\\Size = \# observables $N$};
\path [->](Input) edge node {$\begin{array}{l} \text{Affine mapping ($ \mathbf{W}^{(1)}, \mathbf{b}^{(1)} )$} \\ \& \quad  \text{non-linear mapping $\varphi_{enc}$} \end{array}$} (Latent/Feature); 
\path [->](Latent/Feature) edge node {$\begin{array}{l} \text{Affine mapping ($ \mathbf{W}^{(2)}, \mathbf{b}^{(2)})$} \\ \& \quad  \text{non-linear mapping $\varphi_{dec}$} \end{array}$} (Output); 
\draw [->]  ([xshift=0.2in]Output.east) -- ([xshift=0.2in]Input.east) node [rotate=270, midway, yshift = 0.2in] { Backpropagation};
\draw [decorate, decoration={brace}]  ([xshift=-0.35in, yshift=0.2in]Latent/Feature.west) -- ([xshift=-0.2in]Input.west) node [midway,rotate=90,yshift=0.2in] {Encoder};
\draw [decorate, decoration={brace}]  ([xshift=-0.2in]Output.west) -- ([xshift=-0.35in, yshift=-0.2in]Latent/Feature.west) node [midway,rotate=90,yshift=0.2in] {Decoder};
\end{tikzpicture} 
\caption{Shallow auto-encoder in the space of observables.}
\label{Fig2.1}
\end{minipage}\hfill
\begin{minipage}[]{\columnwidth}
\begin{tikzpicture}
\node[rectangle, text width = 0.6 \columnwidth, text  centered,  draw ] 
	(Input) at (5,0) {\underline{1. \textbf{Input layer}}: \\ $\mathbf{x}_{i}$ \\Size = \# observations $P$} ;
\node[rectangle,  text width = 0.5 \columnwidth, text  centered, below=0.5inof Input, draw] 
	(Hidden)  {\underline{2. \textbf{Latent hidden layer}}: \\ $\mathbf{y'}_{i}=\varphi_{enc}(\mathbf{V^{(1)}} \mathbf{x}_{i} + \mathbf{b}^{(1)}) $ };
\node[rectangle,  text width = 0.6 \columnwidth, text  centered, below=0.5in of Hidden, draw] 
	(Output)  {\underline{3. \textbf{Output layer}}: \\ $\mathbf{\hat{x}_{i}} = \varphi_{dec}(\mathbf{V^{(2)}}\mathbf{y'}_{i} + \mathbf{b}^{(2)}) $ \\
	Size = \# observations $P$};
\path [->](Input) edge node {$\begin{array}{l} \text{Affine mapping ($ \mathbf{V^{(1)}},\mathbf{b}^{(1)}) $} \\  \& \quad \text{non-linear mapping $\varphi_{enc}$ } \end{array}$} (Latent/Feature); 
\path [->](Latent/Feature) edge node {$\begin{array}{l} \text{Affine mapping ($ \mathbf{V^{(2)}}, \mathbf{b}^{(2)})$}  \\ \& \quad  \text{non-linear mapping $\varphi_{dec}$}\end{array}$ } (Output); 
\draw [->]  ([xshift=0.2in]Output.east) -- ([xshift=0.2in]Input.east) node [rotate=270, midway, yshift = 0.2in] { Backpropagation};
\end{tikzpicture}
\caption{Shallow auto-encoder in the space of observations.}
\label{Fig2.1a}
\end{minipage}\hfill
\vskip -0.2in
\end{figure*}

\subsubsection{The encoder stage.}

\smallskip
i) For observables:\\ 
In covariant terns, the weight matrix is  $N \times n$ matrix $\mathbf{W^{(1)} } : \mathbb{O} \rightarrow \hat{\mathbb{O}}$: 
\begin{align}
\mathbf{x_{i}} \rightarrow \mathbf{ y_i} =  \mathbf{\sum_{j=1}^{N} \mathbf{x}_j } W^{(1)}_{ji}. 
\label{0.2}
\end{align}
 It can be translated conveniently in contra-variant terms, as matrix multiplication in the space of observations:
\begin{align}
\mathbf{x_{\mu}} \rightarrow \mathbf{ y_{\mu}} =  \mathbf{x_{\mu} W^{(1)}} = \Bigg \{\sum_{j=1}^{N} X_{\mu j}  W^{(1)}_{ji} \Bigg\}_{i=1}^{N}
\label{0.3}
\end{align}
i.e. one transforms every ''visible'' training observation $\mathbf x_\mu$ into a ''hidden'' n-dim observation  $\mathbf {y_\mu=x_\mu W^{(1)} }$. In Machine Learning, if the columns of $\mathbf W^{(1)}$ are orthonormal, they are referred to as \emph{feature basis vectors} of the training set.

\smallskip
ii) For observations:\\ 
In covariant terms, the weight matrix is  $n \times P$ matrix $\mathbf{V^{(1)} } : \mathbb{O'} \rightarrow \hat{\mathbb{O'}}$: 
\begin{align}
\mathbf{x_{\mu}} \rightarrow \mathbf{ y'_{\mu}} = \sum_{\nu=1}^{P}  V^{(1)}_{\mu \nu}  \mathbf{ \mathbf{x}_{\nu} }.
\label{0.2b}
\end{align}
It can be translated as matrix multiplication  in the space of observables:
\begin{align}
\mathbf{x_{i}} \rightarrow \mathbf{ y'_{i}} =  \mathbf{V^{(1)} x_{i} } = \Bigg \{\sum_{\nu=1}^{P} V^{(1)}_{\mu \nu} X_{\nu i}   \Bigg\}_{\mu=1}^{P}
\label{0.3b}
\end{align}
i.e. one transforms every \emph{''visible''} input  training observable $\mathbf x_i$ into a ''hidden'' n-dim observable $\mathbf {y'_i=  V^{(1)}x_i }$.

\subsubsection{The decoder (reconstruction) stage.}

\smallskip
i) For observables: \\ In covariant terms, the weight matrix is   $n \times N$ matrix $\mathbf{W^{(2)} } : \hat{\mathbb{O}} \rightarrow \mathbb{O}$  which transforms every \emph{''hidden''} training observable $\mathbf y_i$ into a ''output'' n-dim observable $\mathbf{ \hat{x}_i} $ resembling the input observable as much as possible:
\begin{align}
\mathbf{y_{i}} \rightarrow \mathbf{ \hat{x}_i} =  \mathbf{\sum_{j=1}^{N} \mathbf{y}_j } W^{(2)}_{ji}. 
\label{0.4}
\end{align}
In contra-variant terms, this can be  translated as a matrix multiplication for observations:
\begin{align}
\mathbf{y_{\mu}} \rightarrow \mathbf{ \hat{x}_{\mu}} =  \mathbf{y_{\mu} W^{(2)}} = \Bigg\{ \sum_{j=1}^{N} Y_{\mu j}  W^{(2)}_{ji} \Bigg\}_{i=1}^{N}.
\label{0.5}
\end{align}
The weights are considered \emph{tied} if $\mathbf{W^{(2)} }  = \mathbf{W^{(1)T} }$.

\smallskip
ii) For observations: \\ In contra-variant terms, the weight matrix is   $P \times n$ matrix $\mathbf{V^{(2)} } : \hat{\mathbb{O'}} \rightarrow \mathbb{O'}$:
\begin{align}
\mathbf{y'_{\mu}} \rightarrow \mathbf{ \hat{x}_{\mu}} =  \sum_{\nu=1}^{P} V^{(2)}_{\mu \nu} \mathbf{y'}_{\nu}  . 
\label{0.4b}
\end{align}
In contra-variant terms, this can be translated as a matrix multiplication of observables:
\begin{align}
\mathbf{y'_{i}} \rightarrow \mathbf{ \hat{x}_{i}} =  \mathbf{ V^{(2)} y'_{i}} = \Bigg\{ \sum_{\nu=1}^{P} V^{(2)}_{\mu \nu} Y'_{\nu i}   \Bigg\}_{\mu=1}^{P}.
\label{0.5b}
\end{align}
The weights are considered \emph{tied} if $\mathbf{V^{(2)} }  = \mathbf{V^{(1)T} }$.

\bigskip
In terms of the full training matrix $\mathbf X$, one can rewrite the two steps in the respective spaces as:
\begin{align}
\mathbf {Y} &= \mathbf{ X W^{(1)} }, \quad \mathbf{\hat{X}} =\mathbf {Y W^{(2)} }   \nonumber \\
\mathbf {Y'} &= \mathbf{V^{(1)}  X  }, \quad \mathbf{\hat{X}} =\mathbf {V^{(2)} Y'  },
\label{0.6}
\end{align}
where $\mathbf Y$ is a $ P \times n$ matrix and $\mathbf Y'$ is a $n \times N$ matrix. The pairs  $\mathbf{W^{(1)}, W^{(2)} }$ ( resp. $\mathbf{V^{(1)}, V^{(2)} }$) have to be such as to minimize the \emph{reconstruction} error between the input $\mathbf X$ and the output $\mathbf {\hat{X}}$ across all training observations. In matrix terms:

\begin{align}
Recon~Err &= \mathbf{|| X - \hat{X} ||}^2 _F = \nonumber \\
&= Tr\{\mathbf {(X - \hat{X})^T(X-\hat{X}) } \} \nonumber \\
&=  Tr\{\mathbf {(X - \hat{X})(X-\hat{X})^T } \} 
\label{0.7}
\end{align}
where $||.||^2_F$ is the squared \emph{Frobenius norm} of a matrix i.e. sum of squares of its elements. As a function of the encoding and decoding matrices, the reconstruction error is:

\begin{align}
Recon ~Err = f(\mathbf{W^{(1)},W^{(2)}}) = \nonumber \\
= Tr\{\mathbf {(X- X W^{(1)}W^{(2)})^T(X-X W^{(1)} W^{(2)}) } \} ,
\label{0.7b}
\end{align}

\begin{align}
Recon ~ Err = f(\mathbf{V^{(1)},V^{(2)}}) = \nonumber \\
= Tr\{\mathbf {(X- V^{(2)}V^{(1)}X )(X-V^{(2)} V^{(1)}X )^T } \} 
\label{0.7c}
\end{align}
where $Tr{}$ is the Trace matrix operator.  A more detailed form of the error function is given in Appendix \ref{Error functions}.

\subsection{Restricted Boltzmann Machine.}
\label{Restricted Boltzmann}        
 
The Restricted Boltzmann Machine (RBM) is a neural net with a probability density assumed to be of the shape described by the Gibbs distribution in statistical physics. It  appears to have been the first universal net in the sense of being capable of  classification, density estimation, generation etc. It turns out, when the so-called \emph{Contrastive Divergence} training method is used (\cite{Hinton02}), the RBM can be thought of as a special case of a generative shallow auto-encoder with the following features: 

i) tied weights: $\mathbf{W^{(2)} }  = \mathbf{W^{(1)T} }$,

ii) tied non-linearities (optional): $\varphi = \varphi_{enc} = \varphi_{dec}$, 

iii) denoising (opional): random sampling takes place in both the latent and the input/output layers, 

iv) hidden layer is binary (optional), 

v) log-likelihood is approximate: cost to be minimized consists of reconstruction error only.
 
\begin{figure}[!ht]
\vskip 0.2in
\begin{center}
\begin{tikzpicture}
\node[rectangle, text width =0.6 \columnwidth,  text  centered,  draw ] 
	(Input) at (5,0) {\underline{1. \textbf{Input layer}}: \\ $\mathbf{x}_{\mu}(0) = \mu$-th data vector} ;
\node[rectangle,  text width = 0.6 \columnwidth,  text  centered, below=0.5in of Input, draw] 
	(Latent/Feature)  {\underline{2. \textbf{Latent hidden layer}}: \\ $\mathbf{y}_{\mu}(j)= \text{random sampling}$ \\ \text{with mean} \\ $\varphi(\mathbf{x}_{\mu}(0) \mathbf{W^{(1)}} + \mathbf{b}^{(1)}) $ };
\node[rectangle,  text width = 0.6 \columnwidth,  text  centered, below=0.5inof Latent/Feature, draw] 
	(Output)  {\underline{3. \textbf{Output layer}}: \\ $\mathbf{\hat{x}_{\mu}}(j)= $\\$ =\varphi(\mathbf{y}_{\mu}(j) \mathbf{W}^{(1)T} + \mathbf{b}^{(2)}) $ 
	};

\path [->](Input) edge node {} (Latent/Feature); 
\path [->](Latent/Feature) edge node {}( Output); 
\path [ ->] (Output.west) edge [bend left=30] node [rotate=90, midway, yshift = 0.2in] {$j$ -th iteration} (Input.west); 

\draw [->]  ([xshift=0.2in]Output.east) -- ([xshift=0.2in]Input.east) node [rotate=270, midway, yshift = 0.2in] { Backpropagation after j iterations};
\end{tikzpicture}
\caption{Restricted Boltzmann Machine in the space of observables.} 
\label{Fig2.1.1}          
\end{center}
\vskip -0.2in
\end{figure}
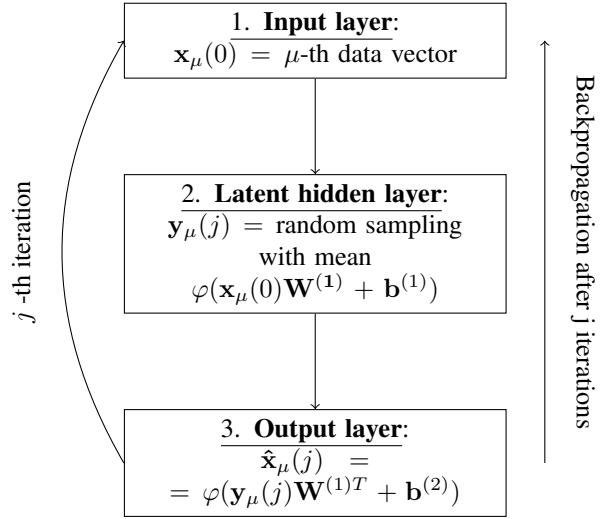

The architecture is drawn in Figure \ref{Fig2.1.1} (in the space of observables). 
 
\section{Dimension reduction: exact linear algebraic solution.}
\label{Dimension reduction}

Dimension Reduction is  a search for structures in a lower dimensional space which encapsulate as much as possible the structure of the original data. We will work in parallel in the observables and observations spaces:

i) In the space of training observables $\mathbb O$ , we want to reduce the number of observables from $N$ to $n <= N$ i.e. transform the N-dim row-vectors $\mathbf x_\mu, \mu = 1,..., P,$  into n-dim vectors for some $n <= N$ and then reconstruct them back into N-dimensional vectors with the least loss of information. 

ii) In the dual space of training observations $\mathbb O'$ , we want to reduce the number of observations from $P$ to $n <= P$ i.e. transform the P-dim column vectors $\mathbf x_i, i = 1,..., N,$  into n-dim vectors for $n <= P$ and then reconstruct them back into P-dimensional vectors with the least loss of information.
We will present here the exact linear-algebraic solution of the Dimension reduction problem and show how it naturally leads to different recipes for numerical solutions outlined in the  Section \ref{Numerical optimization}.

\subsection{Singular value decomposition solution.}
\label{Singular value solution}
Recall the truncated singular value decomposition of the matrix $\mathbf X$ from (\ref{1.1.3}), (\ref{1.1.4.0}) for the case $P >= N$:

\begin{equation}
\mathbf{X = V \Lambda W}^T,
\label{3}
\end{equation}
where $\mathbf V$ is $P\times N$ one-sided orthogonal matrix ($\mathbf {V}^T \mathbf{V = I}_P$) , $\mathbf \Lambda$ is $N\times N$ diagonal matrix and $\mathbf W$ is $N \times N$ orthogonal matrix ($\mathbf {W W}^T = \mathbf{W}^T \mathbf{W = I}$; will assume for simplicity that $rank(\mathbf X) = M = N <= P$).  Note that he matrix $\mathbf{W}$ is the same matrix as in oscillator discussion (\ref{2.0.6}).
 Due to the orthogonality of $\mathbf V$ and $\mathbf W$, one has for the two different Gram matrices, introduced in (\ref{0.1.1b}) and  (\ref{0.1a}):

\begin{align}
\mathbf{G' } = \mathbf{X} \mathbf{X}^T = \mathbf{V  \Lambda^2 V}^T, \label{5a}\\
\mathbf{G} =  \mathbf{X}^T \mathbf{X} = \mathbf{W \Lambda^2 W}^T.\label{5b} 
\end{align}

The matrix  $\mathbf W$ (resp.   $\mathbf V$) define  an orthogonal transformation in the training observables space $\mathbf{W}: \mathbb{O} \rightarrow \mathbb{O}$ (resp. in the training observations space $\mathbf{V}: \mathbb{O'} \rightarrow \mathbb{O'}$) - see Section \ref{The training set} for details: 
\begin{align}
\mathbf{W: x_{i}} \rightarrow  \mathbf{y_i} =  \mathbf{\sum_{j=1}^{N} \mathbf{x}_j } W_{ji} \label{5.0a} \\
\mathbf{V: x_{\mu}} \rightarrow   \mathbf{y'_{\mu}} = \mathbf{\sum_{\nu=1}^{P}} V_{\nu \mu} \mathbf {{x}_{\nu} }.  \label{5.0b}
\end{align}
They can be expressed as matrix multiplications if one switched from training observables to observations and vice versa:
\begin{align}
\mathbf{W: x_{\mu}} \rightarrow \mathbf{ y_{\mu}} &=  \mathbf{x_{\mu} W} = \Bigg \{\sum_{j=1}^{N} x_{\mu j}  W_{ji} \label{5.0.0a} \Bigg\}_{i=1...N}\\
\mathbf{V: x_{i}} \rightarrow \mathbf{ y'_{i}} &=  \mathbf{V^T x_{i} } = \Bigg \{\sum_{\nu=1}^{P} V_{\nu \mu}x_{\nu i} , \Bigg\}_{\mu=1...P} \label{5.0.0b} 
\end{align}
or in terms of the  full training matrix $\mathbf X$:
\begin{align}
\mathbf {W: X \rightarrow Y = X W }  \label{5.0.0.0a} \\
\mathbf{V: X \rightarrow Y' = V}^T \mathbf{X },  \label{5.0.0.0b}
\end{align}
where $\mathbf Y$ is $ P \times N$ matrix, $\mathbf Y'$ is $ N \times N$ matrix. The orthogonality of the new basis of observations (resp. observables) in the plain Euclidean metric follows directly from (\ref{5b}) (resp.(\ref{5a})):
\begin{align} 
\mathbf{ y^T_{i} y_{ j} } &=  \mathbf{ \lambda_{i}^2 \delta_{ij }} \Leftrightarrow \mathbf {Y}^T \mathbf{Y = \Lambda}^2 \nonumber   \Leftrightarrow \\ &\Leftrightarrow \mathbf{(XW)}^T \mathbf{(X W)  } =  \mathbf{\Lambda}^2  \\
 \mathbf{ y'}_{\mu } \mathbf{y'}^T_{\nu }  &= \lambda_{\mu}^2 \delta_{\mu \nu}   \Leftrightarrow \mathbf{Y' Y'}^T = \mathbf{\Lambda}^2 \Leftrightarrow  \nonumber \\ & \Leftrightarrow    (\mathbf{V}^T \mathbf{X}) (\mathbf{V}^T \mathbf{X})^T   =  \mathbf{\Lambda}^2   \\  
\mathbf{ y'}_{\mu } (\mathbf{y}_{i })   &=  \lambda_{i} \delta_{\mu i } \Leftrightarrow \mathbf{V}^T \mathbf{X W  } =  \mathbf{\Lambda} .
\label{5.0.1}
\end{align}
The  norms of both sets of new basis vectors are given by the respective singular values. These  basis can be made orthonormal if we add to the transformation a division by the singular values (will assume for simplicity non-zero singularity values):
\begin{align}
\mathbf{ X  \rightarrow Y } &= \mathbf{X (W \Lambda} ^{-1} )   \label{5.02a} \\
\mathbf{ X  \rightarrow Y' } &= \mathbf{  (V \Lambda} ^{ -1}) ^T \mathbf{X}  \label{5.02b}
\end{align}
and then:
\begin{align}
 \mathbf{ y}^T_{i} \mathbf{y_{ j} } &=  \mathbf{\delta_{ij }}  \label{5.0.3a}\\
 \mathbf{ y'_{\mu } y'}^T_{\nu }   &= \mathbf{ \delta_{\mu \nu} } \label{5.0.3b}\\
\mathbf{ y'_{\mu } (y_{i})  } &=  \mathbf{\lambda_{i}}^{-1} \delta_{\mu i} , \label{5.0.3c}
\end{align}
where $\mathbf Y$ is $ P \times N$ matrix, $\mathbf Y'$ is $ N \times N$ matrix. It will be convenient here to introduce also the \emph{''quasi square roots''}  of the covariance matrices i.e. define
\begin{align}
\mathbf{H}  & =\mathbf{\Lambda W}^T  \Rightarrow \mathbf {G = H}^{T}\mathbf{H} \label{5.1a} \\
\mathbf{H' } & =  \mathbf{ V \Lambda } \Rightarrow \mathbf {G' = H' H'}^{T} \label{5.1b}  
\end{align}

In this context, the particular solution of the Dimension reduction problem is easily recognized (\cite{Eckart36}): The n-rank matrix $\mathbf{\hat{X}}$, $n <= N$, which minimizes the Frobenius norm $ Recon~Err =\mathbf{|| X - \hat{X} ||}^2_F$ from (\ref{0.7}): 
\begin{align}
\mathbf{\hat{X}  =  \hat{V} \hat{\Lambda}  \hat{W}}^T  ,
\label{5.1}
\end{align}
where $\mathbf {\hat{W}}$ (resp. $\mathbf {\hat{V}}$) is the $N \times n$ sub-matrix of $\mathbf W$ (resp. $P \times n$ sub-matrix of $\mathbf V$ ) formed by the n columns of $\mathbf W$ (resp. $\mathbf V$) corresponding to the n largest singular values $\{\lambda_i\}_{i=1}^n$. The physics and financial mathematics interpretation of this problem are discussed in Appendix \ref{Dimension reduction with}. Because the orthogonality of $\mathbf{\hat{W}}$ is partially preserved after the truncation i.e. $\mathbf {\hat{W}}^T \mathbf{\hat{W}}$ $ = \mathbf{\hat{I}}$ $=\mathbf{I}_n$, where $\mathbf{I}_n$ is the $n \times n$ identity matrix, this is equivalent to:
\begin{align}
\mathbf{\hat{X}   = X \hat{W} \hat{W}}^T ,
\label{5.1.1}
\end{align}
and in the training observations space:
\begin{align}
\mathbf{\hat{X}  = \hat{V} \hat{V}}^T \mathbf{X  }.
\label{5.2}
\end{align}

The exact general solution for Shallow auto-encoder, discussed in Sub-section \ref{Shallow auto-encoder}, tied weights, and the Dimension reduction problem here is therefore from (\ref{5.1.1}) and (\ref{5.2}): 
\begin{align}
\mathbf{ W}^{(1)}  & = \mathbf{\hat{W} \hat{S}}, \mathbf{ W}^{(2)}  = \mathbf{\hat{S}}^{-1} \mathbf{\hat{W}}^T  \nonumber \\
\mathbf{ V}^{(1)}  & = \mathbf{\hat{S} \hat{V}}^T , \mathbf{ V}^{(2)}   = \mathbf{ \hat{V} \hat{S}}^{-1}
\label{5.3}
\end{align}  
for an arbitrary invertible $n \times n$ transformation  $\mathbf{ \hat{S}}$ which can be called \emph{mixing} matrix. Equivalently,
\begin{align}
\mathbf{ Y}  & = \mathbf{X \hat{W} \hat{S}} ,\quad \mathbf{\hat{X}}   = \mathbf{ Y \hat{S}}^{-1} \mathbf{\hat{W}}^T  \nonumber \\
\mathbf{ Y'}  & = \mathbf{\hat{S} \hat{V}}^T \mathbf{ X} , \quad \mathbf{\hat{X}}   = \mathbf{ \hat{V} \hat{S}^{-1} Y'  }
\label{5.4}
\end{align}
or, in vector terms,
 \begin{align}
\mathbf{ y_\mu}  & = \mathbf{x_{\mu} \hat{W} \hat{S}}, \quad \mathbf{\hat{x}_{\mu}}   = \mathbf{y_{\mu}\hat{S}}^{-1} \mathbf{\hat{W}}^T   \nonumber \\
\mathbf{ y'_i}  & = \mathbf{\hat{S} \hat{V}}^T \mathbf{ x}_i ,\quad  \mathbf{\hat{x}}_i   = \mathbf{ \hat{V} \hat{S}}^{-1} \mathbf{y'}_i  .
\label{5.5}
\end{align}

It  is important to stress that the truncated rectangle  matrices   $\mathbf {\hat{W}}$ and $\mathbf {\hat{V}}$ are only ``quasi-orthogonal'' i.e.
\begin{align}
\mathbf{ \hat{W}}^T \mathbf{\hat{W} = I}_n ,\mathbf{ \quad \hat{V}}^T \mathbf{\hat{V}} = \mathbf{I}_n , \nonumber \\
\mathbf{ \hat{W} \hat{W}}^T \sim \mathbf{I}_N , \mathbf{\quad \hat{V} \hat{V}}^T \sim \mathbf{I }_P,
\label{5.5b}
\end{align}
and the degree of non-orthogonality in the bottom products is proportional to the ``noise'' i.e. $N - n$. One can easily see from (\ref{5.4}) that the  same  quasi-orthogonality carries over to the transition matrices:
\begin{align}
\mathbf{ W}^{(2)} \mathbf{ W}^{(1)} =  \mathbf{ V}^{(1)} \mathbf{ V}^{(2)} = \mathbf{I}_n, \nonumber \\
\mathbf{ W}^{(1)} \mathbf{W}^{(2)} \sim \mathbf{I}_N , \quad  \mathbf{V}^{(2)} \mathbf{V}^{(1)} \sim \mathbf{I }_P.
\label{5.5.1}
\end{align}
Very importantly, the definition (\ref{0.6}) of the transition matrices implies that the magnitude of the discrepancy at the bottom inequalities above is  the ``error''  (the  Frobenius norm (\ref{0.7}))  
$\mathbf{|| X - \hat{X} ||}^2 _F$  between $\mathbf X$ and $\mathbf {\hat{X}},$ in the exact solution. In some sense, through ``mixing'', the error, due to the truncation from $N$ to $n$ dimensions, is spread out across all $N$ dimensions (in the case of observables). As $n$ gets smaller and smaller, the product $\mathbf{ W}^{(1)} \mathbf{W}^{(2)}$ deviates more and more from the identity matrix $\mathbf{I}_N$, both on and off the diagonal. 

\begin{figure}[!ht]
\vskip 0.2in
\begin{center}
\includegraphics[width=\columnwidth]{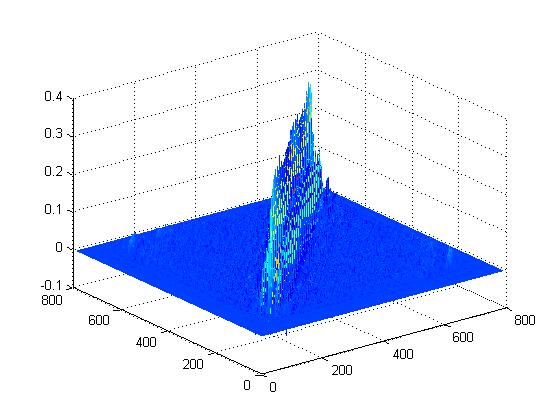}
\vskip 0.2in
\includegraphics[width=\columnwidth]{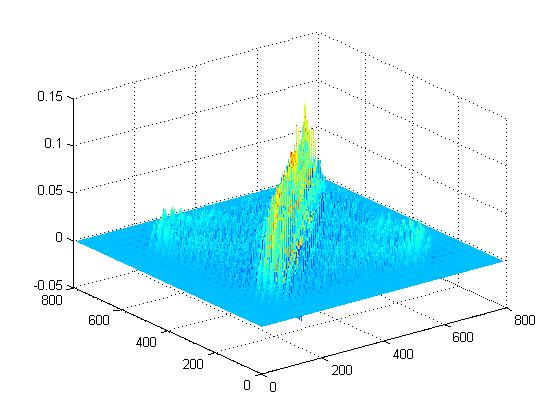}
\caption{The  matrix product $\mathbf{ \hat{W} \hat{W}}^T$ $=\mathbf{ W}^{(1)} \mathbf{W}^{(2)}$ of Dimension reduction mapping in observables space, which is supposed to approximate $\mathbf{I}_N$, for $n=100$ (top) and $n=20$ (bottom), for the first 5,000 MNIST images ($P = 5,000; N=784$). Although the product is quasi-diagonal, the diagonal elements are not even  remotely close to $1$ and even negative elements pop up. They get closer and approach $1$ only for $n > 500$.  Unlike $\mathbf{ V^{(2)} V^{(1)}}$ in the top plot of Figure \ref{Fig4.3} (no Dimension reduction), this ``approximation'' of  $\mathbf{I}_N$ still looks very homogeneous. The sorted diagonal elements are plotted in Figure \ref{Fig4.2}.}
\label{Fig4.1}
\end{center}
\vskip -0.2in
\end{figure}

We show in Figure \ref{Fig4.1} the product $\mathbf{ \hat{W} \hat{W}}^T$ $=\mathbf{ W}^{(1)} \mathbf{W}^{(2)}$ for the MNIST dataset for two cases: $n=100$ and  $n=20$ $\ll N = 784$.

\begin{figure}[!ht]
\vskip 0.2in
\begin{center}
\includegraphics[width=\columnwidth]{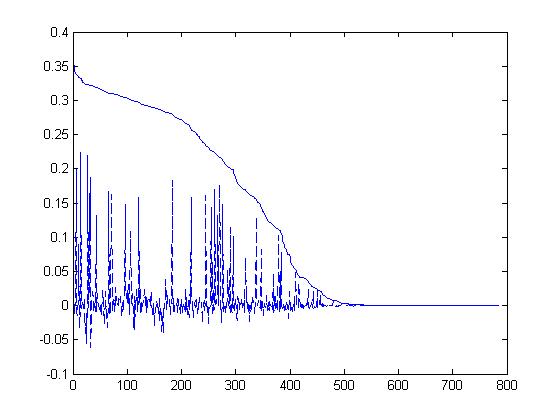}
\vskip 0.2in
\includegraphics[width=\columnwidth]{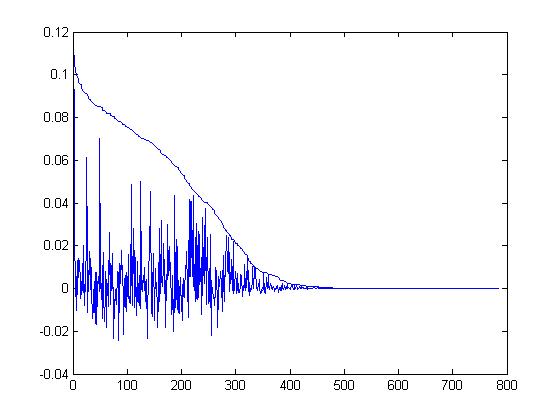}
\caption{The  diagonal (solid line) and first off-diagonal (dashed line) elements of the Dimension reduction mapping  $\mathbf{ \hat{W} \hat{W}}^T$ $=\mathbf{ W^{(1)} W^{(2)}}$  for $n= 100$ (top) and $n=20$ (bottom), sorted in descending order, for the first 5,000 MNIST images (i.e., $P = 5,000$). The pixels with high values on the chart retain the most from the original information. The information carried by the pixels with zeros on the chart, has no connection whatsoever with their original information. }
\label{Fig4.2}
\end{center}
\vskip -0.2in
\end{figure} 

\begin{figure}[!ht]
\vskip 0.2in
\begin{center}
\includegraphics[width=\columnwidth]{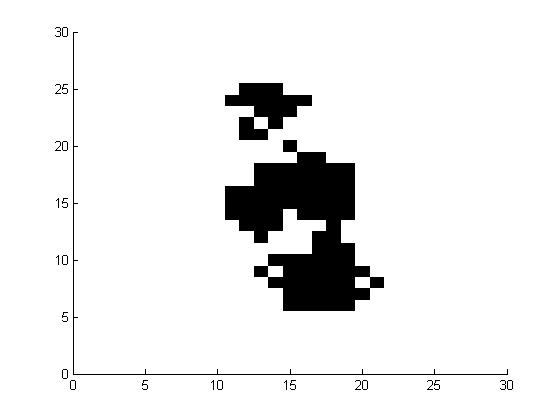}
\vskip 0.2in
\includegraphics[width=\columnwidth]{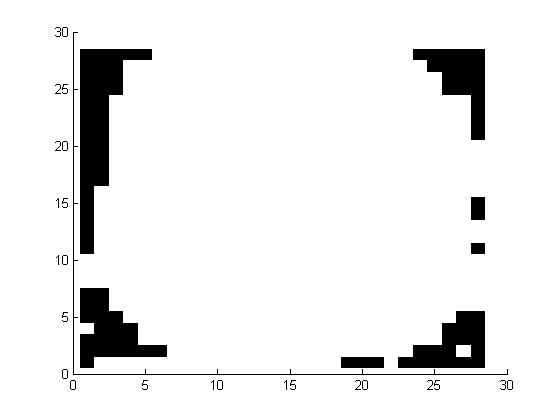}
\caption{The  top 100 (top) and the bottom 100 (bottom) pixels, in the ranking from Figure \ref{Fig4.2}, plotted in black in the standard 28 x 28 MNIST grid ($P = 5,000; n=100$).}
\label{Fig4.2a}
\end{center}
\vskip -0.2in
\end{figure}

Surprisingly, there is an order of magnitude deviation from $\mathbf{I}_N,$ for the diagonal elements. The sorted diagonal elements themselves and the respective off-diagonal elements are plotted in Figure \ref{Fig4.2}. The diagonal elements correspond to the amount of original information retained in a given pixel. The Dimension reduction introduces ``spacial'' noise at \emph{every pixel}: only a third of the information or less (for $n=100$) in a pixel corresponds to its original information, the rest comes from its neighboring pixels. The more the reduction, i.e. the smaller $n$, the greater the spatial spreading of information. Moreover, for a significant number of  pixels on the right-hand side of each graph, the information carried after Dimension reduction has no connection whatsoever with the original information! In fact, Dimension reduction  has wiped out their information entirely. This is qualitatively  different from the space of observations , where every observations retains some resemblance to the original, as it should be , and there are observations which are left completely intact (cf. Figure \ref{Fig4.4})! The top 100 and the bottom 100 pixels, ranked according to their diagonal element in $\mathbf{ \hat{W} \hat{W}}^T$ are plotted in Figure \ref{Fig4.2a}. Naturally, the top pixels are in the middle and very ``busy'' for a typical image, while the bottom ones are on the edges and hardly add any information.

The surprisingly large number of observables (pixels) dropped out by the Dimension reduction in this example is not a general phenomenon but is due  to the inequality between pixels and is an artifact of  the MNIST dataset.\footnote{The statistics of the example look as follows:
\begin{align}
\frac{\mathbf{||\hat{X} ||}^2 _F}{\mathbf{|| X ||}^2 _F} = \frac{\sum_{i=1}^n \lambda^2_i}{\sum_{i=1}^N \lambda^2_i}  \approx
\begin{cases}
\frac{0.1481}{0.1930} \approx 0.76 , \quad n = 20, \\
\frac{0.1790}{0.1930}, \approx 0.93, \quad n = 100, \nonumber 
\end{cases}
\\
\mathbf{|| X - \hat{X} ||}^2 _F  \approx
\begin{cases}
 \approx 0.0432 , \quad n = 20, \\
 \approx 0.0124, \quad n = 100, \\
 \approx 0.00028, \quad n = 500. \nonumber
\end{cases}
\end{align}
}

\begin{figure}[!ht]
\vskip 0.2in
\begin{center}
\includegraphics[width=\columnwidth]{SVD_V_VT.jpg}
\vskip 0.2in
\includegraphics[width=\columnwidth]{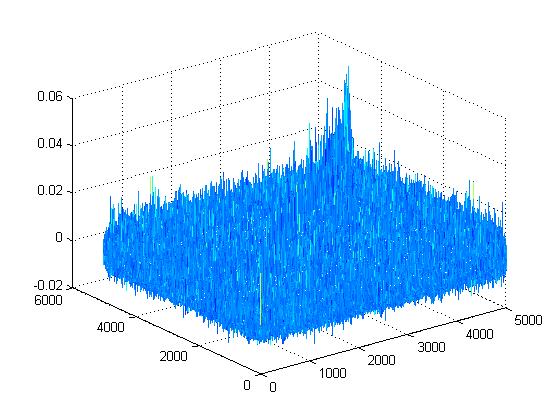}
\caption{The  matrix product $\mathbf{ \hat{V} \hat{V}}^T$ $=\mathbf{ V}^{(2)} \mathbf{V}^{(1)}$ of Dimension reduction (\ref{0.-4a})  induced by $\mathbf X$ itself when $P  \gg N$, which is supposed to approximate $\mathbf{I}_P$, for $n=N = 784$ on the top. At the bottom, we have imposed an additional Dimension reduction in the space of observables i.e. $n=100$. Both use the first 5,000 MNIST images. Because $P =5,000 \gg N$, it does not look close to an identity matrix even for $n=N$ (top). Unlike $\mathbf{ W^{(1)} W^{(2)}}$ in Figure \ref{Fig4.1}, this ``approximation'' of $\mathbf{I}_P$ looks a lot more non-homogeneous, due to the random order of observations in MNIST.  The sorted diagonal and first off-diagonal elements are plotted in Figure \ref{Fig4.4}.}
\label{Fig4.3}
\end{center}
\vskip -0.2in
\end{figure}

\begin{figure}[!ht]
\vskip 0.2in
\begin{center}
\includegraphics[width=\columnwidth]{diagV_VT_sorted.jpg}
\vskip 0.2in
\includegraphics[width=\columnwidth]{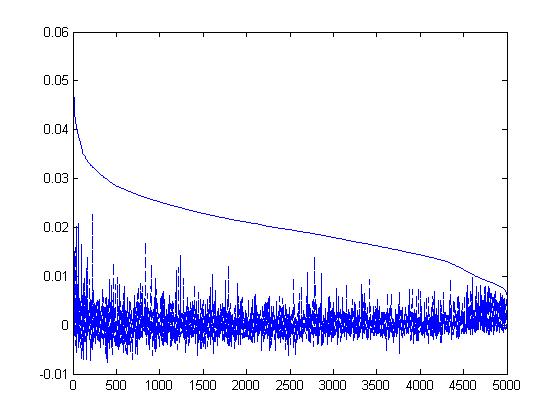}
\caption{The  diagonal (solid line) and first off-diagonal (dashed line) elements of the Dimension reduction mapping $\mathbf{ \hat{V} \hat{V}}^T$ $=\mathbf{ V^{(2)} V^{(1)}}$  for $n=N = 784$ (top) and $n=100$ (bottom), sorted in descending order, for the first 5,000 MNIST images (i.e. $P = 5,000$). The violent drop in the top plot is due to the  Dimension reduction (\ref{0.-4a}) induced by $\mathbf X$ itself when $P  \gg N$. The effect of the additional Dimension reduction in the space of observables, from $N = 784$ to $n=100 \ll N$, is plotted at the bottom and  dampens all values down indiscriminately.}
\label{Fig4.4}
\end{center}
\vskip -0.2in
\end{figure}

\begin{figure}[!ht]
\vskip 0.2in
\begin{center}
\includegraphics[width=\columnwidth]{P1000_K600_diagV_VT_sorted.jpg}
\vskip 0.2in
\includegraphics[width=\columnwidth]{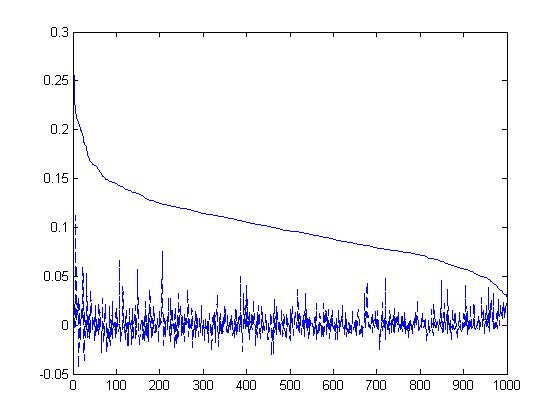}
\caption{Same as Figure \ref{Fig4.4}, but for $P= 1,000 \approx N = 784$ (produced by using only the first 1,000 MNIST images). As in  Figure \ref{Fig4.4}, we have $n=N = 784$ (top) and $n=100$ (bottom). The absence of a violent drop on the top confirms that it emerges only when $P \gg N$ and is due to the Dimension reduction in observations space.}
\label{Fig4.5}
\end{center}
\vskip -0.2in
\end{figure}

In the typical case, when we have many more observations than observables i.e. $P  \gg N$, the first inequality $\mathbf{ W}^{(1)} \mathbf{W}^{(2)} = \mathbf{I}_N$ in (\ref{5.5.1}) will become an equality  when $n=N$ (for the exact solution). Because of the massive Dimension reduction (\ref{0.-4a}) induced by $\mathbf X$ itself when $P  \gg N$ ,  that would not be the case for the second inequality, i.e.
\begin{align}
\mathbf{ W}^{(1)} \mathbf{W}^{(2)} = \mathbf{I}_N ,  \mathbf{V}^{(2)} \mathbf{V}^{(1)} \sim \mathbf{I}_P, \quad P \gg N=n.
\label{5.5.1a}
\end{align}
We plot the matrix $\mathbf{V}^{(2)} \mathbf{V}^{(1)}$ in Figure \ref{Fig4.3} for the same MNIST images. The case $n=N$ is on the top and it is clearly a far cry from $\mathbf{I}_P$. The sorted diagonal elements of the matrix are plotted in Figure \ref{Fig4.4}. The violent drop on the top, for $N=n$, is due to the Dimension reduction (\ref{0.-4a}) from $P=5000$ to $N=784$ in the space of observations.  To confirm that the deviation from identity of the diagonal elements of  $\mathbf{ \hat{V} \hat{V}}^T$ $=\mathbf{ V}^{(2)} \mathbf{V}^{(1)}$ is due to the Dimension reduction (\ref{0.-4a}) induced by $\mathbf X$ itself when $P  \gg N$, we plot them sorted again in Figure \ref{Fig4.5} but this time for $P =1000 \approx N$. Unlike the space of observables, even for arbitrary small $n$,  there are no observations which are wiped out. For $n=N$, there is a small number of observations which are left completely intact by the transformation $\mathbf{V}^{(2)} \mathbf{V}^{(1)}$ (top chart in Figure \ref{Fig4.4}).

For future references, we will compute here  the various ``inner products'' between our data matrices, using the decompositions (\ref{5a}), (\ref{5b}) of $\mathbf {G, G'}$ and (\ref{5.4}). In particular, they give explicit expression for the reduced Gram matrices:

\begin{align}
\hat{\mathbf{G}} = \hat{\mathbf{X}}^T \hat{\mathbf{X}} , \label{5a.1} \\
\hat{\mathbf{G}'} = \hat{\mathbf{X}} \hat{\mathbf{X}}^T. \label{5b.1}
\end{align}

i) For observables:
\begin{align}
\mathbf{ Y}^T \mathbf{Y }  &= \sum_{\mu} \mathbf{y}^T_{\mu}\mathbf{y_{\mu}} = \mathbf{ \hat{S}}^T \mathbf{\hat{W}}^T \mathbf{X}^T \mathbf{X \hat{W} \hat{S}}= \mathbf{\hat{S}}^T \mathbf{\hat{\Lambda}}^2 \mathbf{\hat{S}  },  \nonumber \\
\mathbf{ Y}^T\mathbf{X}   &=  \sum_{\mu} \mathbf{y}^T_{\mu} \mathbf{x_{\mu}} =\mathbf{ \hat{S}}^T \mathbf{\hat{W}}^T \mathbf{X}^T \mathbf{X} =  \mathbf{\hat{S}}^T \mathbf{\hat{\Lambda}}^2 \mathbf{\hat{W}}^T  , \nonumber \\
\mathbf{ Y}^T \mathbf{\hat{X}}   &=  \sum_{\mu} \mathbf{y}^T_{\mu} \mathbf{\hat{x}_{\mu}} = \mathbf{Y}^T \mathbf{Y}  \mathbf{\hat{S}}^{-1}\mathbf{\hat{W}}^T =  \mathbf{\hat{S}}^T \mathbf{\hat{\Lambda}}^2  \mathbf{\hat{W}}^T , \nonumber
\end{align}

\begin{align}
\mathbf{\hat{X}}^T \mathbf{\hat{X}}   &=  \sum_{\mu} \mathbf{\hat{x}}^T_{\mu} \mathbf{\hat{x}_{\mu}} =\mathbf{\hat{W}} \mathbf{\hat{S}}^{-1 T} \mathbf{Y}^T \mathbf{Y}  \mathbf{\hat{S}}^{-1} \mathbf{\hat{W}}^T = \mathbf{\hat{W}} \mathbf{\hat{\Lambda}}^2  \mathbf{\hat{W}}^T  , \nonumber \\
\mathbf{X}^T \mathbf{\hat{X}}   &=  \sum_{\mu} \mathbf{x}^T_{\mu}\mathbf{\hat{x}_{\mu}} =\mathbf{X}^T \mathbf{Y}  \mathbf{\hat{S}}^{-1} \mathbf{\hat{W}}^T \mathbf{\hat{W}} \mathbf{\hat{\Lambda}}^2  \mathbf{\hat{W}}^T , \nonumber \\
\mathbf{ \hat{Y}}^T \mathbf{\hat{X}}   &=  \sum_{\mu} \mathbf{\hat{y}}^T_{\mu} \mathbf{\hat{x}_{\mu}} =\mathbf{ \hat{S}}^T \mathbf{\hat{W}}^T \mathbf{\hat{X}}^T \mathbf{X} =  \mathbf{\hat{S}}^T \mathbf{\hat{\Lambda}}^2 \mathbf{\hat{W}}^T  ,  
\label{5.6}
\end{align}

\smallskip
ii) For observations:
\begin{align}
\mathbf{ Y' Y'}^T   &= \sum_{i} \mathbf{y'}_{i} \mathbf{y'}^T_{i} = \mathbf{ \hat{S}}^T \mathbf{\hat{V}}^T \mathbf{X X}^T \mathbf{\hat{V} \hat{S} }= \mathbf{\hat{S}}^T \mathbf{\hat{\Lambda}}^2 \mathbf{\hat{S}  } , \nonumber \\
\mathbf{ Y' X}^T   &=  \sum_{i} \mathbf{y'}_{i} \mathbf{x}^T_{i} =\mathbf{ \hat{S}}^T \mathbf{\hat{V}}^T \mathbf{X X}^T =  \mathbf{\hat{S}}^T \mathbf{\hat{\Lambda}}^2 \mathbf{\hat{V}}^T  , \nonumber \\
\mathbf{ Y' \hat{X}}^T   &=  \sum_{i} \mathbf{y'}_{i} \mathbf{\hat{x}}^T_{i} =\mathbf{Y' Y'}^T  \mathbf{\hat{S}}^{-1} \mathbf{\hat{V}}^T =  \mathbf{\hat{S}}^T \mathbf{\hat{\Lambda}}^2  \mathbf{\hat{V}}^T  , \nonumber
\end{align}

\begin{align}
\mathbf { \hat{X} \hat{X}}^T   &=  \sum_{i} \mathbf{\hat{x}}_{i} \mathbf{\hat{x}}^T_{i} =\mathbf{\hat{V} \hat{S}}^{-1 } \mathbf{Y' Y'}^T  \mathbf{\hat{S}}^{-1T} \mathbf{\hat{V}}^T = \mathbf{\hat{V}\hat{\Lambda}}^2  \mathbf{\hat{V}}^T  , \nonumber \\
\mathbf{X \hat{X}}^T   &=  \sum_{i} \mathbf{x}_{i} \mathbf{\hat{x}}^T_{i} =\mathbf{X Y'}^T  \mathbf{\hat{S}}^{-1T} \mathbf{\hat{V}}^T = \mathbf{\hat{V}\hat{\Lambda}}^2  \mathbf{\hat{V}}^T    ,\nonumber \\
\mathbf{ \hat{Y}' \hat{X}}^T   &=  \sum_{i} \mathbf{\hat{y}'}_{i} \mathbf{\hat{x}}^T_{i} =\mathbf{ \hat{S}}^T \mathbf{\hat{V}}^T \mathbf{\hat{X} X}^T =  \mathbf{\hat{S}}^T \mathbf{\hat{\Lambda}}^2 \mathbf{\hat{V}}^T .
\label{5.6b}
\end{align}
In particular, when $\mathbf{\hat{S}}$ is diagonal, the column vectors $\{\mathbf y_i\}$ of $\mathbf y$ are orthogonal among themselves and so are the row vectors $\{\mathbf y'_{\mu}\}$ of $\mathbf y'$. As we will see below, the diagonality of $\mathbf{\hat{S}}$ is not a necessary condition for orthogonality of $\{\mathbf y_i\}$.

One can also introduce  $\mathbf{\hat{H}} $,$\mathbf{\hat{H'}} $ as  the ``truncated'' versions of the ``quasi square root''  matrices introduced in (\ref{5.1a}), (\ref{5.1b}) of dimensions $n \times N$ and $P \times n$ respectively:
\begin{align}
\mathbf{\hat{H}} & =\mathbf{\hat{\Lambda} \hat{W}}^T  \label{5.6.1a} \\
\mathbf{\hat{H'} } & =  \mathbf{ \hat{V} \hat{\Lambda} }.  \label{5.6.1b}  
\end{align}
One can check that, in the sense of the approximate ``identities'' (\ref{5.5b}), (\ref{5.5.1}), one has
 \begin{align}
\mathbf{\hat{H}}^T \mathbf{\hat{H} } \sim \mathbf{G}, \qquad \mathbf{\hat{H} \hat{H}}^T = \mathbf{\hat{\Lambda}^2} \label{5.6.1.1a} \\
 \mathbf{ \hat{H'} \hat{H'}}^T  \sim \mathbf{G' }, \qquad  \mathbf{ \hat{H'}}^T \mathbf{\hat{H'} }  = \mathbf{\hat{\Lambda}^2}. \label{5.6.1.1b} 
 \end{align}

\subsection{Special choices for the mixing matrix.}
\label{Special choices}

We will  now go back to the exact linear-algebraic solution and put under the microscope some special choices for the (in general) arbitrary $n \times n$ invertible mixing  matrix $\mathbf { \hat{S}}$ introduced in (\ref{5.3}). For brevity, we will present the choices in the space of observables only: 

\begin{flalign}
&\textbf{Scenario (a)} : \mathbf {\hat{S} = I} ,& \nonumber \\ 
&\textbf{Scenario (b)}: \mathbf {\hat{S} =  \hat{\Lambda}^{-1}} ,&
\label{5.7a}
\end{flalign}
Lets us also consider the same  two choices  but, more generally, multiplied  on the right by an arbitrary orthogonal matrix $\mathbf{\hat{R}}  = \mathbf{(\hat{R}}^T)^{-1}:$
\begin{flalign}
&\textbf{Scenario (c)}:  \mathbf {\hat{S} =   \hat{R}}^T   \Leftrightarrow \mathbf{W}^{(2)} = \mathbf{W}^{(1)T}  , & \nonumber \\
&\textbf{Scenario (d)}: \mathbf {\hat{S}} =  \mathbf{\hat{\Lambda}}^{-1} \mathbf{ \hat{R}}  . &
\label{5.7b}
\end{flalign}
Note that, very importantly, Scenario (c)  is \emph{the most general possible scenario} of tied weights i.e.  $\mathbf{W}^{(2)}$  $= \mathbf{W}^{(1)T}$ in $\mathbb O$ and $\mathbf{V}^{(2)}$  $= \mathbf{V}^{(1)T}$ in $\mathbb O'$. In what follows, the equations for either scenario will carry the resp suffix (a), (b), (c), (d). As above,  we will continue  to treat separately the space of observables and observations.

\bigskip
\bigskip
i) For observables: \\
One gets directly from (\ref{5.3}) for the  resp. scenarios,

\begin{subequations}
\label{5.8}
\begin{align}
\mathbf{W}^{(1)}  &= \mathbf {\hat{W} }  ,  \mathbf{W}^{(2)}  = \mathbf {\hat{W}}^T ,  \\
 \mathbf{W}^{(1)} &= \mathbf{ \hat{W}\hat{\Lambda}}^{-1} = \mathbf{\hat{H}} ^{-1}  ,   \mathbf{W}^{(2)} = \mathbf{ \hat{\Lambda} \hat{W}}^T = \mathbf{\hat{H}  },\\
 \mathbf{W}^{(1)} &= \mathbf{ \hat{W}   \hat{R}} ^T   ,   \mathbf{W}^{(2)}=   \mathbf{\hat{R}\hat{W}}^T, \\
 \mathbf{W}^{(1)} &= \mathbf{ \hat{W}  \hat{\Lambda}}^{-1} \mathbf{\hat{R} } = \mathbf{\hat{H}}^{-1} \mathbf{\hat{ R}} ,   \boxed{\mathbf{W}^{(2)}=   \mathbf{\hat{R}}^T \mathbf{\hat{H}} }
\end{align}
\end{subequations}
and hence,
\begin{subequations}
\label{5.9}
\begin{align}
\mathbf{Y } = \mathbf {X \hat{W}  } = \mathbf { \hat{V} \hat{\Lambda}} = \mathbf{ \hat{H'}   }, \\
 \mathbf{Y} = \mathbf{ X \hat{W}\hat{\Lambda}}^{-1} = \mathbf{ X \hat{H}} ^{-1}   = \mathbf{\hat{V}}, \\
 \mathbf{Y} = \mathbf{ \hat{H'} \hat{R}}^T , \\
   \boxed{ \mathbf{Y} = \mathbf{ \hat{V} \hat{R}} },   
\end{align}
\end{subequations}
(see (\ref{5.6.1b}) for the definition of the quasi square root matrix $\mathbf{\hat{H'} }$ and recall that  $\mathbf {\hat{V}}$ is the $P \times n$ sub-matrix of $\mathbf V$ formed by the n columns of $\mathbf V$ corresponding to the $n$ largest singular values ).  From (\ref{5.6}) one gets the Euclidean inner products of the column vectors in the latent layer:
\begin{subequations} 
\label{5.9.1}
\begin{align}
&\mathbf{ y}^T_i \mathbf{y}_j   = \lambda^2_j \delta_{ij}  , \\
&\mathbf{ y}^T_i \mathbf{y}_j   =  \delta_{ij}  ,  \\
&\mathbf{ y}^T_i \mathbf{y}_j   = \{ \mathbf{\hat{R} \hat{\Lambda}}^2 \mathbf{\hat{R}}^T  \}_{ij} ,  \\
&\boxed{ \mathbf{ y}^T_i \mathbf{y}_j   =  \delta _{ij}} ,
\end{align}
\end{subequations}
where $\delta_{ij} =1$ for i=j, $\delta_{ij} =0$ for $i\neq j$ and $\{\lambda_i\}$ are the diagonal elements of $\mathbf \Lambda$. The equations in Scenario (d) are boxed because they encapsulate the most general scenario of orthonormal observables $\mathbf Y$ in the hidden layer.

For future references, lets compute the self-products of the transition matrices in different scenarios: 
\begin{subequations}
\label{5.10.1a}
\begin{align}
&\mathbf{W}^{(1)T} \mathbf{W}^{(1)}  = \mathbf {I}_n   ,  \mathbf{W}^{(2)} \mathbf{W}^{(2)T} = \mathbf{I}_n,  \\
&\mathbf{W}^{(1)T} \mathbf{W}^{(1)}  = \mathbf {\hat{\Lambda}}^{-2}  ,   \mathbf{W}^{(2)} \mathbf{W}^{(2)T} = \mathbf {\hat{\Lambda}}^2 ,\\
&\mathbf{W}^{(1)T} \mathbf{W}^{(1)}  = \mathbf {I}_n   ,  \mathbf{W}^{(2)} \mathbf{W}^{(2)T} = \mathbf{I}_n, \\
& \mathbf{W}^{(1)T} \mathbf{W}^{(1)} = \mathbf{\hat{R}}^T \mathbf{\hat{\Lambda}}^{-2} \mathbf{\hat{R}  }  ,   \mathbf{W}^{(2)} \mathbf{W}^{(2)T}  = \mathbf{\hat{R}}^T \mathbf{\hat{\Lambda}}^2 \mathbf{\hat{R}  }
\end{align}
\end{subequations}
Similarly, for the permuted self-products:
\begin{subequations}
\label{5.10.1}
\begin{align}
&\mathbf{W}^{(1)} \mathbf{W}^{(1)T}  \sim \mathbf {I}_N   ,  \mathbf{W}^{(2)T} \mathbf{W}^{(2)} \sim \mathbf{I}_N,  \\
&\mathbf{W}^{(1)} \mathbf{W}^{(1)T}  \sim \mathbf {G}^{-1}  ,   \mathbf{W}^{(2)T} \mathbf{W}^{(2)} \sim \mathbf {G} ,\\
&\mathbf{W}^{(1)} \mathbf{W}^{(1)T}  \sim \mathbf {I}_N   ,  \mathbf{W}^{(2)T} \mathbf{W}^{(2)} \sim \mathbf{I}_N, \\
& \boxed{ \mathbf{W}^{(1)} \mathbf{W}^{(1)T} \sim \mathbf {G}^{-1}  ,   \mathbf{W}^{(2)T} \mathbf{W}^{(2)} \sim \mathbf{G}   },
\end{align}
\end{subequations}
The ``approximate'' sign ``$\sim$'' means ``similar in the sense of (\ref{5.5.1})'' (recall the''quasi-square root'' approximate decomposition  (\ref{5.6.1.1a}) of $\mathbf {G}$).

\bigskip
\bigskip
ii) For observations: \\
Similarly to above, from (\ref{5.3}) ,

\begin{subequations}
\label{5.8b}
\begin{align}
&\mathbf{V}^{(1)}  = \mathbf {\hat{V}}^T   ,  \mathbf{V}^{(2)}  = \mathbf {\hat{V} },  \\
& \mathbf{V}^{(1)} = \mathbf{ \hat{\Lambda}}^{-1} \mathbf{\hat{V}}^T = \mathbf{\hat{H'}} ^{-1 } ,  \mathbf{V}^{(2)} = \mathbf{  \hat{V} \hat{\Lambda} = \hat{H'}  }, \\
&\boxed{ \mathbf{V}^{(1)} = \mathbf{  \hat{R}}^T \mathbf{\hat{V}}^T  ,  \mathbf{V}^{(2)} = \mathbf{  \hat{V} \hat{R}  } }, \\
&\mathbf{V}^{(1)} = \mathbf{\hat{\Lambda}}^{-1}  \mathbf{\hat{R} \hat{V}}^T,  \mathbf{V}^{(2)} = \mathbf{  \hat{V} \hat{R}}^T  \mathbf{\hat{\Lambda}},
\end{align}
\end{subequations}
and hence,
\begin{subequations}
\label{5.9b}
\begin{align}
&\mathbf{Y' } = \mathbf {\hat{V}}^T \mathbf{X  } = \mathbf { \hat{\Lambda} \hat{W}}^T  = \mathbf{ \hat{H}   } ,\\
&\mathbf{Y'} = \mathbf{\hat{\Lambda}}^{-1} \mathbf{\hat{V}}^T \mathbf{X } = \mathbf{ \hat{H'}} ^{-1} \mathbf{X} = \mathbf{\hat{W}}^T , \\
&\boxed{\mathbf{Y'} = \mathbf{\hat{R}}^T \mathbf{\hat{V}}^T \mathbf{X } = \mathbf{ \hat{R}}^T \mathbf{\hat{H}}} ,\\
&\mathbf{Y'} = \mathbf{ \hat{\Lambda}}^{-1} \mathbf{\hat{R} \hat{V}}^T \mathbf{X }  .
\end{align}
\end{subequations}
 From (\ref{5.6b}) , one gets the Euclidean inner products of the column vectors in the hidden layer:
\begin{subequations} 
\label{5.10b}
\begin{align}
&\mathbf{ y'}_{\mu} \mathbf{y'}^T_{\nu}   = \lambda^2_{\nu} \delta_{\mu \nu}  ,   \\
&\mathbf{ y'}_{\mu} \mathbf{y'}^T_{\nu}   =  \delta_{\mu \nu}  , \\
&\mathbf{ y'}_{\mu} \mathbf{y'}^T_{\nu}   =\{ \mathbf{\hat{R}}^T \mathbf{\hat{\Lambda}}^2 \mathbf{\hat{R} } \}_{\mu \nu}  ,   \\
&\mathbf{ y'}_{\mu} \mathbf{y'}^T_{\nu}   = \{  \mathbf{ \hat{\Lambda}}^{-1} \mathbf{\hat{R} \hat{\Lambda}}^2 \mathbf{\hat{R}}^T  \mathbf{\hat{\Lambda}}^{-1} \}_{\mu \nu}   .
\end{align}
\end{subequations}

For future references, let us compute the self-products of the transition matrices in different scenarios. 
\begin{subequations}
\label{5.10b.1}
\begin{align}
&\mathbf{V}^{(1)} \mathbf{V}^{(1)T}  = \mathbf {I}_n   ,  \mathbf{V}^{(2)T} \mathbf{V}^{(2)} = \mathbf{I}_n,  \\
&\mathbf{V}^{(1)} \mathbf{V}^{(1)T} = \mathbf {\hat{\Lambda}}^{-2}  ,   \mathbf{V}^{(2)T} \mathbf{V}^{(2)} = \mathbf {\hat{\Lambda}}^2 ,\\
&\boxed { \mathbf{V}^{(1)} \mathbf{V}^{(1)T}  = \mathbf {I}_n  ,  \mathbf{V}^{(2)T} \mathbf{V}^{(2)} = \mathbf{I}_n }, \\
&\mathbf{V}^{(1)} \mathbf{V}^{(1)T}= \mathbf {\hat{\Lambda}}^{-2} ,   \mathbf{V}^{(2)T} \mathbf{V}^{(2)} =\mathbf {\hat{\Lambda}}^2 ,
\end{align}
\end{subequations}
(recall the''quasi-square root'' decomposition  (\ref{5.6.1.1b}) of $\mathbf {\hat{G'}}$). Similarly, for the permuted self-products: 
\begin{subequations}
\label{5.10b.2}
\begin{align}
&\mathbf{V}^{(1)T} \mathbf{V}^{(1)}  \sim \mathbf {I}_P   ,  \mathbf{V}^{(2)} \mathbf{V}^{(2)T} \sim \mathbf{I}_P,  \\
&\mathbf{V}^{(1)T} \mathbf{V}^{(1)}  \sim \mathbf {G'}^{-1}  ,   \mathbf{V}^{(2)} \mathbf{V}^{(2)T} \sim \mathbf {G'} ,\\
&\boxed{ \mathbf{V}^{(1)T} \mathbf{V}^{(1)}  \sim \mathbf {I}_P   ,  \mathbf{V}^{(2)} \mathbf{V}^{(2)T} \sim \mathbf{I}_P }, \\
&\mathbf{V}^{(1)T} \mathbf{V}^{(1)} \sim  \mathbf{\hat{V} \hat{R} }^T  \mathbf{\hat{\Lambda}}^{-2} \mathbf{\hat{R} \hat{V}}^T   ,  \nonumber \\ 
&\mathbf{V}^{(2)} \mathbf{V}^{(2)T}  \sim  \mathbf{\hat{V} \hat{R}}^T  \mathbf{\hat{\Lambda}}^2 \mathbf{\hat{R} \hat{V}}^T .
\end{align}
\end{subequations}
The ``approximate'' sign ``$\sim$''  means ``similar in the sense of (\ref{5.5.1})'' (recall the''quasi-square root'' decomposition  (\ref{5.6.1.1b}) of $\mathbf {\hat{G'}}$).

\subsection{Duality and orthogonality.}
\label{Orthogonality in}

A comparison between (\ref{5.9}d) and (\ref{5.8b}c), on the one hand, and (\ref{5.8}d) and (\ref{5.9b}c) on the other, reveals a nice \emph{duality} between the hidden nodes in one of the two training spaces ($\mathbb O$ or $\mathbb O'$) and the decoding matrix, for Scenarios $\mathbf{(c)}$ and $\mathbf{(d)}$ (the dualty for Sub-scenarios $\mathbf{(a)}$ and $\mathbf{(b)}$ is listed below, in (\ref{5.10d})):
 \begin{align}
\mathbf{Y } ~ from ~\mathbf{(d})~ in ~\mathbb{O} &=   \mathbf { V}^{(2)} ~ from ~ \mathbf{(c)} ~in~ \mathbb{O'}, \nonumber \\
\mathbf{W}^{(2)}  ~ from ~\mathbf{(d})~ in ~\mathbb{O} &=   \mathbf { Y' } ~ from ~ \mathbf{(c)} ~in~ \mathbb{O'}.
 \label{5.10c}
\end{align}
 In other words, when the Dimension reduction problem is exactly solved, the decoding transition matrix $\mathbf{W}^{(2)}$  in the space of observables for Scenario $\mathbf{(d)}$ coincides with the observations  $\mathbf{Y'}$ in the hidden layer in the space of observations for Scenario $\mathbf{(c)}$. But as already emphasized after the definition (\ref{5.7b}), and as the boxed formulas above for $\mathbf{V}^{(2)}$ confirm, Scenario $\mathbf{(c)}$ is always true, as long the observation weights are tied i.e.  $\mathbf{V}^{(2)}$  $= \mathbf{V}^{(1)T}$ in $\mathbb O'$. 
 
On the other hand, as seen in (\ref{5.9.1}d), the latent observables  $\mathbf{Y}$ in Scenario $\mathbf{(d)}$ are orthonormal. This is a highly-nontrivial and often desirable feature which is hard to achieve organically in   numerical algorithms. The orthogonalization algorithm is in Algorithm~\ref{Othogonalization algorithm}.

\begin{algorithm}[!ht]
   \caption{Othogonalization algorithm for observables (in $\mathbb{O}$):}
   \label{Othogonalization algorithm}
\begin{algorithmic}
   \STATE {\bfseries Input:} $\mathbf{X} $
   \smallskip
   \STATE {\bfseries Initialize:}  
   \STATE ~ i) weights: 
   \STATE In $\mathbb O': \mathbf{V}^{(2)}(0)= \mathbf{V}^{(1)T}(0)$ random   , 
   \STATE In  $ \mathbb O: \mathbf {W}^{(2)}(0) = \mathbf {Y'}(0), \quad \mathbf {W}^{(1)}$ not used  
    \smallskip
   \STATE ~ ii) layers: \\
   \STATE In $\mathbb O'$: $\mathbf{ X}(0) = \mathbf{X} $, \\
   \STATE In  $ \mathbb O$: $\mathbf{ Y}(0)$ $=\mathbf V^{(2)}(0)$ 
   
   \REPEAT
         
   \FOR{$m=1$ {\bfseries to} $M$}
   {
   \STATE   In $\mathbb O':$ back-propagation,
   \STATE In $\mathbb O: \mathbf {W}^{(2)}(m) =  \mathbf {W}^{(2)}(m) + \Delta \mathbf {Y'}(m) $
   $\mathbf {Y}(m) = \mathbf {Y}(m) + \Delta \mathbf {V}^{(2)}(m) $ 
   }
   \ENDFOR
   \UNTIL \\{  In $\mathbb O': \mathbf{V}^{(1)} \mathbf{V}^{(2)}= \mathbf{V}^{(2)T} \mathbf{V}^{(2)} = \mathbf{\hat{I}_n} \quad \Leftrightarrow $\\
   In $\mathbb O: \mathbf {Y}^T \mathbf{Y} = \mathbf{\hat{I}}_n$   }
\end{algorithmic}
\end{algorithm}

For completeness, we also list below the duality between Scenarios (a) and (b):
  \begin{align}
\mathbf{Y } ~ from ~\mathbf{(a})~ in ~\mathbb{O} &=   \mathbf { \hat{V}}^{(2)}  ~ from ~ \mathbf{(b)} ~in~ \mathbb{O'}, \nonumber \\
\mathbf{Y } ~ from ~\mathbf{(b})~ in ~\mathbb{O} &=   \mathbf { \hat{V}}^{(2)}  ~ from ~ \mathbf{(a)} ~in~ \mathbb{O'}, \nonumber \\
\mathbf{Y' } ~ from ~\mathbf{(a})~ in ~\mathbb{O} &=   \mathbf { \hat{W}}^{(2)}  ~ from ~ \mathbf{(b)} ~in~ \mathbb{O'}, \nonumber \\
\mathbf{Y' } ~ from ~\mathbf{(b})~ in ~\mathbb{O} &=   \mathbf { \hat{W}}^{(2)}  ~ from ~ \mathbf{(a)} ~in~ \mathbb{O'}.
 \label{5.10d}
\end{align}

Alternatively, if the goal is to achieve orthogonality for the hidden observations in $\mathbb O'$, a ``dual'' to the above algorithm has to be followed, with  Scenario (d) replaced by:
  
\bigskip
\textbf{Scenario (d')}: $\mathbf {\hat{S} =   \hat{R} \hat{\Lambda}}^{-1} $ .

\section{Numerical optimization.}
\label{Numerical optimization}
In practical applications, the dimensions of the training matrix $\mathbf X$ could be in the thousands or millions, so, exact algebraic solutions as above are not really feasible. The transition matrices will likely be determined iteratively and thus ``evolve'' in \emph{optimization} time. 

\subsection{Target identities for numerical optimization.}
\label{Target identities}

As a prelude to the numerical estimation methods for $\mathbf{W^{(1)}}$ and  $\mathbf{W^{(2)}},$ let us highlight some interesting phenomena emerging in the exact solution (\ref{5.6}), (\ref{5.6b}) \footnote{For brevity, we will focus on observables only, in the absence of random sampling.}. The observables in the hidden layer i.e. the column vectors of $\mathbf Y$  are orthogonal to the reconstruction error $\mathbf{ \hat{X} - X}$ because, according to the second and third of (\ref{5.6}), one has:
\begin{align}
\mathbf{ Y^T (\hat{X} - X) }= \mathbf{ Y^T \Delta \mathbf{X}  } = 0,
\label{5.6c}
\end{align}
where
\begin{align}
\Delta \mathbf{X} := \mathbf{\hat{X} - X}.
\label{5.6c.1}
\end{align}
Intuitively, if the solution is exact, the error is confined to the subspace spanned by the N - n eigenvectors of $\mathbf{ X^T X}$ corresponding to the N - n smallest eigenvalues (and similarly for observations). 

\subsubsection{Case of general un-tied weights.}

In an iterative numerical scheme, the reconstruction error $\mathbf{ X - \hat{X}}$ at iterative step $m$\footnote{Note that the iterative step $m$  in a numerical optimization procedure is not to be confused with the iterative step $j$ in the tied layer nets (Sub-section \ref{Core architecture}).}  is given by  (\ref{5.6c.1}) and Figure \ref{Fig5.1}:
\begin{align}
\Delta \mathbf{X}(m) =\mathbf{X}(m) - \mathbf{X}(0)
\label{5.6c.2}
\end{align}
and the above exact solution can be rewritten at step $m$ as:
\begin{align}
\mathbf{ Y}^T(m) \Delta \mathbf{X}(m)   = 0.
\label{5.6d}
\end{align}

The minimization of this inner products  via incremental changes of $\mathbf{W}^{(1)}  $ and $\mathbf{W}^{(2)}  $ is the objective of   the numerical method: at iteration  $m$, $\mathbf{W}^{(1)}  $ is assumed known  and $\mathbf {W}^{(2)} $ is adjusted by  an incremental amount $\Delta \mathbf {W}^{(2)}(m)$ $= \mathbf {W}^{(2)}(m+1) - \mathbf {W}^{(2)}(m),$ so as to minimize the left hand side of (\ref{5.6d}) i.e.
\begin{align}
\Delta \mathbf {W}^{(2)}(m) = -\delta \mathbf{  Y}^T(m) \Delta \mathbf{X}(m)   
\label{5.6e}
\end{align}
for a small ``learning parameter'' $\delta$. To complete the iterative step, one then assumes  $\mathbf{W}^{(2)}  $ fixed and adjusts $\mathbf {W}^{(1)} $.

\subsubsection{Case of tied weights.}
In this case, we can not optimize the two weight matrices independently. Instead, we want to construct an identity similar to (\ref{5.6c}) which involves both the reconstruction error  $ \Delta \mathbf{X} := \mathbf{\hat{X} - X}$ and the ``propagated'' reconstruction error in the latent layer $ \Delta \mathbf{Y} := \mathbf{\hat{Y} - Y}$ $= \mathbf{\hat{X}\mathbf{W}^{(1)}   - Y}$. To accomplish this, note that we can reduce the sum of suitable inner products of the two errors  to the difference of $\mathbf{ X^T Y}$ and  $\mathbf{ \hat{X}^T \hat{Y}}$, which  are equal in the exact solution (\ref{5.6}): 
\begin{align}
&\Delta \mathbf{X} ^T \mathbf{ Y}  + \mathbf{ \hat{X}}^T \Delta \mathbf{Y } = \nonumber \\
&=(\mathbf{\hat{X}} ^T -\mathbf{X}^T)\mathbf{ Y}  + \mathbf{ \hat{X}}^T \mathbf{(\hat{Y} - Y) } = \nonumber \\
&= - \mathbf{ X}^T \mathbf{Y} + \mathbf{\hat{X}}^T \mathbf{\hat{Y}}  = 0,
\label{5.6f}
\end{align}
or, after expanding the left-hand side:
\begin{align}
&(-  \mathbf{X}^T \mathbf{X} + \mathbf{\hat{X}}^T \mathbf{\hat{X}} ) \mathbf{W}^{(1)} = \nonumber \\
&=  (\mathbf{ - G + \hat{G}}) \mathbf{W}^{(1)} = \nonumber \\
&= \Delta \mathbf{ G} \mathbf{W}^{(1)} = 0,
\label{5.6f.1}
\end{align}
for $\Delta \mathbf{G}$ $=\mathbf{\hat{G}} - \mathbf{G}$. These identities can be re-written using an iteration index $m$, instead of the hat symbol,
\begin{align}
& \Delta\mathbf{X}^T(m) \mathbf{Y}(m ) + \mathbf{ X}^T(m) \Delta \mathbf{Y}(m)  = \nonumber \\
& (\mathbf{X}^T(m) - \mathbf{X}^T(0)) \mathbf{Y}(m ) + \nonumber \\
&+ \mathbf{ X}^T(m) ( \mathbf{X}(m) \mathbf{W}^{(1)}(m)- \mathbf{Y}(m))  = \nonumber \\
& = \mathbf{ X}^T(m) \mathbf{X}(m)\mathbf{W}^{(1)}(m) - \mathbf{X}^T(0) \mathbf{Y}(m)  = 0,
\label{5.6g}
\end{align}
which in expanded form renders:
\begin{align}
&\left(   \mathbf{ X}^T(m) \mathbf{X}(m) - \mathbf{X}^T(0) \mathbf{X}(0) \right) \mathbf{W}^{(1)}(m) = \nonumber \\
 &= (\mathbf{G}(m) - \mathbf{G}(0) ) \mathbf{W}^{(1)}(m) = \nonumber \\
 &=\Delta \mathbf{G}(m) \mathbf{W}^{(1)}(m)  = 0,
\label{5.6g.1}
\end{align}
for $\Delta \mathbf{G}(m) $ $=\mathbf{G}(m) - \mathbf{G}(0)$. This is the weight update term for the Restricted Boltzmann Machine (cf. (\ref{5.11i})).

\subsubsection{Case of incremental tied weights.}
Assume that instead of (\ref{5.6c.2}), the reconstruction error $\Delta \mathbf{X}$ at step $m,$ is given by an incremental change $\Delta \mathbf{X}(m) $ $=\mathbf{X}(m+1) - \mathbf{X}(m)$ and $\mathbf{Y}(m) = \mathbf{X}(m) \mathbf{W}^{(1)}(m).$ In equlibrium, the iteration steps $m$ and $m+1$
are approximately equal, therefore:
\begin{align}
& \Delta\mathbf{X}^T(m) \mathbf{Y}(m ) + \mathbf{ X}^T(m+1) \Delta \mathbf{Y}(m)  = \nonumber \\
& (\mathbf{X}^T(m+1) - \mathbf{X}^T(m)) \mathbf{Y}(m ) + \nonumber \\
& +\mathbf{ X}^T(m+1) ( \mathbf{Y}(m+1) - \mathbf{Y}(m))  = \nonumber \\
& = - \mathbf{X}^T(m) \mathbf{Y}(m) + \mathbf{ X}^T(m+1) \mathbf{Y}(m+1)  \approx 0,
\label{5.6g.2}
\end{align}
or, in expanded form, ignoring the higher order difference between $\mathbf{W}^{(1)}(m)$ and $\mathbf{W}^{(1)}(m+1)$:
\begin{align}
&\left(  - \mathbf{ X}^T(m) \mathbf{X}(m) + \mathbf{X}^T(m+1) \mathbf{X}(m+1) \right) \mathbf{W}^{(1)}(m) = \nonumber \\
 &= ( - \mathbf{G}(m) + \mathbf{G}(m+1)) \mathbf{W}^{(1)}(m) = \nonumber \\
 &=  \Delta \mathbf{G}(m) \mathbf{ W}^{(1)}(m)  \approx 0,
\label{5.6g.3}
\end{align}
where we changed the definition of $\Delta \mathbf{G}(m) $ to be instead the incremental change $\mathbf{G}(m+1) - \mathbf{G}(m)$.

In each of the tied weight cases,  $\mathbf {W}^{(1)} $  changes in iteration step $m$ with an incremental amount $\Delta \mathbf {W}^{(1)}(m)$ $= \mathbf {W}^{(1)}(m+1) - \mathbf {W}^{(1)}(m)$ so as to minimize (\ref{5.6g.1}) or (\ref{5.6g.3}) i.e.
\begin{align}
\Delta \mathbf {W}^{(1)}(m) = -\delta  \Delta \mathbf{G}(m) \mathbf{W}^{(1)}(m)   
\label{5.6h}
\end{align}
  for a small ``learning parameter'' $\delta$.

\subsection{Minimizing reconstruction error: back-propagation from Variational Calculus.}
\label{Minimizing reconstruction}

\begin{figure}[!ht]
\vskip 0.2in
\begin{center}
\begin{tikzpicture}
\centering
\tikzstyle{ann} = [draw=none,fill=none]
\node[rectangle, text width = 0.3\columnwidth, text  centered,  draw ] 
	(Input_t) at (5,0) { \underline{\textbf{1.Input Layer}}: \\ $\mathbf{X}(0)$ };
\node[rectangle,  text width = 0.3\columnwidth, text  centered, below=of Input_t, draw] 
	(Hidden_t)  { \underline{\textbf{2.Latent Layer}}: \\$\mathbf{Y}(m)$ };
\node[rectangle, text width =0.3\columnwidth, text  centered, below=of Hidden_t, draw ] 
	(Output_t)  {\underline{\textbf{3.Output Layer}}: \\ $\mathbf{X}(m)$ };

\node[ann ,text width = 0.3\columnwidth, text  centered,  above=0.5cm of Input_t  ] (t) { Step m} ;
\path [->](Input_t) edge node {$\mathbf{W}^{(1)}(m)$} (Hidden_t);
\path [->](Hidden_t) edge node {$\mathbf{W}^{(2)}(m)$} (Output_t);

\node[rectangle, text width = 0.3\columnwidth,text  centered,  right=1in of Input_t ,draw ] 
	(Input_t1) at (5,0) {\underline{\textbf{1.Input Layer}}: \\ $\mathbf{X}(0)$ };
\node[rectangle,  text width = 0.3\columnwidth, text  centered, below=of Input_t1, draw] 
	(Hidden_t1)  { \underline{\textbf{2.Latent Layer}}: \\$\mathbf{Y}(m+1)$ };
\node[rectangle, text width = 0.3\columnwidth,text  centered,,below=of Hidden_t1, draw ] (Output_t1)  { \underline{\textbf{3.Output Layer}}: \\$\mathbf{X}(m+1)$ };
\node[ann ,text width = 0.3\columnwidth, text  centered,  above=0.5cm of Input_t1  ] (t1) { Step m+1} ;

\path [->](Input_t1) edge node {$\mathbf{W}^{(1)}(m+1)$} (Hidden_t1);
\path [->](Hidden_t1) edge node {$\mathbf{W}^{(2)}(m+1)$} (Output_t1);
\path [->](t) edge node [above] {} (t1);

\node[rectangle, text width = 0.3\columnwidth, text  centered,  right=2.5in of Input_t ] 
	(Input_t2) at (5,0) { };
\node[ann ,text width = 0.3\columnwidth,text  centered,  above=1cm of Input_t2  ] (t2) {} ;

\path [->](t1) edge node  [above] {}(t2);
\end{tikzpicture}
\caption{Shallow  auto-encoder in optimization time.}
\label{Fig5.1}
\end{center}
\vskip -0.2in
\end{figure}
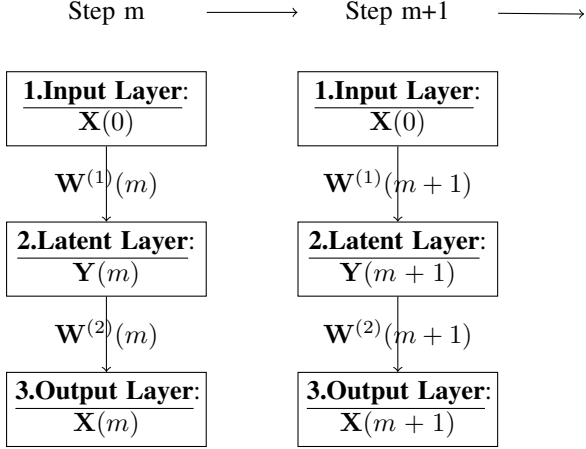

Let us introduce an iteration step index a.k.a. an optimization time index $m = 0, 1,2,...$ and rewrite the decoding and encoding stage of the Shallow auto-encoder from Section \ref{Neural network}, using the optimization index (and Figure \ref{Fig5.1}):
\begin{align}
\mathbf {X} &= \mathbf{X}(0), \nonumber \\
 \mathbf {Y} &= \mathbf{Y}(m) = \mathbf{X}(0) \mathbf{W}^{(1)}(m) \nonumber \\
\mathbf {\hat{X}} &= \mathbf{X}(m) = \mathbf{Y}(m)  \mathbf{W}^{(2)}(m) , \nonumber \\
 \mathbf {\hat{G}} &= \mathbf {X^T \hat{X}}  = \mathbf{G}(m).
\label{5.10}
\end{align}

In back-propagation, one starts typically with random weights/biases and the goal is to decrease at every step the reconstruction error $Recon~Err(m)$  from  (\ref{0.7}):
\begin{align}
Recon~Err(m) =  Tr\left\{\Delta \mathbf{X}(m)  \Delta \mathbf{X}(m)^T\right\}  
\label{5.11}
\end{align}
where $\Delta \mathbf{X}(m) = \mathbf{X}(m) - \mathbf{X}(0)$. This is accomplished by modifying $\mathbf{W}^{(1)}(m)$ and $ \mathbf{W}^{(2)}(m)$ at every time step and hopefully reaching an equilibrium after sufficiently many steps $M$. 

\subsubsection{Case of general un-tied weights.}

One can describe the minimization of   $Recon~Err$ (\ref{5.11}), subject to the constraints (\ref{5.10}), as the minimization of the \emph{action} functional $\mathcal{S} (\mathbf{X,W,Z})$:
\begin{align}
 \mathcal{S} (\mathbf{X,W,Z}) &= Tr \sum_{m=1}^M \left \{ 
 \frac{1}{2}\Delta \mathbf{X}(m)  \Delta \mathbf{X}(m)^T \right.+ \nonumber \\
  &+  \mathbf{\hat{Z}} (m)\left(\mathbf{X}(m) - \mathbf{Y}(m)  \mathbf{W}^{(2)}(m)\right)^T + \nonumber \\
  &+ \left.  \mathbf{Z} (m)\left(\mathbf{Y}(m) - \mathbf{X}(0)  \mathbf{W}^{(1)}(m)\right)^T \right \}, 
  \label{5.11b}
\end{align}
for some, yet to be determined, \emph{Lagrangian coefficients} $\mathbf{\hat{Z}}(m) \in \mathbb{R}^{ P \times N}$, $ \mathbf{Z}(m)\in \mathbb{R}^{ P \times n}$, $m =0,1,2,...,M.$ Note that the Trace is taken on matrices of size $P \times P$ i.e. operators in the space of observations. In the continuous limit, the sum $\Sigma _m$ becomes an integral in the traditional sense of variational calculus, \cite{Gelfand63} (see in particular Appendix II, where optimal control problem is discussed).

The stationary solutions are found by zeroing the partial derivatives of our action with respect to the ``operators'' $\mathbf{X,W,Z}$  and their ``derivatives''. But our functional does not have explicit finite differences which are the discrete analogue of derivatives ($\Delta \mathbf{X}$ is not a finite difference in time). Taking partial derivatives of the action\footnote{We use  standard matrix calculus  to take derivatives of Tr() of matrix products.}, one easily gets for the stationary solution:
\begin{align}
 \frac { \partial \mathcal{S}}{ \partial \mathbf{X}(m)} &= \mathbf{\hat{Z}} (m) + (\mathbf{X}(m) - \mathbf{X}(0)) = 0, \nonumber \\
\frac { \partial \mathcal{S}}{ \partial \mathbf{Y}(m)} &= \mathbf{Z} (m) -   \mathbf{\hat{Z}} (m) \mathbf{W}^{(2)}(m)^T= 0,  \nonumber \\
\frac { \partial \mathcal{S}}{ \partial \mathbf{W}^{(2)}(m)} &= -  \mathbf{Y}(m)^T \mathbf{\hat{Z}} (m)= 0 , \nonumber \\
\frac { \partial \mathcal{S}}{ \partial \mathbf{W}^{(1)}(m)} &= -  \mathbf{X}(0)^T \mathbf{Z} (m) = 0. 
\label{5.11c}
\end{align}
Note that, contrary to (\ref{5.11b}), where we have operators in the space of observations, the partial derivatives with respect to $\mathbf{W}^{(1)}(m)$ and $ \mathbf{W}^{(2)}(m)$ are  operators in the space of observables. From the first equation, the Lagranian vector-coefficient $-\Delta \mathbf{X}(m)=\mathbf{\hat{Z}}(m)$ is  the negative reconstruction error vector (whose norm we are seeking to minimize) at the output layer, for every $m$. From the second equation:
\begin{align}
\mathbf{Z} (m) =   \mathbf{\hat{Z}} (m)  \mathbf{W}^{(2)}(m)^T, 
\label{5.11d}
\end{align}
hence, we can interpret  the Lagrangian coefficients $\mathbf{Z}(m),$  $m < M, $ as the ``back-propagated'' reconstruction error, using for back-propagation the transpose transition operator. It is important to stress that this is NOT a back-propagarion in time but merely in the layers of our net! From the last two equation in (\ref{5.11c}), one deduces the updating equations for the weight matrices at step $m$:
\begin{align}
\Delta \mathbf{W}^{(2)}(m) = -\delta  \mathbf{Y}(m)^T \Delta \mathbf{X} (m) \nonumber \\
\Delta \mathbf{W}^{(1)}(m) = -\delta \mathbf{X}(0)^T  \mathbf{Z} (m)
\label{5.11e}
\end{align}
for a small $\delta$, reconstruction error $-\Delta \mathbf{X}(m) = \mathbf{\hat{Z}}(m)$ as above, and back-propagated reconstruction error.

 The zeroing of partial derivatives at a fixed point of time $t=m$ was done in \cite{LeCun88}\footnote{In this reference, the letter $m$ is used to denote the index of the hidden layers of a multi-layered network and NOT iterative optimization steps, as we do. We have only one hidden layer here and do not need an extra index for the separate layers.}. 

\subsubsection{Case of tied weights: Restricted Boltzman Machine.}

In the special case when  $\mathbf{W}^{(2)}(m)^T = \mathbf{W}^{(1)} (m)$, the $N \times n$ partial derivative with respect to the only weight matrix  $\mathbf{W}^{(1)}$ becomes:
\begin{align}
\frac { \partial \mathcal{S}}{ \partial \mathbf{W}^{(1)}(m)} &= -  \mathbf{\hat{Z}}(m)^T \mathbf{Y}(m)  -  \mathbf{X}(0)^T \mathbf{Z} (m) =0. 
\label{5.11f}
\end{align}
Because $\Delta \mathbf{X}(m) = - \mathbf{\hat{Z}}(m)$, $\Delta \mathbf{Y}(m) = -\mathbf{Z}(m)$, where $\Delta \mathbf{Y}(m) =\mathbf{X}(m)\mathbf{W}^{(1)}(m) - \mathbf{Y}(m),$ from (\ref{5.11c}), (\ref{5.11d}), this being approximately zero translates into:
\begin{align}
\Delta \mathbf{W}^{(1)}(m) \approx -\delta \left( \Delta \mathbf{X}(m)^T  \mathbf{Y} (m) + \mathbf{X}(0)^T \Delta \mathbf{Y} (m)\right),
\label{5.11g}
\end{align}
for a small $\delta$. Up to the higher order term $\Delta \mathbf{X}(m)^T  \Delta \mathbf{Y} (m) $ and (\ref{5.10}), this can be expressed as:
\begin{align}
& \Delta \mathbf{W}^{(1)}(m) \approx \nonumber \\
& \approx -\delta \left( \Delta \mathbf{X}(m)^T  \mathbf{Y} (m) + \mathbf{X}(m)^T \Delta \mathbf{Y} (m) \right) \approx \nonumber \\
& \approx - \delta \left( -\mathbf{X}(0)^T \mathbf{X} (0)\mathbf{W}^{(1)}(m) + \mathbf{X}(m)^T \mathbf{X} (m)\mathbf{W}^{(1)}(m) \right),
\label{5.11h}
\end{align}
the last line resulting from the cancellation of the term $\mathbf{X}(m)^T \mathbf{Y}(m).$ This is the update rule of the Restricted Boltzmann Machine, \cite{Hinton06-2}. It can be approximately expressed in terms of the Gram matrix $\mathbf{G}(m) = \mathbf{X}(m)^T \mathbf{X}(m)$ as in (\ref{5.6g.1}):
\begin{align}
\Delta \mathbf{W}^{(1)}(m) \approx -\delta \Delta \mathbf{G}(m) \mathbf{  W}^{(1)}(m),
\label{5.11i}
\end{align}
for $\Delta \mathbf{G}(m) $ $=\mathbf{G}(m) - \mathbf{G}(0)$.

\bibliographystyle{icml2016}
\bibliography{bibliography}

\begin{appendices}
\section{Error function.}
\label{Error functions}
We will provide the more detailed formulas for the error function from Section \ref{Neural network}.

In expanded form, one can re-write the reconstruction errors introduced in (\ref{0.7})- (\ref{0.7c}) as (recall that $\{x_i\}_{i=1}^{N}$ are column-vectors in $\mathbb{R}^P$ and  $\{x_{\mu}\}_{\mu=1}^{P}$ are row-vectors in $\mathbb{R}^N)$:
\begin{align}
Recon~Err =\sum_{i=1}^N \mathbf{(x_i -\hat{x}_i)^T(x_i -\hat{x}_i)} = \nonumber \\
\sum_{\mu=1}^P \mathbf{(x_\mu -\hat{x}_\mu)(x_\mu -\hat{x}_\mu)^T}, 
\label{0.8}
\end{align}
\begin{align}
&Recon~Err(\mathbf{W^{(1)},W^{(2)}}) = \nonumber \\
 &=\sum_{\mu=1}^P \mathbf{(x_\mu - x_\mu W^{(1)}  W^{(2)}) (x_\mu - x_\mu W^{(1)}  W^{(2)})^T} = \nonumber \\
&=\sum_{i=1}^N \mathbf{(x_i - \sum_{jk}x_{k}} W^{(1)}_{kj}  W^{(2)}_{ji})^T \mathbf{(x_i -  \sum_{jk}x_{k}} W^{(1)}_{kj}  W^{(2)}_{ji})
\label{0.8b}
\end{align}
\begin{align}
&Recon~Err(\mathbf{V^{(1)},V^{(2)}}) = \nonumber \\
&=\sum_{i=1}^N \mathbf{(x_i - V^{(2)}  V^{(1)} x_i} )^T\mathbf{(x_i - V^{(2)}  V^{(1)} x_i} ) = \nonumber \\
 &=\sum_{\mu=1}^P (\mathbf{x_{\mu}} - \sum_{\nu\kappa} V^{(2)}_{\mu \nu}  V^{(1)}_{\nu \kappa} \mathbf{x_{\kappa}}) (\mathbf{x_{\mu}} -  \sum_{\nu\kappa}V^{(2)}_{\mu \nu}  V^{(1)}_{\nu \kappa} \mathbf{x_{\kappa}})^T. 
\label{0.8c}
\end{align}

\section{Back-propagation and partial derivatives of error function.}
\label{Relationship between}

We will show  how back-propagation naturally arises when partial derivatives of error function w.r.t. weight matrices are zero.

The differential of the error  $Err(\mathbf{W^{(1)},W^{(2)}})$  from (\ref{0.7}), (\ref{0.7b}), as a function of $\mathbf{W^{(1)}}$ and $\mathbf{W^{(2)}},$ can be written as:
\begin{align}
dErr = Tr \bigg\{ \frac{\partial Err}{\partial \mathbf{W}^{(1)}} d\mathbf{W}^{(1)} +  \frac{\partial Err}{\partial \mathbf{W}^{(2)}}  d\mathbf{W}^{(2)} \bigg\},
\label{2.2.10}
\end{align}
where,
\begin{align}
 \frac{\partial Err}{\partial \mathbf{W}^{(1)}}^T &= -2 (\mathbf{XW}^{(2)T})^T\mathbf{(X - \hat{X})},
  \label{2.2.3.1}
  \end{align}
  \begin{align}
 \frac{\partial Err}{\partial \mathbf{W}^{(2)}}^T  & = -2 \mathbf{Y}^T\mathbf{(X - \hat{X}) },
 \label{2.2.3.2}
\end{align}
the first equation above being the product of our layer $\mathbf X^T$ and  the "back-propagated error" $\mathbf{(X - \hat{X})W}^{(2)T}.$

The iterative algorithm for computing the transition matrices is given by:
\begin{align}
\Delta \mathbf{W}^{(1)} &=  -\delta \frac{\partial Err}{\partial\mathbf{W}^{(1)}}^T \\
\Delta \mathbf{W}^{(2)} &=  -\delta \frac{\partial Err}{\partial\mathbf{W}^{(2)}}^T,
 \label{2.2.3.3}
\end{align}
for some small $\mathbf {\delta}$ and partial derivatives given by (\ref{2.2.3.1}), (\ref{2.2.3.2}). Obviously this process can continue backwards if our network  had more than two layers.

\section{Dimension reduction with  oscillators. Financial Mathematics interpretation.}
\label{Dimension reduction with}
	
We will rephrase here the problem of Dimension reduction from Section \ref{Dimension reduction} in both physics and financial mathematics terms.

Lets start by interpreting each of the P-dimensional observables (vector-columns) $\{\mathbf{x}_i\}_{i=1}^N$ as 1-dimensional interacting quasi-particles.  Each of the observations (row-vectors) $\{\mathbf{x}_{\mu}\}_{\mu=1}^P$ are discrete snapshots in otherwise continuous time of the locations of the N quasi-particles i.e. the index $\mu$ plays a role of discrete time. The Dimension reduction problem from Section \ref{Dimension reduction} is now the problem of finding n ''synthetic'' quasi-particles ($n < M = rank(\mathbf {x}) < min(P,N)$) which ''approximate'' best the original N quasi-particles i.e. retain most of the energy of the original system.  There is a natural financial mathematics equivalent: simply think of the observables $\{\mathbf{x}_i\}_{i=1}^N$ as ''assets''. The financial formulation of the problem from Section \ref{Dimension reduction} is to ''replicate'' any portfolio of the $N$ original assets with only $n \leq N$ new ''synthetic'' assets i.e. assets which are in turn portfolios of the original assets, \cite{Alexander02}. Replication is in the sense of minimizing the variance of the difference portfolio.

Understanding the \emph{true} dynamics in time i.e in the index of observations $\mu$ is the ultimate goal. Here we will consider a toy model of \emph{oscillators} with non-interacting masses i.e. Euclidean kinetic energy 
and potential energy given by the Gram matrix $\mathbf G$ of the training observables from (\ref{0.1a}). The oscillator has among other benefits the property that its energy is proportional to the trace of the variance of its coordinates. The comparison of its auto-covariance matrix against the Gram matrix of observations $\mathbf G'$ will give us a perspective of how far the dynamics of the training set is from the oscillator dynamics.

Lets consider $N$ coupled oscillator \emph{quasi-particles} given by the $N$ observables (column-vectors) $\{\mathbf{x}_i\}_{i=1}^N$, each of them thought of as 1-dimensional oscillator quasi-particle,  with pairwise interaction (potential energy)  given by their Gram matrix $\mathbf{G} = \mathbf {X^T X}$ (or its inverse) and non-interacting momenta. Every ''synthetic'' quasi-particle now is a super-position of the observables i.e. P-dimensional column-vector (using the basis $\{\mathbf{e}_\mu\}_{\mu=1}^P$, introduced in Section \ref{The training set}):
\begin{align}
\mathbf{q} = \sum_{i=1}^N q_{i} \mathbf{x}_{i} = \sum_{\mu=1}^P q_{\mu} \mathbf{e}_{\mu},
 \label{2.0.0}
\end{align}
where, by (\ref{0.-3a}):
\begin{align}
q_{\mu} = \sum_{i=1}^N X_{\mu i} q_{i}.
\label{2.0.0.1}
\end{align}
The squared norm is given by the definition (\ref{0.1a}) and (\ref{2.0.0.1}):
\begin{align}
\mathbf{||q||}^2  &=\mathbf{q}^T\mathbf{q} = \sum_{\mu=1}^{P} q_{\mu}^2 =  \sum_{i,j=1}^{N}G_{ij} q_i q_j  ,
\label{2.0.0.2}
\end{align}
or:
\begin{align}
\mathbf{||q||}^2 &=  \sum_{i,j=1}^{N}G^{-1}_{ij} q_i q_j.
\label{2.0.0.2a}
\end{align}
We will replace the discrete ''time'' index $\mu$ with a continuous time parameter $t$ i.e. think of our momenta and coordinates as the usual functions of time $\{p_i(t),q_i(t)\},i=1,...,N$. The row-vectors $ \mathbf{p}(t) $ $=\{ p_1(t), p_2(t),...,p_N(t)\}$ and $ \mathbf{q}(t) $ $=\{ q_1(t), q_2(t),...,q_N(t)\}$ become observations at a fixed point of time $t.$ Conversely, for a fixed $i$,  $ p_i(.)$ and $ q_i(.)$ are observables and can be thought of as column-vectors in infinite dimensional space of continuous time.
This model has a natural financial mathematics equivalent: think of the  quasi-particle $\mathbf q$ as a ''portfolio'' of  the assets with weights given by the column-vector $\{q_i\}_{i=1}^N$ and variance $\mathbf{||q||^2}$. 

Let us start with the free-momenta Hamiltonian of the oscillator quasi-particle with non-interacting momenta:
\begin{align}
\mathbf{\mathcal{H}(p,q)} = \frac{1}{2} \sum_{i=1}^{N}p_i^2   + \frac{1}{2} \sum_{i,j=1}^{N}G^{(-1)}_{ij} q_i q_j.  
\label{2.0.1}
\end{align}

For simplicity, we will use in the future the metric (\ref{2.0.0.2}) for interaction of quasi-particles.

 The variation Principle of Least Action postulates that the equations of motion minimize the action, given by:
\begin{align}
\mathbf{S} = \int  \sum_{i=1}^{N}p_i \dot{q}_i   - \mathcal{H}\mathbf{(p,q)} dt, 
\label{2.0.2}
\end{align} 
and from it, one easily derives the familiar Lagrangian equations for the coordinates:
\begin{align}
 \ddot{q_i }   +  \sum_{j=1}^N G_{ij} q_j = 0, i = 1,...,N. 
\label{2.0.3}
\end{align} 
For simplicity of the presentation, we will assume that the rank $M$ of the training matrix is full i.e. $M = N \leq P$. Lets recall that for given initial conditions $p(0), q(0)$, the one-dimensional oscillator $\ddot{q} + \lambda^2 q = 0$ has a solution which in matrix terms is:
\begin{align}
\begin{pmatrix} p(t) \quad q(t) \end{pmatrix} = 
\begin{pmatrix} p(0) \quad  q(0) \end{pmatrix}
\begin{pmatrix} \cos( \lambda t) &   \frac{1}{\lambda} \sin(\lambda t)  \\ 
 - \lambda  \sin(\lambda t) &  \cos (\lambda t)  \end{pmatrix}.
\label{2.0.3.1}
\end{align} 
The solution of our multi-dimensional oscillator is a straightforward generalization for the row-vectors $ \mathbf{p}(t) $ $=\{ p_1(t), p_2(t),...,p_N(t)\}$ and $ \mathbf{q}(t) $ $=\{ q_1(t), q_2(t),...,q_N(t)\}$ at a given time $t$:
\begin{align}
\begin{pmatrix} \mathbf{p}(t) \quad \mathbf{q} (t) \end{pmatrix} &= \nonumber \\
\begin{pmatrix} \mathbf{p}(0) \quad \mathbf{q}(0) \end{pmatrix}
&\begin{pmatrix} \cos(\sqrt{\mathbf{G}} t) &     \sqrt{\mathbf{G}}^{-1} \sin(\sqrt{\mathbf{G}} t) \\ 
 - \sqrt{\mathbf{G}} \sin(\sqrt{\mathbf{G}} t) &  \cos (\sqrt{\mathbf{G}} t)  \end{pmatrix},
\label{2.0.3.2}
\end{align} 
where $\mathbf {\sqrt{\mathbf{G}}}$ is defined as $\mathbf { W \Lambda W^T, }$  where $\mathbf {G }=  \mathbf{W \Lambda^2  W^T}$ is the usual Singular value decomposition (Sub-section \ref{Singular value}) of $\mathbf G$ and the trigonometric matrix functions are resp. the real and imaginary part of the exponential  $e^{i\sqrt{\mathbf{G}}t}$, \cite{Ford65}. The matrix $\mathbf \Lambda^2$ is diagonal with real valued diagonal elements ordered in descending order $ \lambda_1^2 \geq \lambda_2^2 , ... \geq \lambda_N^2 \geq 0$ and $\mathbf W$ is orthogonal i.e.
\begin{align}
 \mathbf {W} := \begin{pmatrix}  
w_{11} & w_{12} &... & w_{1N} \\ 
w_{21} & w_{22} &... & w_{2N} \\ 
. & . &... & . \\ 
w_{N1} & w_{N2} &... & w_{NN} 
\end{pmatrix}, \mathbf{W W^T = I}.
\label{2.0.6}
\end{align}
One can easily verify that  the N orthogonal  column-vectors   $ \mathbf{w}_k$ $= \{w_{1k} ,  w_{2k},...,  w_{Nk} \}^T$ of $\mathbf{W} $ are eigenvectors (or \emph{''eigenstates''} or \emph{''eigen quasi-particles''}) of the Gram matrix  $\mathbf {G}$ with ${\lambda_k^2}$ are the respective eigenvalues i.e. they solve the equation:
\begin{align}
 \mathbf{Gw}_k =  \lambda_k^2 \mathbf{w}_k, k = 1,...,N. 
\label{2.0.5}
\end{align}
Because $e^{i\sqrt{\mathbf{G}}t}$ $  = \mathbf{W} e^{i \mathbf{\Lambda} t} \mathbf{W^T}$, one can factorize the  solution (\ref{2.0.3.2}) and rewrite it in pure diagonal form as follows:
\begin{align}
\begin{pmatrix} \mathbf{p}(t)  \mathbf{W} \quad \mathbf{q} (t)  \mathbf{W} \end{pmatrix} &= \nonumber \\ 
\begin{pmatrix} \mathbf{p}(0)\mathbf{W} \quad \mathbf{q}(0)\mathbf{W} \end{pmatrix}
&\begin{pmatrix} \cos(\mathbf{\Lambda} t) & \mathbf{\Lambda ^{-1}}\sin(\mathbf{\Lambda} t) \\ 
 -\mathbf{\Lambda }\sin(\mathbf{\Lambda} t) & \cos (\mathbf{\Lambda} t)  \end{pmatrix},
\label{2.0.3.3}
\end{align}
There is an obvious canonical transformation  $\{p_i,q_i\}$   $ \rightarrow \{\tilde{p}_k, \tilde{q}_k\}:$
\begin{align}
\mathbf{W: p(t)} \rightarrow  \mathbf{\tilde{p}}(t)   = \mathbf {p}(t) \mathbf{W} \nonumber \\
 \mathbf{q(t)} \rightarrow \mathbf{\tilde{q}}(t)  = \mathbf {q}(t) \mathbf{W} \Leftrightarrow \nonumber \\
\tilde{q}_k(t) = f_k(\mathbf{q}(t)) = \sum_i q_i(t) W_{ik} ,k=1,...,N,
\label{2.0.7.1}
\end{align}
with a dual transformation in the space of observables:
\begin{align}
\mathbf{W: x}_{i}\rightarrow   \sum_{j=1}^{N} \mathbf{x}_j  \mathbf{W}_{ji} 
\label{2.0.7.1.1} 
\end{align}
The respective generating function $\mathbf {\Phi(q,p)}$  is of Type 2:
\begin{align}
\mathbf {\Phi(q,p)} = \sum_{k=1}^N f_k(\mathbf {q}) \mathbf{\tilde{p}}_k
\label{2.0.7.2}
\end{align} 
In standard calculus of variations, \cite{Gelfand63}, this is equivalent to modifying the action (\ref{2.0.2}) by adding the generating function (\ref{2.0.7.2}) where the summands can be thought of as \emph{Lagrange constraints}:
\begin{align}
\mathbf{S} & = \int  \sum_{i=1}^{N}p_i \dot{q}_i   - \mathbf{\mathcal{H}(p,q)}dt + \sum_{k=1}^N f_k(\mathbf {q}) \mathbf{\tilde{p}}_k \nonumber \\
  & = \int \sum_{i=1}^{N}\tilde{p}_i \dot{\tilde{q}}_i   - \mathbf{\mathcal{H}(\tilde{p},\tilde{q})}dt 
\label{2.0.7.3}
\end{align} 
In the new coordinates $\{\tilde{p}_k,\tilde{q}_k\},k=1,...,N$, the Hamiltonian is diagonalized i.e. we have a superposition of independent oscillations:
\begin{align}
\mathbf{\mathcal{H}(\tilde{p},\tilde{q})} = \frac{1}{2} \sum_{k=1}^{N} \tilde{p}_k^2   + \frac{1}{2} \sum_{k=1}^{N} \lambda_k^2  \tilde{q}_k \tilde{q}_k.  
\label{2.0.8}
\end{align}
We are ready to deal with the free-momenta Dimension reduction problem in energy terms (as articulated in Section \ref{Dimension reduction}). The conserved total energy of the quasi-particle equals its potential energy $\frac{1}{2} \sum_{i,j=1}^{N}G_{ij} q_i q_j$. In particular, in the $k$-th principal state, the energy is $\frac{\lambda_k^2}{2}$. In general, our quasi-particle is found in a super-position of its principal states, given by some arbitrary superposition vector $\{\xi_i\}_{k=1}^{N}$ of unit norm $\mathbf{\xi}^T \mathbf{\xi} = 1,$ with respective energy proportional to:
\begin{align}
\mathbf{ E(\xi)} \sim \frac{1}{2}   \sum_{k=1}^{N} \xi_i \lambda_i^2.
\label{2.0.8.1}
\end{align}
The  maximum contribution to energy comes from the n-dimensional subspace spanned by $n$ eigenvectors corresponding to the n largest eigenvalues $\lambda_1^2 \geq \lambda_2^2 \geq ... \lambda_n^2$ i.e. by the column vectors of the matrix:
\begin{align}
 \mathbf {\hat{W}} := \begin{pmatrix}  
w_{11} & w_{12} &... & w_{1n} \\ 
w_{21} & w_{22} &... & w_{2n} \\ 
. & . &... & . \\ 
w_{N1} & w_{N2} &... & w_{Nn} 
\end{pmatrix}, \mathbf{\hat{W} \hat{W}^T = I}.
\label{2.0.9}
\end{align}
We will refer to it as \emph{reduced principal space}. The new truncated Gram  matrix is:
\begin{align}
\mathbf {\hat{G}  }= \mathbf {\hat{W} \hat{\Lambda}^2 \hat{W}^T}
\label{2.0.10}
\end{align}
where $\mathbf {\hat{\Lambda}^2}$ is the diagonal matrix with diagonal elements $\{ \lambda_1^2 , \lambda_2^2 , ... \lambda_n^2 \}$. Dimension reduction amounts to keeping only the top n energy-contributing solutions from (\ref{2.0.3.2}):
\begin{align}
\begin{pmatrix} \mathbf{\hat{p}}(t) \quad \mathbf{\hat{q}} (t) \end{pmatrix} &= \nonumber \\ 
\begin{pmatrix} \mathbf{p}(0) \quad \mathbf{q}(0) \end{pmatrix}
&\begin{pmatrix} \cos(\sqrt{\mathbf{\hat{G}}} t) &     \sqrt{\mathbf{\hat{G}}}^{-1} \sin(\sqrt{\mathbf{\hat{G}}} t) \\ 
 - \sqrt{\mathbf{\hat{G}}} \sin(\sqrt{\mathbf{\hat{G}}} t) &  \cos (\sqrt{\mathbf{\hat{G}}} t)  \end{pmatrix},
\label{2.0.11}
\end{align}

\begin{align}
\mathbf {\hat{Z} } &= \mathbf{\hat{y} \hat{Q}}, \nonumber \\
 \mathbf {\hat{y} } &= \mathbf{x \hat{W}}, \mathbf{\hat{Q}} = Re \{ e^{i \mathbf{\hat{\Lambda}}^2 t} \}.
\label{2.0.11}
\end{align} 
The degree of success of our Dimension reduction is measured by the proximity to 1 of the ratio $R^2$, defined as:
\begin{align}
0  \leq R^2 := \frac{Tr(\hat {\mathbf{G}})}{Tr(\mathbf{G})} = \frac{\sum_{i=1}^n\lambda^2_i} {\sum_{i=1}^N\lambda^2_i} = \frac{\hat{\Lambda}^2}{\Lambda^2} \leq 1.
\label{2.0.12}
\end{align}
Let us get back to part of the original coordinates via the  ''partial reverse'' canonical transformation:
\begin{align}
p_k &=  f^{-1}_k f_k(\mathbf{p}) = \mathbf { \tilde{p} \hat{W}}, k=1,...,n, \nonumber \\
\hat{q}_k &=  f^{-1}_k f_k(\mathbf{q}) = \mathbf { \tilde{q} \hat{W}}, k=1,...,n, \nonumber \\
\hat{p}_k &= \tilde{p}_k, k = n+1, ..., N, \nonumber \\
\hat{q}_k &= \tilde{q}_k, k = n+1, ..., N.
\label{2.0.13}
\end{align}
We have free motion in the last orthogonal $N - n - 1$ dimensions and the Hamiltonian of the new quasi-particle in $N$ dimensions is:
\begin{align}
\mathbf{\hat{\mathcal{H}}(\hat{p},\hat{q})} &= \frac{1}{2} \sum_{i=1}^{n} p_i^2   + \frac{1}{2} \sum_{i,j=1}^{n}\hat{G}_{ij} \hat{q}_i \hat{q}_j + \mathbf{\hat{\mathcal{H}}_{noise}},
\label{2.0.14}
\end{align}
subject to constraint (see definition in (\ref{2.0.12})):
\begin{align}
 \mathbf{\hat{\mathcal{H}}_{noise}} &=  \frac{1}{2} \sum_{i=n+1}^{N}\tilde{p}_i^2 \leq  \frac{1}{2} (\Lambda^2 - \hat{\Lambda}^2).
\label{2.0.14a}
\end{align}
This is not a closed system in $n$ dimensions: the free momenta $\{\tilde{p}_i\}_{i=n+1}^N$ can take arbitrary values, subject to the constraint in (\ref{2.0.14a}). Because the term $\mathbf{\hat{\mathcal{H}}_{noise}}$ changes the energy arbitrarily, our $N$-dimensional quasi-particle can drift from one energy level to another.

\end{appendices}

\end{document}